\PassOptionsToPackage{table}{xcolor}

\documentclass[journal]{IEEEtran}

\usepackage[table]{xcolor}
\usepackage{times}
\usepackage{siunitx}
\usepackage{epsfig}
\usepackage{graphicx}
\usepackage{amsmath}
\usepackage{amssymb}
\usepackage{placeins}
\usepackage[noend]{algpseudocode}
\usepackage{algorithm,algorithmicx}
\usepackage{lipsum}  
\usepackage[labelformat=simple]{subcaption}

\usepackage[percent]{overpic}
\usepackage{mathtools}
\usepackage{amsfonts,amsmath,amssymb, amsthm}
\usepackage{colortbl}
\usepackage{booktabs}
\usepackage{gensymb}
\usepackage{slashbox}
\usepackage{url}
\usepackage{cite}
\usepackage{titling}

\makeatletter
\DeclareRobustCommand\onedot{\futurelet\@let@token\@onedot}
\def\@onedot{\ifx\@let@token.\else.\null\fi\xspace}

\makeatother

\newcommand{\Section}[1]{\vspace{0pt}\section{#1}\vspace{0pt}}

\renewcommand{\paragraph}[1]{\vspace{2 pt}\noindent\textbf{#1}}

\newcommand{\edit}[1]{\textcolor{black}{#1}}
\newcommand{\editTwo}[1]{\textcolor{black}{#1}}

\def\mathclap#1{\text{\hbox to 5pt{\hss$\mathsurround=0pt#1$\hss}}}

\graphicspath{{.}{./figures/}}

\begin{document}

\title{Keyhole Imaging:\\Non-Line-of-Sight Imaging and Tracking of Moving Objects Along a Single Optical Path}
\date{}

\author{Christopher~A.~Metzler,
	David~B.~Lindell,
	and~Gordon~Wetzstein
	\thanks{C.~Metzler, D.~Lindell, and G.~Wetzstein are with the Department
		of Electrical Engineering, Stanford University, Stanford,
		CA, 94305 USA.}
	\thanks{e-mail: \{cmetzler, gordon.wetzstein\}@stanford.edu.}
}
\renewcommand\footnotemark{}
\renewcommand\footnoterule{}
%
%
%


\maketitle

\begin{abstract}
Non-line-of-sight (NLOS) imaging and tracking is an emerging technology that allows the shape or position of objects around corners or behind diffusers to be recovered from transient, time-of-flight measurements. However, existing NLOS approaches require the imaging system to scan a large area on a visible surface, where the indirect light paths of hidden objects are sampled. In many applications, such as robotic vision or autonomous driving, optical access to a large scanning area may not be available, which severely limits the practicality of existing NLOS techniques. Here, we propose a new approach, dubbed keyhole imaging, that captures a sequence of transient measurements along a single optical path, for example, through a keyhole. Assuming that the hidden object of interest moves during the acquisition time, we effectively capture a series of time-resolved projections of the object's shape from unknown viewpoints. We derive inverse methods based on expectation-maximization to recover the object's shape and location using these measurements. Then, with the help of long exposure times and retroreflective tape, we demonstrate successful experimental results with a prototype keyhole imaging system.
\end{abstract}


\IEEEpeerreviewmaketitle

\newcommand{\albedo}{\rho}
\newcommand{\xlocal}{{\bf x'}}
\newcommand{\xglob}{{\bf x}}
\newcommand{\trans}{\boldsymbol{\theta}}

\newcommand{\noise}{\boldsymbol{\eta}}
\newcommand{\meas}{\mathbf{y}}

\Section{Introduction}
\label{sec:introduction}
\begin{figure}[t]
	\centering
	\begin{overpic}[width=0.995\linewidth]{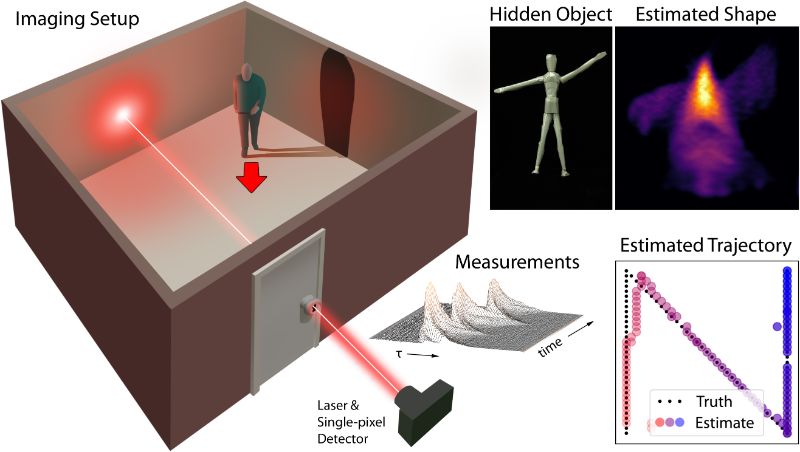}	
		\put (75,7) {{\scriptsize $\uparrow$}}
		\put (75,4) {{\scriptsize $x$}}				
		\put (80,-1) {{\scriptsize $z \rightarrow$}}
	\end{overpic}	
	\caption{\textbf{Keyhole imaging.} A time-resolved detector and pulsed laser illuminate and image a point visible through a keyhole (left). As a hidden person moves, the detector captures a series of time-resolved measurements of the indirectly scattered light (center). From these measurements, we reconstruct both hidden object shape (e.g.,~for a hidden mannequin) and the time-resolved trajectory (right).}
	\label{fig:teaser}
\end{figure}

Imaging or tracking objects outside a camera's direct line of sight has important applications in autonomous driving, robotic vision, and many other areas. By analyzing the time-of-flight of indirectly scattered light, transient non-line-of-sight (NLOS) imaging approaches (e.g.,~\cite{Velten:2012recovering,buttafava2015non,Gariepy:2016,OToole:2018,xin2019theory,Liu:2019,Lindell:2019:Wave,faccio2020non,young2020non}) are a particularly promising approach to seeing around corners at large scales. However, all of these techniques rely on time-resolved measurements of indirect light transport that are captured by sampling a large area of a surface within the line of sight of the imaging system. The sampled area acts as a synthetic aperture and it was shown that the resolution of estimated hidden scenes, and also the accuracy of NLOS tracking, is primarily limited by both the size of the sampling area and the temporal resolution of the imaging system~\cite{OToole:2018}. Unfortunately, in many applications, such as navigation in tight quarters or at long standoffs, optical access to a large sampling surface may not be feasible, making current NLOS techniques impractical.\footnote{\editTwo{Another barrier to eye-safe, real-time NLOS imaging is light throughput. The challenging radiometry associated with keyhole imaging with diffuse objects is discussed in the supplement.}}

Here, we introduce a new form of NLOS imaging where transient measurements are recorded along a single optical path. As shown in Figure~\ref{fig:teaser}, one example of an optical configuration that is enabled by our approach is that of sampling a single point on the wall of a room through a keyhole. Although imaging through a keyhole may be possible by moving a small camera very close to the keyhole, our approach operates at a standoff distance from the keyhole and only requires a single optical path from the light source--detector pair to the sampling point inside the room.

Trying to recover the shape or to track the location of objects in a static room with such a configuration is a highly ill-posed problem.
However, as was recognized by Bouman et al.~\cite{bouman2017turning}, object motion can make otherwise ill-posed NLOS imaging problems tractable.
Unlike their work, however, keyhole imaging only has access to a single optical light path. 
To account for this, our work makes use of the intuition that {\em measurements of a moving object captured with a fixed sensor are equivalent to measurements of a fixed object captured with a moving ``virtual'' sensor}. This is the same principle that underlies problems in other domains, such as inverse synthetic aperture radar (ISAR)~\cite{prickett1980principles}. Just as NLOS imaging is analogous to SAR and tomography~\cite{liu2019analysis}, keyhole imaging of a moving object is analogous to ISAR and unknown-view tomography. Like ISAR and unknown-view tomography, keyhole imaging requires solving a challenging and highly non-convex inverse problem: jointly estimating both the shape of the hidden object and its unknown motion. We address this problem using a variant of the expectation-maximization (EM) algorithm.

To our knowledge, this is the first work to study the keyhole imaging problem, which could enable an entirely new class of NLOS imaging problems to be addressed. In doing so, this work makes several distinct contributions:
\begin{itemize}
	\item We introduce the keyhole imaging problem: NLOS imaging or tracking where the relay wall consists of just a single visible point.
	\item We develop a Bayesian EM-based algorithm that uses the unknown motion of the object to solve the keyhole imaging problem. 
	\item We evaluate the performance of this method for various types of motion, which model both translation and rotation, in extensive simulations.
	\item We build a prototype system and experimentally demonstrate NLOS imaging and tracking through a keyhole.
\end{itemize}

\paragraph{Assumptions and Limitations}
Our present work relies on several assumptions and simplifications that help in tackling the challenging keyhole imaging problem. First, we assume the hidden object exhibits only rigid body motion. Second, we assume the object does not self-occlude nor produce interreflections. Third, to greatly reduce computational complexity, our present reconstructions of the trajectories and the objects (even those that are non-planar, like the mannequin in Figure \ref{fig:teaser}) are 2D. Fourth, it is worth pointing out that our reconstructions have an inherent flip ambiguity. For example, measurements from an object moving from left to right are indistinguishable from measurements of horizontally flipped but otherwise identical object moving from right to left.
\editTwo{Similarly, our method cannot distinguish between an object rotating in place and an object translating along a circular trajectory.} Finally, because of the limited power of our laser source, we covered the hidden objects in retroreflective tape and set the total exposure time to a few minutes so as to increase the strength of the returning signals. \edit{As a result, our reconstructions have access to 2--5$\times$ more photons than one would detect when imaging a diffuse, white, one square meter hidden objects at a two meter standoff with a barely eye-safe 9.9 mW 1550 nm laser over a five second exposure time. See the supplement 
for more information about eye-safe laser power limits and the radiometry associated with room-scale keyhole imaging of diffuse and retroreflective objects.}

\Section{Related Work}
\label{sec:related}

\paragraph{NLOS Imaging and Tracking}
While long theorized~\cite{freund1990looking,Kirmani:2009}, NLOS imaging of static objects was first successfully demonstrated in 2012~\cite{Velten:2012:Visualizing}. Since then, much progress has been made on developing more efficient data acquisition setups and inverse methods~\cite{faccio2020non}. The majority of these works reconstruct hidden scenes from transient measurements captured using pulsed~\cite{Pandharkar:2011,Velten:2012recovering,Wu:2014,Gupta:12,laurenzis2015multiple,tsai2017geometry,arellano2017fast,pediredla2017reconstructing,OToole:2018,Xu:18,buttafava2015non,jin2015recovering,OToole:2018,heide2019non,xin2019theory,Liu:2019,Lindell:2019:Wave,young2020non} or continuous-wave~\cite{heide2014diffuse,kadambi2016occluded,xin2019theory} active illumination, or using a coherent source that creates speckle~\cite{bertolotti2012non,katz2012looking,katz2014non,PrasannaCosi,metzler2020deep}. Acoustic NLOS imaging has also been explored~\cite{lindell2019acoustic,an2019diffraction}, as have methods that forgo active illumination and instead rely opportunistically on occlusions or illumination sources in the scene itself~\cite{torralba2012accidental,bouman2017turning,batarseh2018passive,Thrampoulidis:2018,saunders2019computational}. All these works require a large relay wall to sample the indirect light transport inside their line of sight.

More recently, several groups have developed methods to image or track moving hidden objects. This feat can be accomplished by either ignoring motion and using powerful illumination to form frame-by-frame triangulations~\cite{brooks2019single} or reconstructions~\cite{o2018real,Lindell:2019:Wave} in real-time or by leveraging the differences between subsequent frames with object motion to track them~\cite{klein2016tracking,Gariepy:2016,chan2017non,bouman2017turning,smith2018tracking,Boger:2019}. These methods too rely upon large relay walls.

Keyhole imaging is a new NLOS imaging modality. Unlike other NLOS approaches, it only requires access to a single optical light path. The keyhole imaging inverse problem is significantly more challenging than conventional NLOS imaging because the locations of the sampling points are unknown.

\paragraph{Unknown-view Tomography} The problem of reconstructing a 3D or 2D object from a series of 2D or 1D projections, respectively, taken at unknown view angles is known as unknown-view tomography~\cite{basu2000uniqueness}. This problem occurs in a number of applications; most notably inverse synthetic aperture radar (ISAR)~\cite{prickett1980principles} and single-particle cryogenic electron microscopy~\cite{cheng2015primer,singer2018mathematics}. 

A host of methods exist to solve the unknown-view tomography problem: Moment methods compute perspective invariant features and use these features to directly reconstruct the scene~\cite{wang2019two,bandeira2017optimal,zehni2019geometric}.
Low-dimensional embedding methods estimate the perspective associated with each measurement by embedding it into a low-dimensional space, this estimated perspective then allows for the application of standard recovery algorithms~\cite{coifman2008graph,singer2013two}. Bayesian methods, which are the most popular and successful approach to unknown-view tomography, compute maximum likelihood (ML) or maximum a posteriori (MAP) estimates of the scene under priors on the distribution of the scene, the measurements, and the noise~\cite{basu2000feasibility,scheres2012relion}. Bayesian methods are sample efficient and highly robust to noise, but can be computationally demanding.

Keyhole imaging is a particularly challenging unknown-view tomography problem as it reconstructs an image of a 3D object from unregistered 1D measurements. Nevertheless, we found Bayesian algorithms could be adapted to solve this problem. To overcome their high computational cost we took advantage of GPU computing.

\Section{Keyhole Imaging}
\label{sec:method}
The keyhole imaging setup is illustrated in Figure~\ref{fig:teaser}.
A confocal pulsed light source--detector pair, placed at a standoff distance from the door, illuminates and images a visible point through a small aperture (such as a keyhole). The time-resolved measurements, captured with a single-photon avalanche diode (SPAD) or other detector, contain the temporal response of the emitted light pulse that travels to a visible point through the keyhole, and scatters back from a hidden object. A series of measurements, captured as the hidden object undergoes rigid motion, are used to recover the shape and motion of the hidden object.

In this section, we describe the keyhole measurement model in detail and derive a Bayesian estimation method for recovering the shape and motion of hidden objects. 

\subsection{Observation Model}
Assume that the hidden object is defined by a volumetric albedo $\albedo \left( \xlocal \right)$, where $\xlocal = [x', y', z']^T$ is the object's local coordinate system. \edit{Assuming the object exhibits only rigid body motion,} a transform $\trans_l = \left[ \mathbf{R}_l | \mathbf{t}_l \right] \in \mathbf{R}^{3 \times 4}$ describes the motion of the object at some time $l$ and transforms it into the global coordinate system, using rotation and translation, as $\xglob = \mathbf{R}_l \xlocal + \mathbf{t}_l$. Each time-resolved measurement $\meas_l \in \mathbb{R}^{T}$, represented by a histogram with $T$ bins counting the number of photon-arrival events within a certain time window, is defined by a formation model 
\begin{align}
\meas_l=f(\albedo,\mathbf{\trans}_l)+\noise_l,
\end{align}
where $\noise_l$ denotes (potentially signal dependent) noise and $f(\albedo,\mathbf{\trans}_l)$ is the NLOS imaging forward model
%
\begin{equation}
f(\albedo,\mathbf{\trans}_l) = \int \frac{1}{g \left( \xglob \right) }\albedo \left( \xlocal \right) \delta \left(2 \left\| \xglob \right\|_2 - \tau c \right) d \xlocal.
\label{eqn:keyholemodel}
\end{equation}
Here, $c$ is the speed of light, $\tau$ is the time relative to an emitted laser pulse, the point at which we record the scene is located in the origin of the global coordinate system, and $g \left( \xglob \right)$ describes the falloff of light with distance. For example, hidden objects with Lambertian reflectance exhibit a falloff of $g_{\textrm{diffuse}} \left( \xglob \right) = \left\| \xglob \right\|_2^4$ whereas perfectly retroreflective objects experience $g_{\textrm{retro}} \left( \xglob \right) = \left\| \xglob \right\|_2^2$~\cite{OToole:2018}. Moreover, $g$ can also include angle-dependent factors that are influenced by the surface normals, the normal of the visible wall, or other angle-dependent reflectance characteristics. Using experimental measurements, we found that $g_{\textrm{exp}} \left( \xglob \right) = \left\| \xglob \right\|_2^4 \cos^{-4}(\phi)$, where $\phi$ is the angle between the wall's normal and $\xglob$, best modeled the falloff associated with a patch, oriented parallel to the visible wall, of the store-bought retroreflective tape that we used in our captured results. \edit{See the supplement  
for more information about the origin of a quartic falloff models with retroreflective objects, how we fit the model to our data, and the effects of model mismatch on our reconstructions.}

Note that \eqref{eqn:keyholemodel} is the standard confocal NLOS model~\cite{OToole:2018}, with the hidden object's position transformed by $\trans_l$. \edit{This formulation does not model self-occlusion and in this work we assume the hidden object does not self-occlude.} The keyhole imaging reconstruction problem uses a series of measurements $\meas_1, \ldots,\meas_L$ to reconstruct the albedo $\albedo$ in its local coordinate system. This problem is challenging because the measurements are parameterized by latent hidden object locations, $\trans_1,\ldots,\trans_L$ corresponding to each captured measurement. (In practice, we parameterize each measurement by the position of an isotropic virtual sensor during each measurement. This parameterization allows us to represent rotation and translation of the hidden object as translation of the virtual sensor.)

\subsection{Reconstruction with Expectation-Maximization}

We seek to recover an estimate of the unknown albedo, $\albedo$ from the observations $\mathbf{y}_1,...\mathbf{y}_L$.
This can be done by maximizing the log likelihood of the observed measurements 
\begin{align}\label{eqn:ML}
\mathcal{L}(\albedo; \mathbf{y}) = \log p(\mathbf{y}|\albedo)=\log \int_{\mathbf{\trans}}p(\mathbf{y},\mathbf{\trans}|\albedo)\,d\mathbf{\trans},
\end{align}
where $\mathbf{y}=[\mathbf{y}_1,...\mathbf{y}_L]$ and $\mathbf{\trans}=[\mathbf{\trans}_1,...\mathbf{\trans}_L]$. 

Maximizing the objective \eqref{eqn:ML} directly is numerically unstable; without an excellent initialization, $p(\mathbf{y}|\albedo)$ will start near $0$ and one is tasked with maximizing the $\log$ of $0$. 
Instead, in this work we utilize EM, which efficiently maximizes \eqref{eqn:ML} by solving a series of easy-to-solve least squares problems. 
In particular, EM iteratively applies two steps: (1) an expectation step in which an estimated conditional distribution of $\mathbf{\trans}$ is used to form a lower bound to the log likelihood, and (2) a maximization step, which estimates the albedo given the current conditional distribution of $\mathbf{\trans}$.

\vspace{0.8em}\noindent\textbf{EM Algorithm for Keyhole Imaging}

Given an estimate $\albedo^{(n)}$ of the hidden object's albedo at iteration $n$ of the EM algorithm, we perform the expectation step by finding a lower bound on the log of the likelihood given by \eqref{eqn:ML}. Using Jensen's inequality, it can be shown that a lower bound is given as $Q(\albedo,\albedo^{(n)})$~\cite{dellaert2002expectation}, where, up to additive constants, 
\begin{align}\label{eqn:E_function}
Q(\albedo,&\albedo^{(n)})=\mathbb{E}_{\trans|\mathbf{y},\albedo^{(n)}}[\log p(\mathbf{y},\mathbf{\trans}|\albedo)], \\
&=\int_{\mathbf{\trans}}p(\mathbf{\trans}|\mathbf{y},\albedo^{(n)})\log[ p(\mathbf{y}|\mathbf{\trans},\albedo)p(\mathbf{\trans}|\albedo)]d\mathbf{\trans},\nonumber\\
&=\sum_{l=1}^L \int_{\mathbf{\trans}_l}p(\mathbf{\trans}_l|\mathbf{y}_l,\albedo^{(n)})\log[ p(\mathbf{y}_l|\mathbf{\trans}_l,\albedo)p(\mathbf{\trans}_l|\albedo)]d\mathbf{\trans}_l,\nonumber\\
&\approx \sum_{l=1}^L \sum_{\mathbf{\trans}_{l,k} \in \Omega}p(\mathbf{\trans}_{l,k}|\mathbf{y}_l,\albedo^{(n)})\log[ p(\mathbf{y}_l|\mathbf{\trans}_{l,k},\albedo)p(\mathbf{\trans}_{l,k}|\albedo)],\nonumber \\
&= \sum_{l=1}^L \sum_{\mathbf{\trans}_{l,k} \in \Omega}p(\mathbf{\trans}_{l,k}|\mathbf{y}_l,\albedo^{(n)})\log[ p(\mathbf{y}_l|\mathbf{\trans}_{l,k},\albedo)]. \nonumber
\end{align}
Here, $\Omega$ is the domain of possible hidden object positions (parameterized as translations of a virtual sensor); $p(\mathbf{\trans}_{l,k}|\mathbf{y}_l,\albedo^{(n)})$ is the probability the object is at position $\mathbf{\trans}_{l,k}$ during the $l^{th}$ measurement conditioned on the measurement $\mathbf{y}_l$ and the previous estimate of the object's albedo $\albedo^{(n)}$; and $p(\mathbf{y}_l|\mathbf{\trans}_{l,k},\albedo)$ is the likelihood the measurement $\mathbf{y}_l$ would be observed, conditioned on the object being at position $\mathbf{\trans}_{l,k}$ and the current estimate of the albedo ${\albedo}^{(n)}$. 

In the series of equations following~\eqref{eqn:E_function}, the second line depends on Bayes' rule; the third line holds by assuming the observations are independent; the fourth line uses a discrete approximation of the hidden object location, which is required for our numerical implementation; and the last line's equality holds by assuming that the object's location is independent of its albedo and by dropping an additive $p(\mathbf{\trans}_{l,k})$ term, which does not affect the optimal value of $\albedo$.

In the case of additive white Gaussian noise with variance $\sigma^2$, \eqref{eqn:E_function} simplifies to 
\begin{flalign}\label{eqn:SimplifiedQ}
Q(\albedo,\albedo^{(n)})&\propto \sum_{l=1}^L \sum_{\mathbf{\trans}_{l,k} \in \Omega} w_{\trans_{l,k}}{ \Big (}-\|\mathbf{y}_l-f(\albedo,\mathbf{\trans}_{l,k})\|^2_2{\Big )},
\end{flalign}
where $w_{\trans_{l,k}}=\exp \frac{-\|\mathbf{y}_l-f(\albedo^{(n)},\mathbf{\trans}_{l,k})\|^2_2}{2\sigma^2}$.
(While EM allows for a more accurate Poisson noise model to be used, an i.i.d.~Gaussian model reduces the computational complexity of the algorithm.)
Intuitively, this result computes the sum of squared differences between the measurements and predicted measurements given the current estimate of $\albedo$ and each possible hidden object location. Elements of the sum are weighted by $w_{\trans_{l,k}}$, which is proportional to the conditional probability that the hidden object is at location $\mathbf{\trans}_{l,k}$ for measurement $\mathbf{y}_l$.

Then, the maximization step of the algorithm computes
\begin{align}
\albedo^{(n+1)}=\arg\max_\albedo Q(\albedo,\albedo^{(n)}),
\end{align}
for $n=1...N$. This optimization can be accomplished using gradient descent until convergence.

We summarize the EM algorithm for keyhole imaging in Algorithm~\ref{alg:EM}. 
At a high level, the two steps of EM for keyhole imaging can be understood as follows. First, based on the current estimate of $\albedo$, line~\ref{alg:w} of the algorithm performs the expectation step and estimates the probability of measurement $\mathbf{y}_l$ being captured at each of the possible hidden object locations $\mathbf{\trans}_{l,k}$. This step can also be interpreted as performing a soft assignment of the object position at each iteration. Next, line~\ref{alg:max} computes the maximization step: Based on the soft assignment of hidden object location, the algorithm updates the estimate of the hidden object albedo. Iterating these two steps until convergence produces the final estimate of the hidden object albedo, while also recovering an estimate of the distribution of the hidden object's locations during each measurement.


\begin{algorithm}[t]{
		\caption{EM for Keyhole Imaging}
		\begin{algorithmic}[1]
			\State Initialize: $\albedo^{(0)}$
			\For{n=0, 1, ..., N-1}
			\State Compute $w_{\trans_{l,k}}=\exp \frac{-\|\mathbf{y}_l-f(\albedo^{(n)},\mathbf{\trans}_{l,k})\|^2_2}{2\sigma^2}$~$\forall$~$i,k$ \label{alg:w}
			\State $\albedo^{(n+1)}=\arg\max_\albedo Q(\albedo,\albedo^{(n)})$ \label{alg:max}
			\EndFor
			\State Return $\albedo^{(N)}$
		\end{algorithmic}
		\label{alg:EM}
	}
\end{algorithm}

\paragraph{Adding Priors}
EM can be extended to computing MAP estimates by redefining $Q(\albedo,\albedo^{(n)})$ as
\begin{flalign}\label{eqn:SimplifiedQ_MAP}
&{\Big (}\sum_{l=1}^L \sum_{\mathbf{\trans}_{l,k} \in \Omega} -w_{\trans_{l,k}}\|\mathbf{y}_l-f(\albedo,\mathbf{\trans}_{l,k})\|^2_2 {\Big )}+ \lambda \log p(\albedo) ,
\end{flalign}
where $p(\albedo)$ is a prior on $\albedo$. In this work, we use $\log p(\albedo)=-\|\mathbf{L}\albedo\|_1 - \|\albedo\|_1$, where $\mathbf{L}$ is a Laplacian filter. This expression corresponds to a prior that the hidden object's albedo is smooth and sparse.

\paragraph{Implementation Details}
\edit{We developed an implementation of EM using the GPU-accelerated and free-to-use PyTorch library~\cite{NEURIPS2019_9015}.} We ran EM for 30 iterations. 
Each maximization step was accomplished by running the ADAM optimizer~\cite{kingma2014adam} $n+1$ times, where $n$ is the EM iteration number. During the early iterations of the algorithm, when our estimate of $w$ is unreliable, we do not spend much time optimizing $\albedo$.
We set ADAM's learning rate to $0.1$ and set its $\beta$ values, which control the momentums' decay, to $0.5$ and $0.999$.
We set the regularization parameter $\lambda$ to $2\,000$ for the experimental data and $0$ for the simulated data. 
The noise variance $\sigma$ was set to $200$ for both simulated and experimental data; this value is far larger than the true noise variance, but helped account for model mismatch and the discretization of the trajectory support set $\Omega$. 

During simulations, $\Omega$ was a  $33\times 33$ equispaced grid spanning 1 meter by 1 meter. 
For experiments, $\Omega$ was a $33\times 33$ grid spanning 1 m along the axis parallel to the wall and 15 cm along the axis perpendicular to the wall; this was the range of our translation stages.
We enforced positivity on $\albedo$ by parameterizing it with the variable $\nu$, with $\albedo:=\nu^2$ where the square is taken elementwise.
The variable $\nu$ was initialized with an i.i.d.~Gaussian vector \edit{with mean 0 and variance 1, which is roughly the same as the variance of $\nu$ after the algorithm has converged.}

We found that deterministic annealing helped EM avoid local minima \cite{ueda1998deterministic}.
With deterministic annealing 
\begin{align}
w_{\trans_{l,k}}={\Big [}\exp \frac{-\|\mathbf{y}_l-f(\albedo^{(n)},\mathbf{\trans}_{l,k})\|^2_2}{2\sigma^2}{\Big ]}^\beta,
\end{align}
where $\beta\in (0,1]$ is an inverse temperature parameter that increases iteration to iteration. We set $\beta^{(n+1)}=1.3\cdot\beta^{(n)}$, with $\beta^{(0)}$ set such that $\beta^{(N-1)}=1$. Annealing serves to make the distribution of the estimated object locations more uniform during the earlier iterations of the algorithm, when the algorithm has a less accurate estimate of $\albedo$. \edit{Accordingly, as demonstrated in the supplement, with annealing EM is insensitive to how it is initialized.} 


\Section{Evaluation and Analysis}
\label{sec:analysis}

\subsection{Simulation Setup}\label{ssec:SimSetup}

\begin{figure}[t]
	\centering
	
	\begin{subfigure}[t]{.15\textwidth}
		\centering
		\begin{overpic}[width=\textwidth]{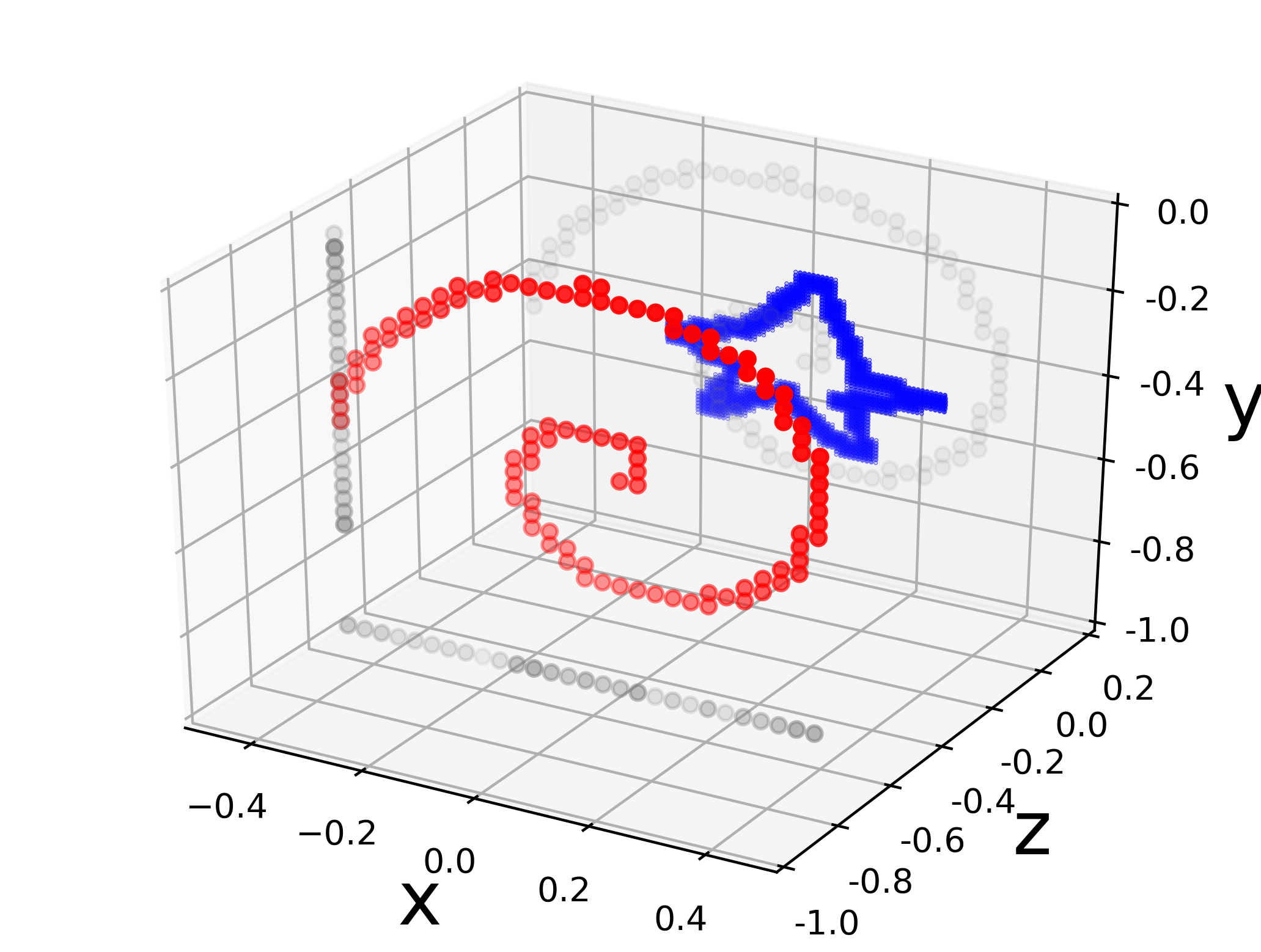}
			\put (10,80)  {Constant $z$ Plane}
		\end{overpic}
		\caption{}
	\end{subfigure}%
	~
	\begin{subfigure}[t]{.15\textwidth}
		\centering
		\begin{overpic}[width=\textwidth]{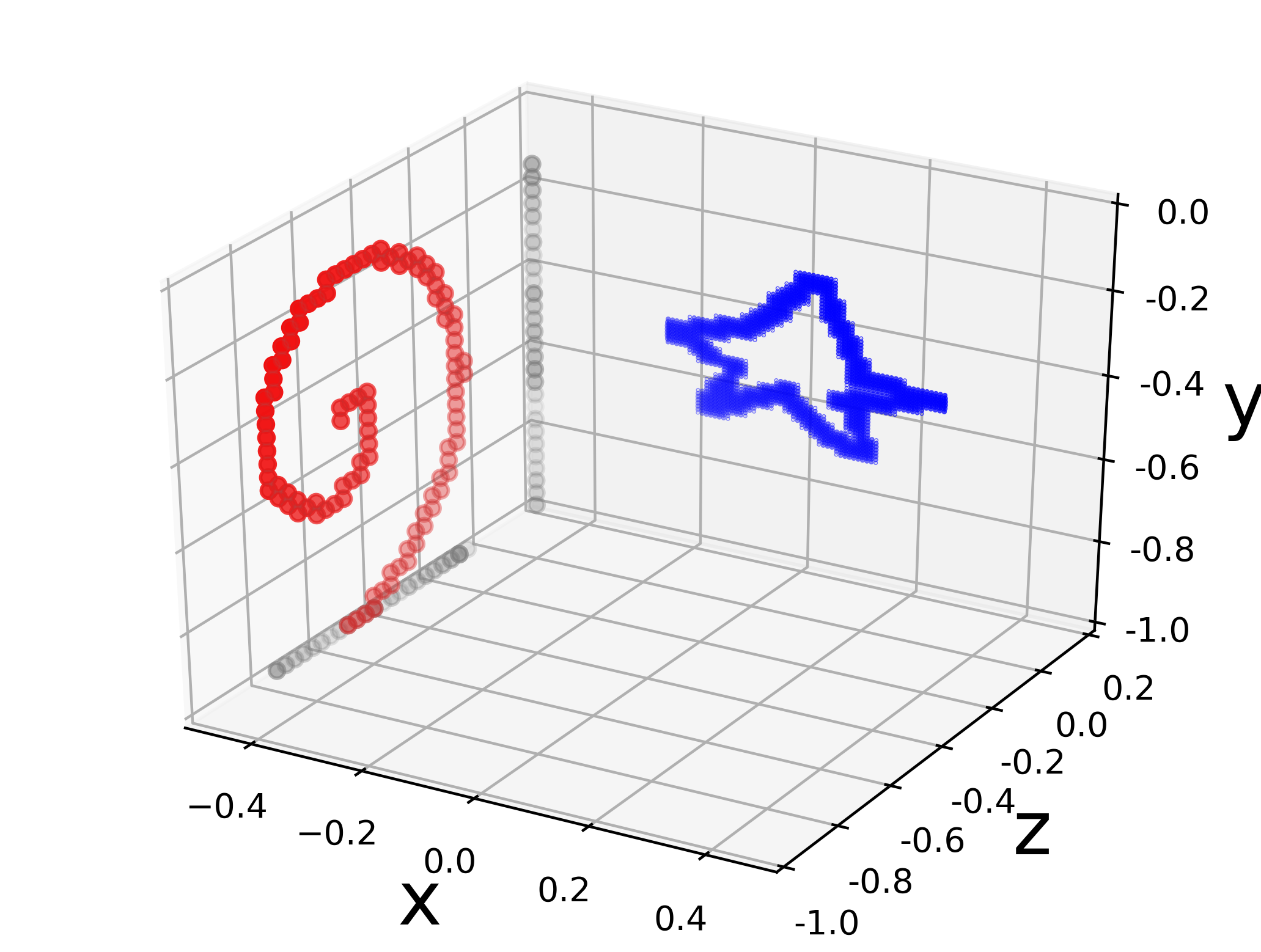}
			\put (10,80)  {Constant $x$ Plane}
		\end{overpic}
		\caption{}
	\end{subfigure}%
	~
	\begin{subfigure}[t]{.15\textwidth}
		\centering
		\begin{overpic}[width=\textwidth]{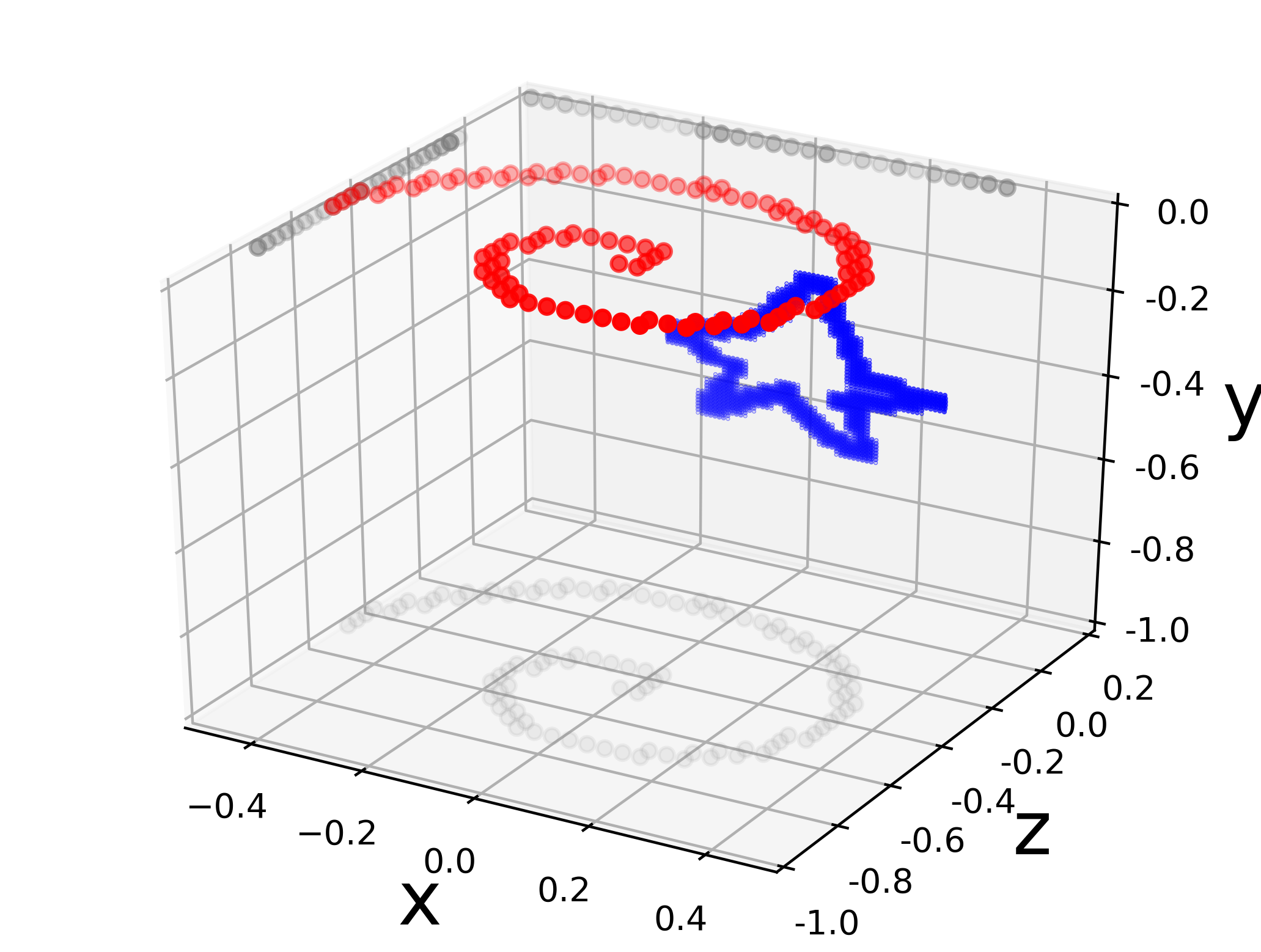}
			\put (10,80)  {Constant $y$ Plane}
		\end{overpic}
		\caption{}
	\end{subfigure}%
	
	\begin{subfigure}[t]{.15\textwidth}
		\centering
		\includegraphics[width=\textwidth]{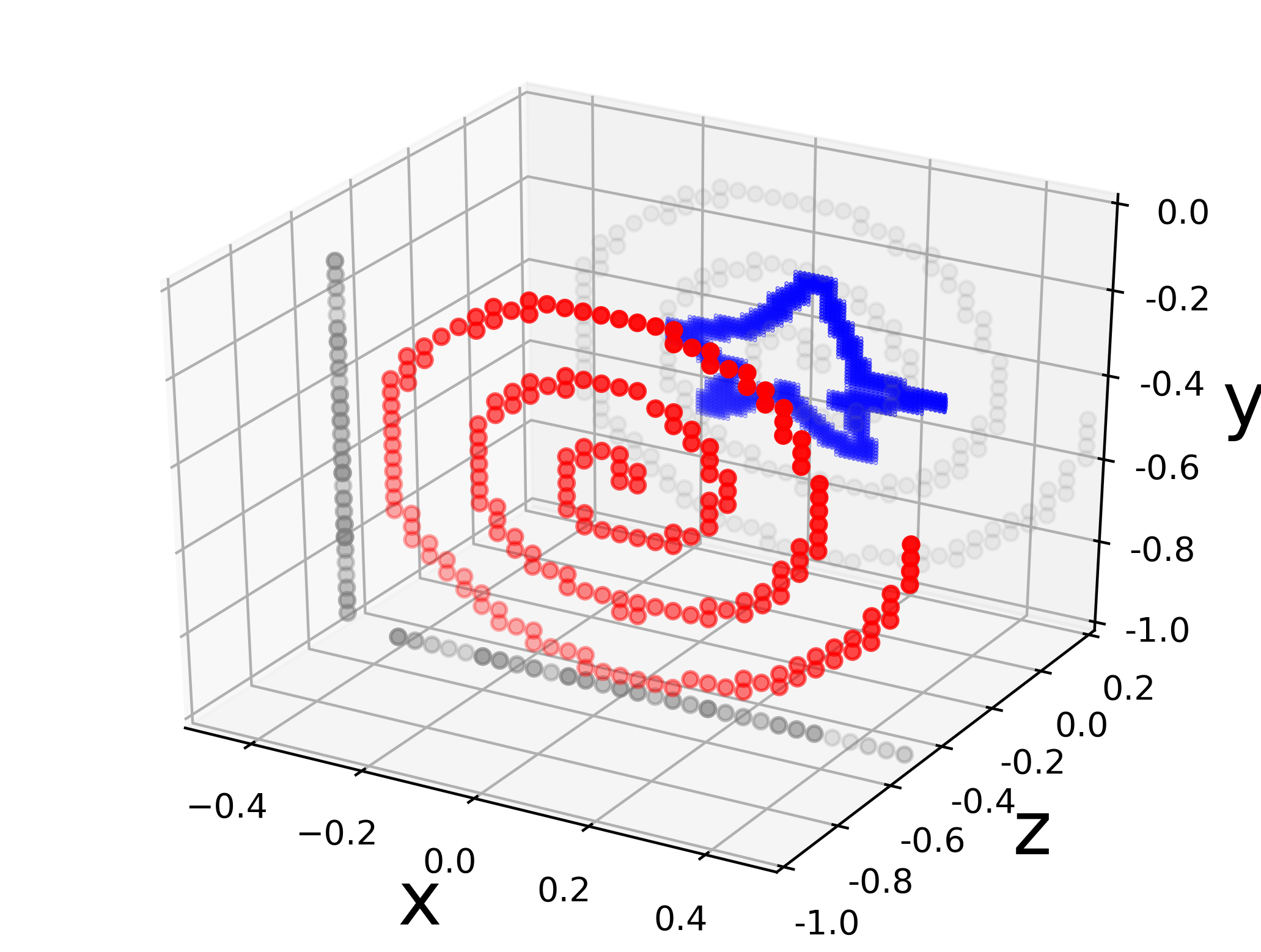}
		\caption{}
	\end{subfigure}%
	~
	\begin{subfigure}[t]{.15\textwidth}
		\centering
		\includegraphics[width=\textwidth]{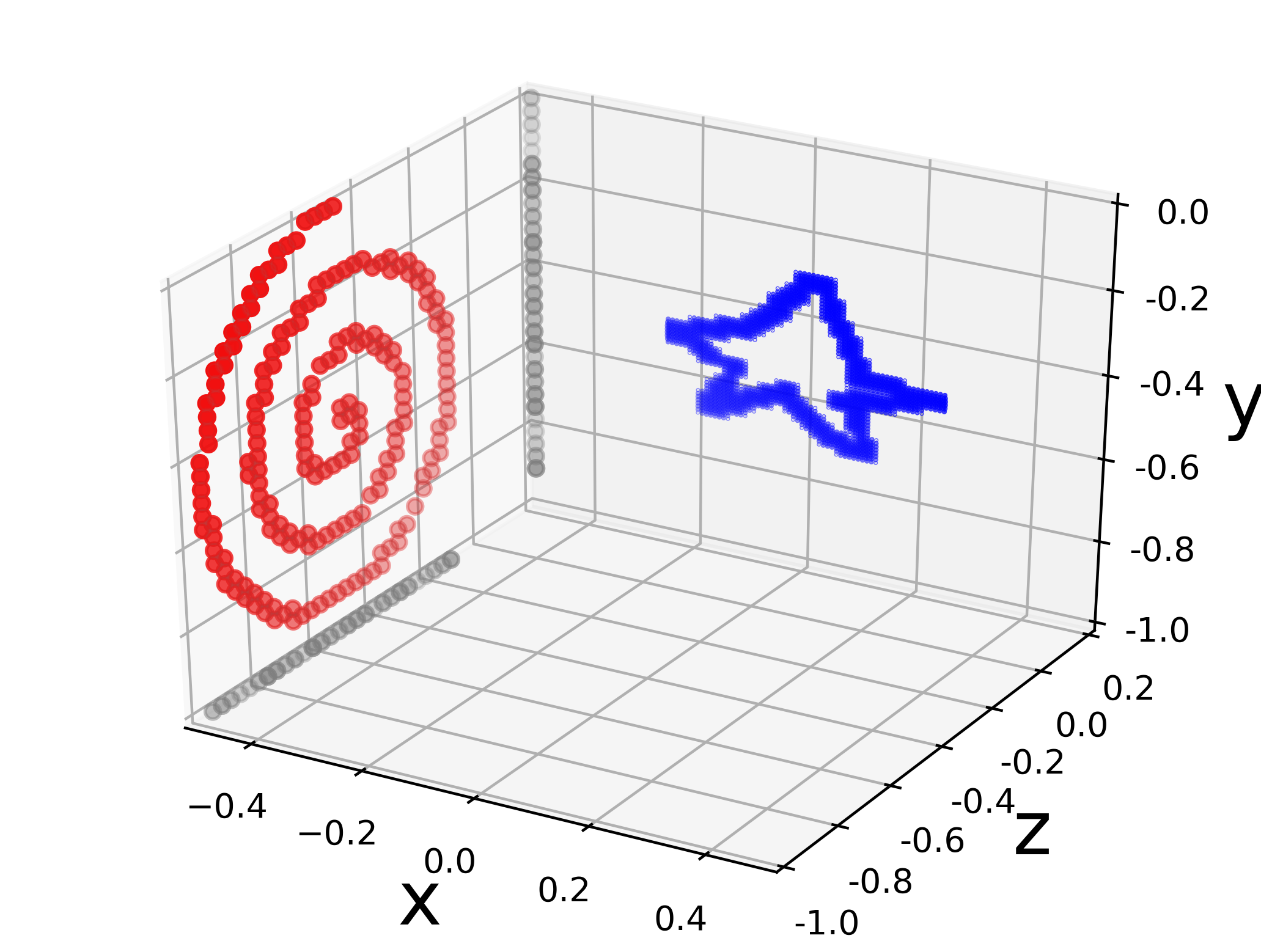}
		\caption{}
	\end{subfigure}%
	~
	\begin{subfigure}[t]{.15\textwidth}
		\centering
		\includegraphics[width=\textwidth]{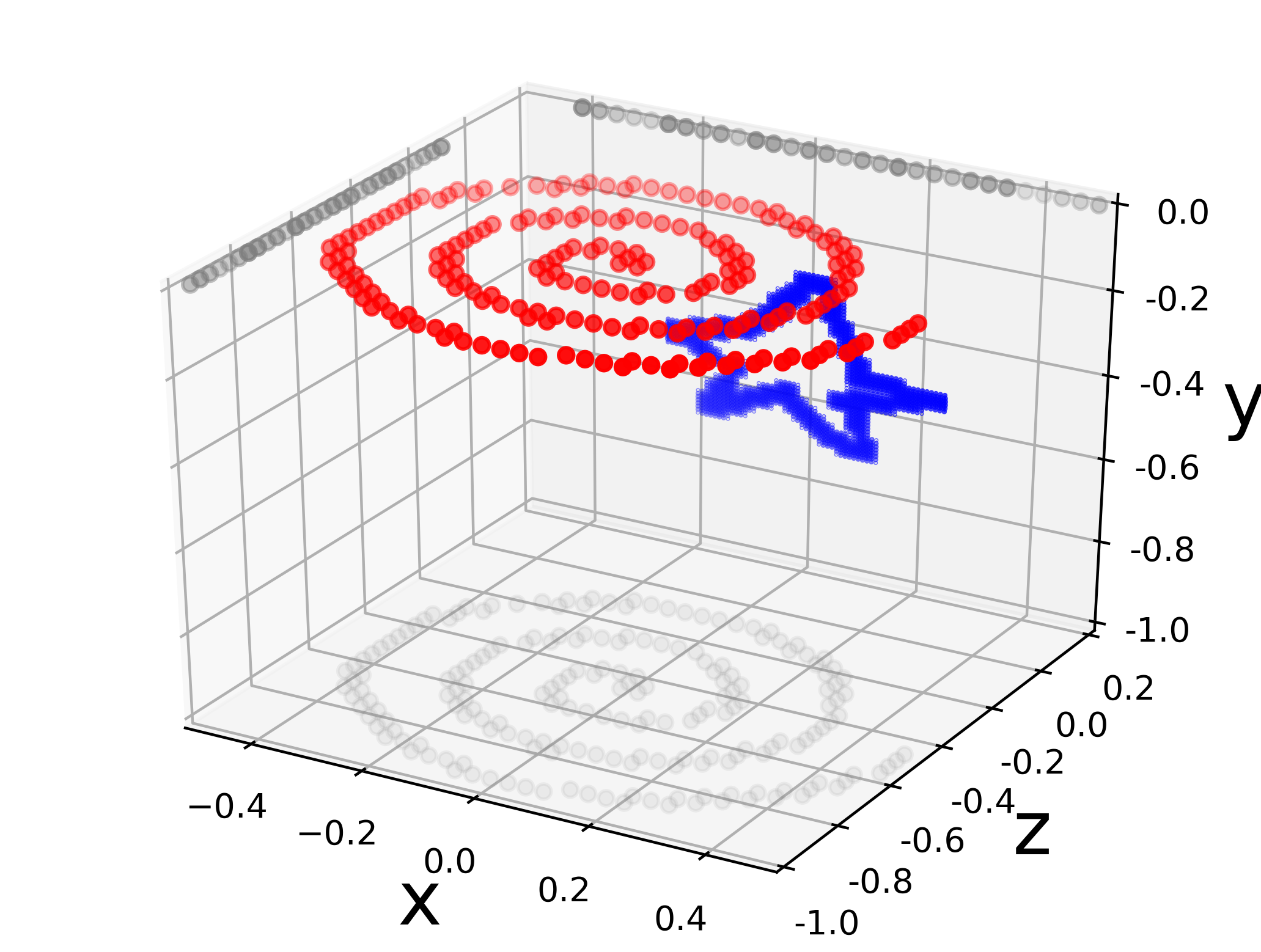}
		\caption{}
	\end{subfigure}%
	
	\begin{subfigure}[t]{.15\textwidth}
		\centering
		\includegraphics[width=\textwidth]{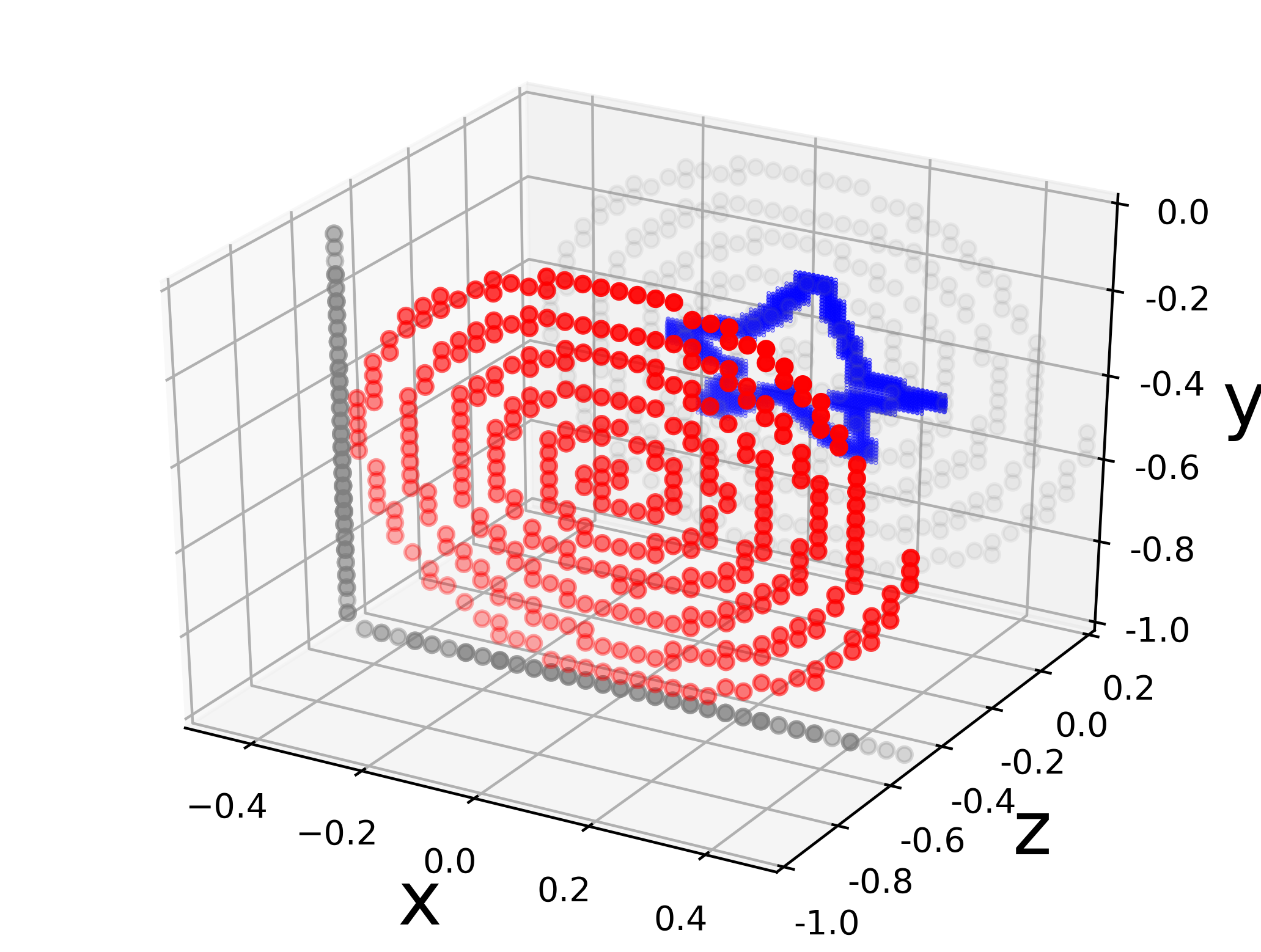}
		\caption{}
	\end{subfigure}%
	~
	\begin{subfigure}[t]{.15\textwidth}
		\centering
		\includegraphics[width=\textwidth]{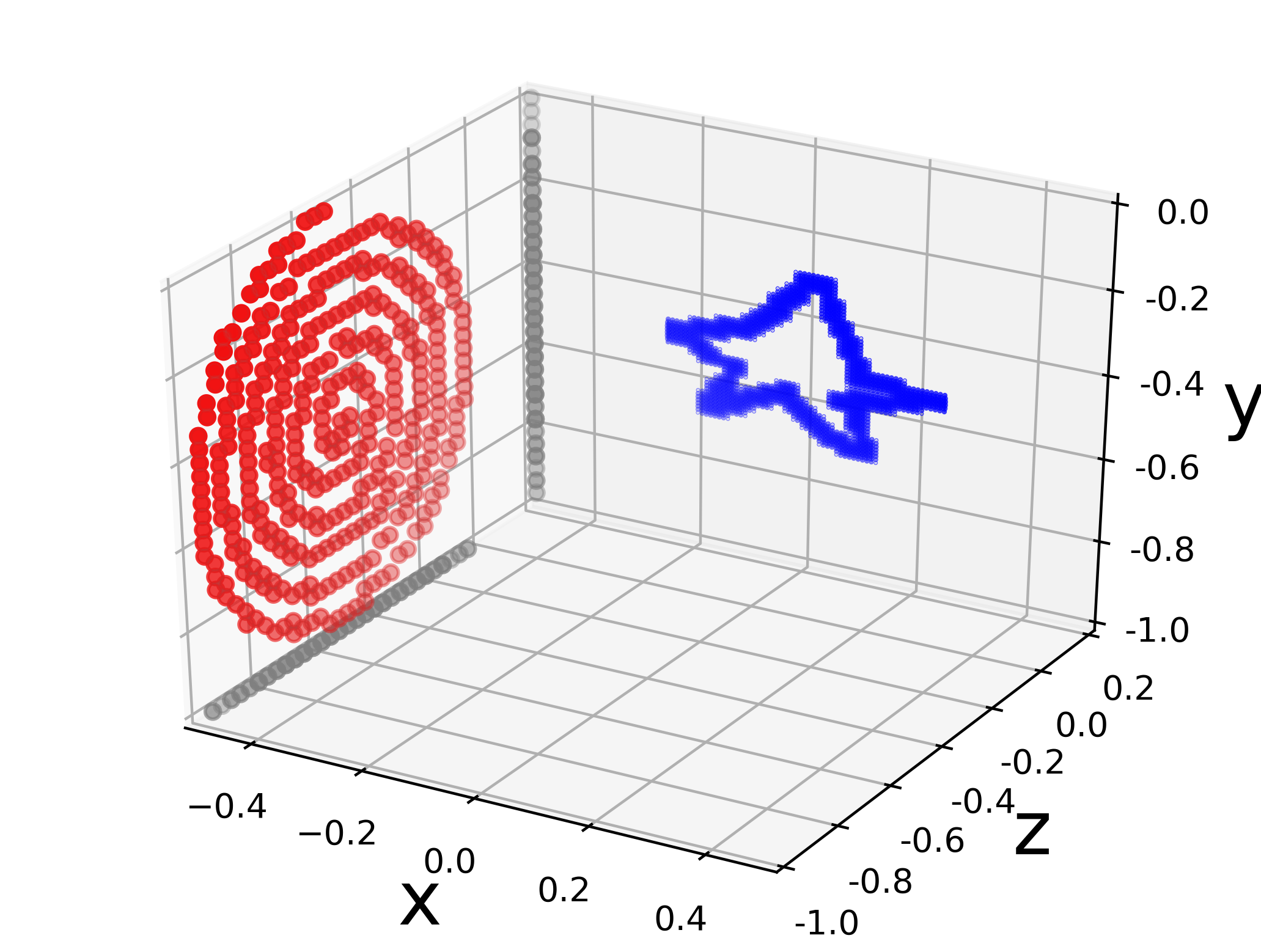}
		\caption{}
	\end{subfigure}%
	~
	\begin{subfigure}[t]{.15\textwidth}
		\centering
		\includegraphics[width=\textwidth]{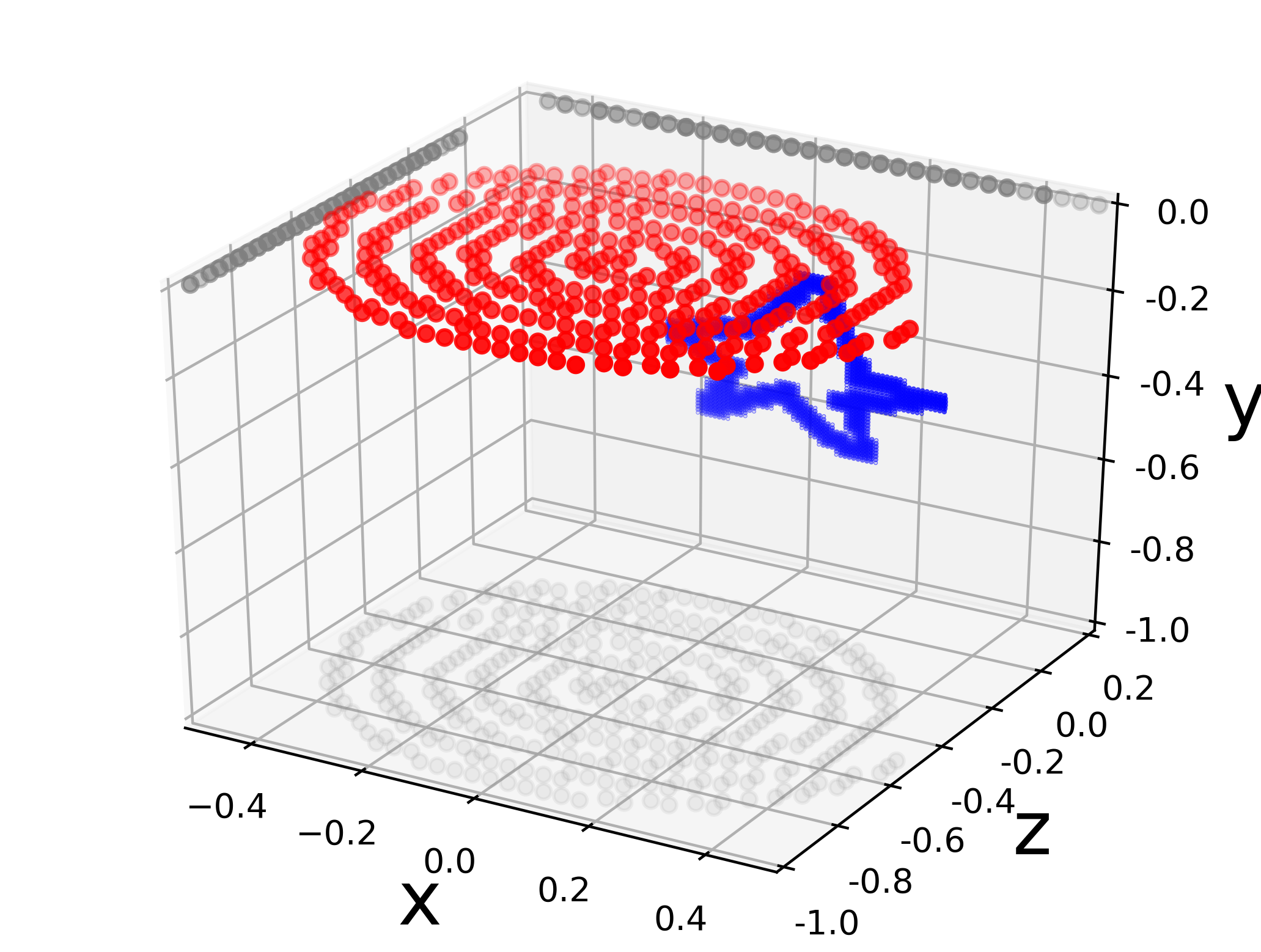}
		\caption{}
	\end{subfigure}%
	
	\begin{subfigure}[t]{.15\textwidth}
		\centering
		\includegraphics[width=\textwidth]{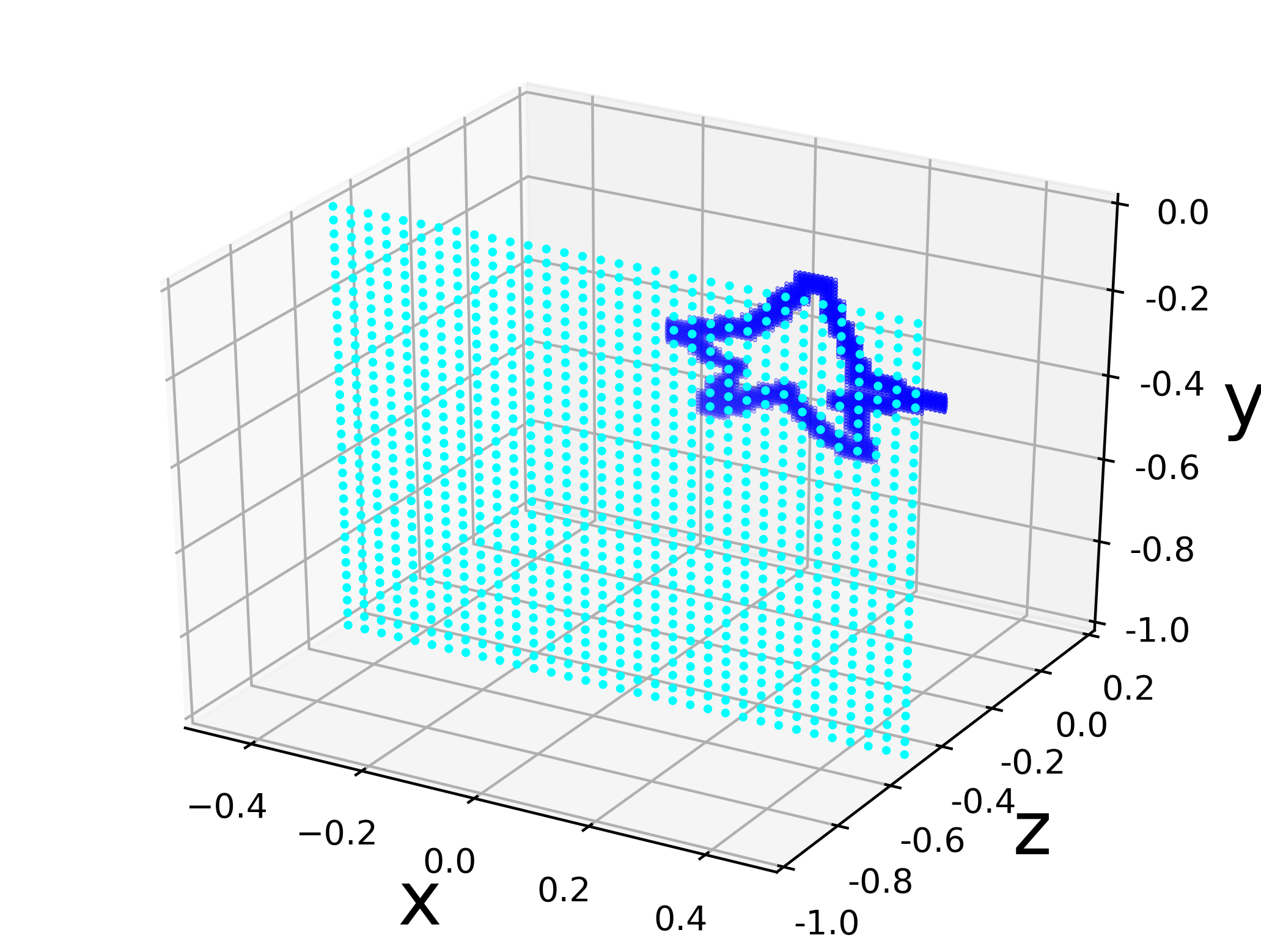}
		\caption{}
	\end{subfigure}%
	~
	\begin{subfigure}[t]{.15\textwidth}
		\centering
		\includegraphics[width=\textwidth]{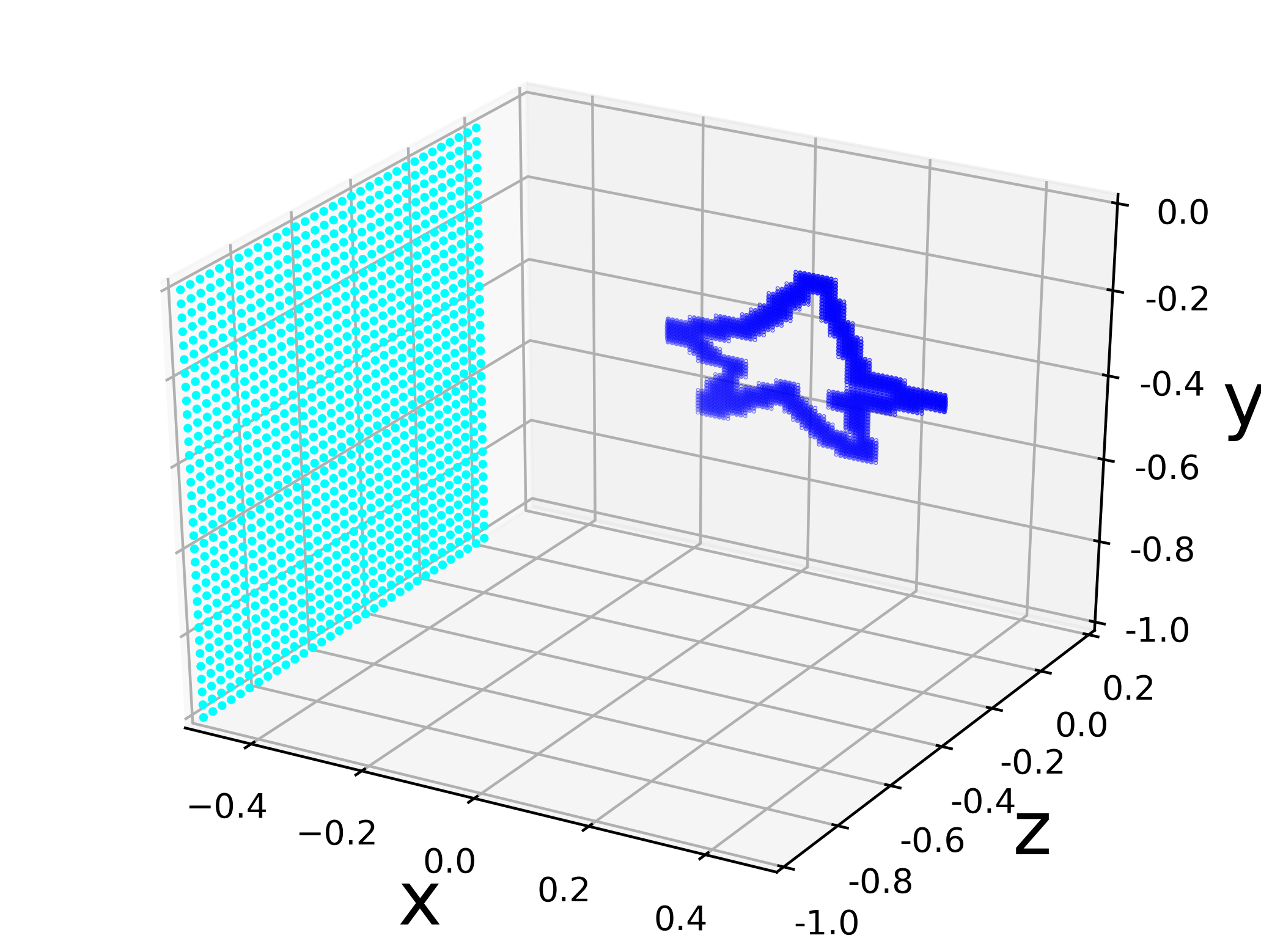}
		\caption{}
	\end{subfigure}%
	~
	\begin{subfigure}[t]{.15\textwidth}
		\centering
		\includegraphics[width=\textwidth]{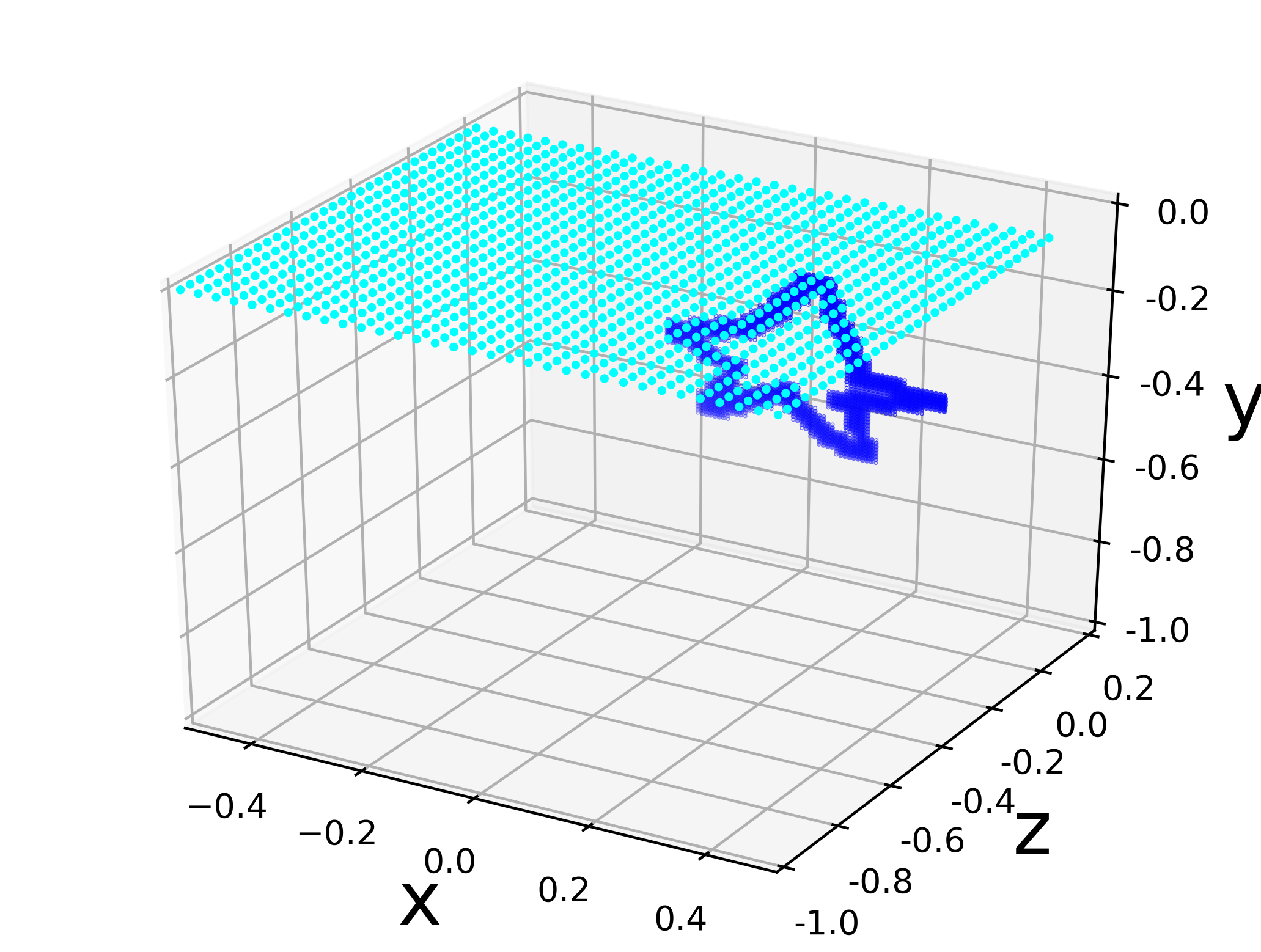}
		\caption{}
		\label{fig:2l}
	\end{subfigure}%
	
	\caption{\textbf{Simulation setup.} We simulated reconstructing objects which followed the nine sampling trajectories (a--i) shown in the top three rows. Each plot displays an example object in blue (the star, also show in Figure \ref{fig:SimResults}) and the locations of the virtual sensor locations, which are determined by the object trajectories, in red. The plots in the bottom row (j--l) show the allowable virtual sensor locations used by EM when reconstructing any of the three trajectories displayed above them.}
	\label{fig:SimSetup}
\end{figure}

\newcommand{\mywidth}{.09\textwidth}
\begin{figure*}
	\centering
	\hspace*{1.2cm}
	\begin{subfigure}[t]{\mywidth}
		\centering
			\begin{overpic}[width=\textwidth]{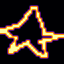}	
			\put (-70,45) {{\scriptsize \begin{tabular}{@{}c@{}} Ground \\ Truth \end{tabular}}}			
			\put (38,110) {{ 1}}					
		\end{overpic}	
	\end{subfigure}%
	~
	\begin{subfigure}[t]{\mywidth}
		\centering
		\begin{overpic}[width=\textwidth]{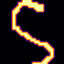}		
			\put (38,110) {{ 2}}
		\end{overpic}		
	\end{subfigure}%
	~
	\begin{subfigure}[t]{\mywidth}
		\centering
		\begin{overpic}[width=\textwidth]{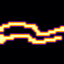}		
			\put (38,110) {{ 3}}
		\end{overpic}
	\end{subfigure}%
	~
	\begin{subfigure}[t]{\mywidth}
		\centering
		\begin{overpic}[width=\textwidth]{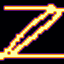}		
			\put (38,110) {{ 4}}
		\end{overpic}
	\end{subfigure}%
	~
	\begin{subfigure}[t]{\mywidth}
		\centering
		\begin{overpic}[width=\textwidth]{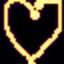}		
			\put (38,110) {{ 5}}
		\end{overpic}
	\end{subfigure}%
	~
	\begin{subfigure}[t]{\mywidth}
		\centering
		\begin{overpic}[width=\textwidth]{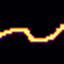}		
			\put (38,110) {{ 6}}
		\end{overpic}
	\end{subfigure}%
	~
	\begin{subfigure}[t]{\mywidth}
		\centering
		\begin{overpic}[width=\textwidth]{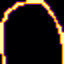}		
			\put (38,110) {{ 7}}
		\end{overpic}
	\end{subfigure}%
	~
	\begin{subfigure}[t]{\mywidth}
		\centering
		\begin{overpic}[width=\textwidth]{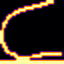}		
			\put (38,110) {{ 8}}
		\end{overpic}
	\end{subfigure}%
	~
	\begin{subfigure}[t]{\mywidth}
		\centering
		\begin{overpic}[width=\textwidth]{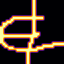}		
			\put (38,110) {{ 9}}
		\end{overpic}
	\end{subfigure}%
	
	\vspace{2 mm}
	\hspace*{1.2cm}
	\begin{subfigure}[t]{\mywidth}
		\centering
		\begin{overpic}[width=\textwidth]{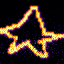}
			\put (37,-17)  {\scriptsize 0.73}
			\put (-105,45) {{\scriptsize \begin{tabular}{@{}c@{}c@{}} Known\\ Trajectory  \\Reconstructions\end{tabular}}}	
		\end{overpic}
	\end{subfigure}%
	~
	\begin{subfigure}[t]{\mywidth}
		\centering
		\begin{overpic}[width=\textwidth]{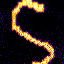}
			\put (37,-17)  {\scriptsize 0.60}
		\end{overpic}
	\end{subfigure}%
	~
	\begin{subfigure}[t]{\mywidth}
		\centering
		\begin{overpic}[width=\textwidth]{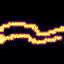}
			\put (37,-17)  {\scriptsize 0.77}
		\end{overpic}
	\end{subfigure}%
	~
	\begin{subfigure}[t]{\mywidth}
		\centering
		\begin{overpic}[width=\textwidth]{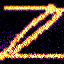}
			\put (37,-17)  {\scriptsize 0.73}
		\end{overpic}
	\end{subfigure}%
	~
	\begin{subfigure}[t]{\mywidth}
		\centering
		\begin{overpic}[width=\textwidth]{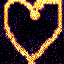}
			\put (37,-17)  {\scriptsize 0.66}
		\end{overpic}
	\end{subfigure}%
	~
	\begin{subfigure}[t]{\mywidth}
		\centering
		\begin{overpic}[width=\textwidth]{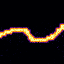}
			\put (37,-17)  {\scriptsize 0.78}
		\end{overpic}
	\end{subfigure}%
	~
	\begin{subfigure}[t]{\mywidth}
		\centering
		\begin{overpic}[width=\textwidth]{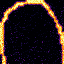}
			\put (37,-17)  {\scriptsize 0.63}
		\end{overpic}
	\end{subfigure}%
	~
	\begin{subfigure}[t]{\mywidth}
		\centering
		\begin{overpic}[width=\textwidth]{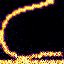}
			\put (37,-17)  {\scriptsize 0.69}
		\end{overpic}
	\end{subfigure}%
	~
	\begin{subfigure}[t]{\mywidth}
		\centering
		\begin{overpic}[width=\textwidth]{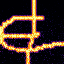}
			\put (37,-17)  {\scriptsize 0.71}
		\end{overpic}
	\end{subfigure}%
	
	\vspace{17 pt}
	\hspace*{1.2cm}
	\begin{subfigure}[t]{\mywidth}
		\centering
		\begin{overpic}[width=\textwidth]{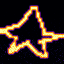}
			\put (37,-17)  {\scriptsize 0.81}
			\put (-105,45) {{\scriptsize \begin{tabular}{@{}c@{}c@{}} Unknown\\ Trajectory \\ Reconstructions  \end{tabular}}}	
		\end{overpic}
	\end{subfigure}%
	~
	\begin{subfigure}[t]{\mywidth}
		\centering
		\begin{overpic}[width=\textwidth]{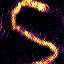}
			\put (37,-17)  {\scriptsize 0.59}
		\end{overpic}
	\end{subfigure}%
	~
	\begin{subfigure}[t]{\mywidth}
		\centering
		\begin{overpic}[width=\textwidth]{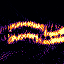}
			\put (37,-17)  {\scriptsize 0.60}
		\end{overpic}
	\end{subfigure}%
	~
	\begin{subfigure}[t]{\mywidth}
		\centering
		\begin{overpic}[width=\textwidth]{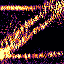}
			\put (37,-17)  {\scriptsize 0.55}
		\end{overpic}
	\end{subfigure}%
	~
	\begin{subfigure}[t]{\mywidth}
		\centering
		\begin{overpic}[width=\textwidth]{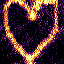}
			\put (37,-17)  {\scriptsize 0.53}
		\end{overpic}
	\end{subfigure}%
	~
	\begin{subfigure}[t]{\mywidth}
		\centering
		\begin{overpic}[width=\textwidth]{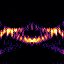}
			\put (37,-17)  {\scriptsize 0.61}
		\end{overpic}
	\end{subfigure}%
	~
	\begin{subfigure}[t]{\mywidth}
		\centering
		\begin{overpic}[width=\textwidth]{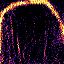}
			\put (37,-17)  {\scriptsize 0.58}
		\end{overpic}
	\end{subfigure}%
	~
	\begin{subfigure}[t]{\mywidth}
		\centering
		\begin{overpic}[width=\textwidth]{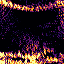}
			\put (37,-17)  {\scriptsize 0.58}
		\end{overpic}
	\end{subfigure}%
	~
	\begin{subfigure}[t]{\mywidth}
		\centering
		\begin{overpic}[width=\textwidth]{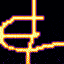}
			\put (37,-17)  {\scriptsize 0.79}
		\end{overpic}
	\end{subfigure}%
	\vspace{12 pt}
	
	\caption{\textbf{Simulation results.} Top: The nine test objects. Middle: Known-trajectory reconstructions using gradient descent with simulated measurements that follow trajectory (f) from Figure \ref{fig:SimSetup} and have an SNR of 15. Bottom: Unknown-trajectory reconstructions using EM with the same simulated measurements. Below each reconstruction we report the disambiguated SSIM.}
	\label{fig:SimResults}
\end{figure*}


We first investigate the keyhole imaging problem in simulation. 
The keyhole measurements are simulated from nine different binary objects, drawn from the HaSyV2 dataset \cite{thoma2017hasyv2}, which are illustrated in the top row of Figure \ref{fig:SimResults}. Each of these objects has a resolution of $64\times 64$ and is $50$ cm tall and wide in the simulator. Our simulated SPAD measurements have a temporal resolution of 16 ps. We apply Poisson noise to the measurements such that they have the desired signal-to-noise ratio (SNR) for each test.

For each of the nine objects, we simulate measurements from nine distinct trajectories of varying lengths.  The virtual sensor locations corresponding to each of these trajectories is shown in Figure \ref{fig:SimSetup}. Recall measurements of a moving object with a fixed sensor location are equivalent to measurements of a fixed object with a moving virtual sensor. 
Three of these trajectories (left column) have the virtual sensor locations restricted to a constant $z$ plane; this corresponds to object roll and translation up-and-down and side-to-side.
Another three of these trajectories (middle column) have the virtual sensor locations restricted to a constant $x$ plane; this corresponds to object pitch and translation up-and-down and forward-and-backward.
The last three of these trajectories (right column) have the virtual sensor locations restricted to a constant $y$ plane; this corresponds to object yaw and translation side-to-side and forward-and-backward.
The horizontal motion in the last of these trajectories is arguably the most realistic; for instance, this is the type of motion exhibited by cars, rolling chairs, and shopping carts. 
The last row of Figure \ref{fig:SimSetup} presents the $33\times 33$ grids $\Omega$ of allowable virtual sensor locations that we use with EM.

Because the EM algorithm does not know the object trajectories beforehand, the algorithm's reconstructions have certain ambiguities/invariances. 
Constant $z$ trajectories are rotation invariant: given a reconstructed trajectory and an object that fits the measurements, one could rotate both clockwise without changing the fit.
Similarly, constant $x$ trajectories are invariant to vertical flips of the trajectory and object and constant $y$ trajectories, which are what we test with real data in the next section, are invariant to horizontal flips.

When considered as object/trajectory pairs, none of the reconstructions are invariant to translations. 
However, an object/trajectory pair can be expressed as an equivalent pair where the object is translated one direction and the trajectory is translated equally in the opposite direction. Thus, in comparing our reconstructed objects to the ground truth, we need to compare across translations as well.

\subsection{Simulation Results}\label{ssec:SimResults}
In order to form a performance baseline, we first reconstruct the hidden objects assuming their trajectories are known. 
This is accomplished by maximizing the log-likelihood
\begin{flalign}\label{eqn:KnownLocation}
&{\Big (}\sum_{l=1}^L -\|\mathbf{y}_l-f(\albedo,\mathbf{\trans}_l)\|^2_2{\Big )} + \lambda \log p(\albedo),
\end{flalign}
where the objects' locations over time, $\mathbf{\trans}_1,...\mathbf{\trans}_L$, are given. 
We use 200 iterations of gradient descent (GD), along-side the ADAM optimizer, to maximize \eqref{eqn:KnownLocation}. \edit{As in the EM case, the non-negative albedo $\rho$ was parameterized as $\rho=\nu^2$ and the variable $\nu$ was initialized with an i.i.d.~Gaussian vector with mean 0 and variance 1.}

We compare EM, which does not know the trajectories, and GD, which does, qualitatively in Figure~\ref{fig:SimResults} and quantitatively in Table \ref{tab:SimResults}. 
Table \ref{tab:SimResults} reports reconstruction accuracy in terms of disambiguated Structural Similarity (SSIM)~\cite{ssim}, which we define as
\begin{align}\label{eqn:SSIMdef}
\max_{\text{rtf} \in \text{ RTF}}\text{SSIM}(\albedo,\text{rtf}(\hat{\albedo})),
\end{align}
where $\hat{\albedo}$ denotes the reconstructed albedo and $\text{RTF}$ denotes the set of all possible rotations, translations, and flips of the object. We test over all rotations in 5\degree~increments and all translations in 1 pixel increments.


Figure~\ref{fig:SimResults} compares reconstructions between GD and EM when the object follows trajectory (f) from Figure \ref{fig:SimSetup} and the measurements have an SNR of 15. The results show EM performs nearly as well as GD. 
\edit{However, because there is a flip ambiguity, that is an object moving left would produce the same measurements as a flipped version of an object moving right, EM occasionally finds two mirrored objects/trajectories and it struggles to decide between the two. Scenarios like these (which are more common at lower resolutions) result in artifacts like those demonstrated for objects 6 and 8 in Figure~\ref{fig:SimResults}, where we can see the traces of two flipped versions of the object overlaid on top of one another.} 
Table~\ref{tab:SimResults} demonstrates EM performs only a little worse on average than GD across a variety of SNRs and trajectories. 
\edit{Additional simulation results, including rotation-only trajectories and 3D trajectories, are provided in Appendix \ref{sec:ExtraSim}.}


\newcommand\mc[1]{\multicolumn{1}{l}{#1}} 

\rowcolors{2}{white}{gray!25}
\begin{table}[t!]%
	\centering
	{\small
		\begin{tabular}{l|cccccc}
			\rowcolor{white}
			\toprule
			\mc{}& \multicolumn{2}{c}{SNR = 5}& \multicolumn{2}{c}{SNR = 15}& \multicolumn{2}{c}{SNR = 50} \\ 
			\cmidrule(lr){2-3}\cmidrule(lr){4-5}\cmidrule(lr){6-7}
			\rowcolor{white}
			\mc{}&  GD & EM & GD&EM & GD&EM \\
			\midrule
			Trajectory (a) & \textbf{0.56}&0.55 & \textbf{0.66}&0.57 &\textbf{0.75} &0.58\\
			Trajectory (b) & \textbf{0.55}&0.49 & \textbf{0.64}&0.50 &\textbf{0.72} &0.51\\
			Trajectory (c) & \textbf{0.59}&0.57 & \textbf{0.69}&0.57 &\textbf{0.77} &0.59\\
			Trajectory (d) & \textbf{0.59}&0.55 & \textbf{0.68}&0.58 &\textbf{0.77} &0.57\\
			Trajectory (e) & \textbf{0.58}&0.48 & \textbf{0.66}&0.48 &\textbf{0.73} &0.50\\
			Trajectory (f) & \textbf{0.62}&0.60 & \textbf{0.70}&0.63 &\textbf{0.79} &0.64\\
			Trajectory (g) & \textbf{0.61}&0.58 & \textbf{0.70}&0.58 &\textbf{0.80} &0.59\\
			Trajectory (h) & \textbf{0.60}&0.51 & \textbf{0.68}&0.52 &\textbf{0.76} &0.53\\
			Trajectory (i) & \textbf{0.65}&0.61 & \textbf{0.72}&0.63 &\textbf{0.81} &0.65\\
			\bottomrule
		\end{tabular}
	}
	\caption{Comparison of the mean disambiguated SSIM (higher is better) across the 9 tests images with various trajectories and SNRs. EM works best with long trajectories, but is relatively robust to noise.}
	\label{tab:SimResults}
\end{table}%
\rowcolors{2}{}{}

\Section{Experimental Validation}
\label{sec:results}

\subsection{Experimental Setup}

\begin{figure}
	\centering
	\includegraphics[width=\columnwidth]{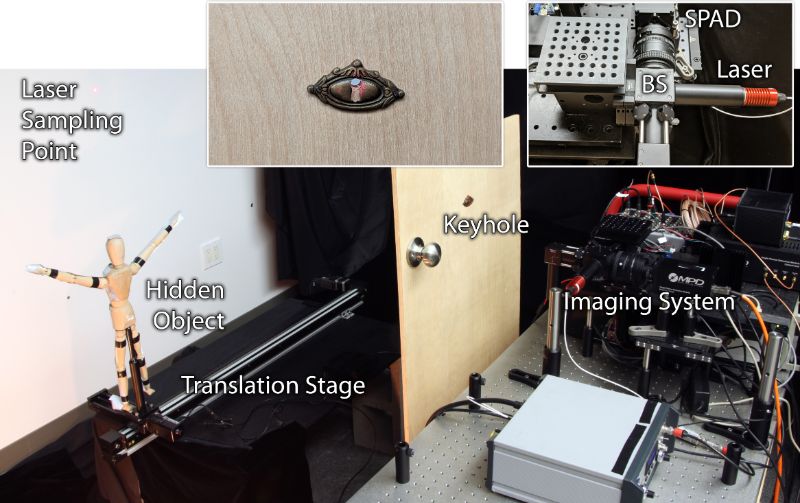}
	\caption{\textbf{Experimental setup.} Our optical system sends a laser pulse through the keyhole of a door. On the other side of the door, the hidden object moves along a translation stage. When third-bounce photons return, they are recorded and time-stamped by a SPAD. Top-right inset: A beam splitter (BS) is used to place the laser and SPAD in a confocal configuration.}
	\label{fig:Exp_setup}
\end{figure}

Our prototype system is illustrated in Figure \ref{fig:Exp_setup}. 
The optical setup consists of a \SI{670}{\nano\meter} pulsed laser source (ALPHALAS PICOPOWER-LD-670-50) operating with a \SI{10}{\mega\hertz} repetition rate, an average power of approximately \SI{0.1}{\milli\watt}, and a pulse width of \SI{30}{\pico\second}. 
This laser is in a confocal configuration with a fast-gated single-pixel SPAD detector (Micro Photon Devices PDM series SPAD, \SI{50}{\micro\meter} $\times$ \SI{50}{\micro\meter} active area), which allows us to gate out direct bounce photons, capturing only indirect photons from the hidden object.
A time-correlated single photon counter (PicoQuant PicoHarp 300) takes as input a trigger signal from the laser and photon detection event triggers from the SPAD and forms time-stamped histograms of photon arrival times with \SI{16}{\pico\second} bin widths.

We captured keyhole measurements of the objects  illustrated in the top row of Figure \ref{fig:ExpResults}. The objects are each 50~cm tall and covered in retroreflective tape. The objects were affixed to two Zaber translation stages: a \SI{1}{\meter} linear stage (X-BLQ1045-E01) and a \SI{15}{\centi\meter} stage (T-LSR150A).  Using the stages, objects can be translated to any point within a \SI{1}{\meter} by \SI{15}{\centi\meter} horizontal plane, where the long axis ($x$-axis) is aligned parallel to the wall and the short axis ($z$-axis) is aligned perpendicular to the wall \edit{such that at their closest point the objects are \SI{.62}{\centi\meter} from the wall}. In the captured measurements, objects are translated to between 66 and 165 discrete locations. To capture adequate signal given the limited laser power of our prototype, the objects hold position at each location for 10 seconds. Example trajectories are shown in the last column of Figure \ref{fig:ExpResults}.

\subsection{Experimental Results}\label{ssec:ExpResults}
\begin{figure}[t]
	\centering
	\begin{subfigure}[t]{.105\textwidth}
		\centering
		\begin{overpic}[angle=0,origin=c,width=\textwidth]{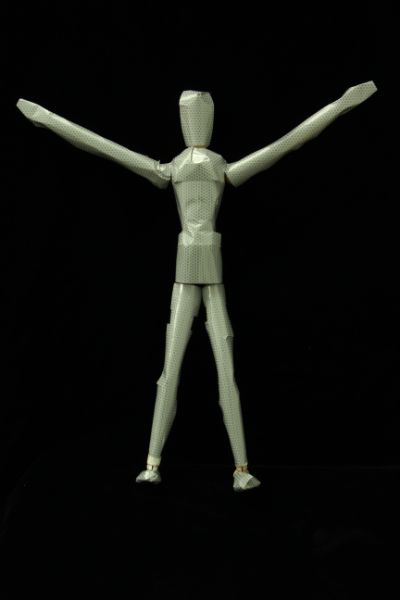}
		\end{overpic}
	\end{subfigure}%
	~ 
	\begin{subfigure}[t]{.105\textwidth}
		\centering
		\begin{overpic}[angle=0,origin=c,width=\textwidth]{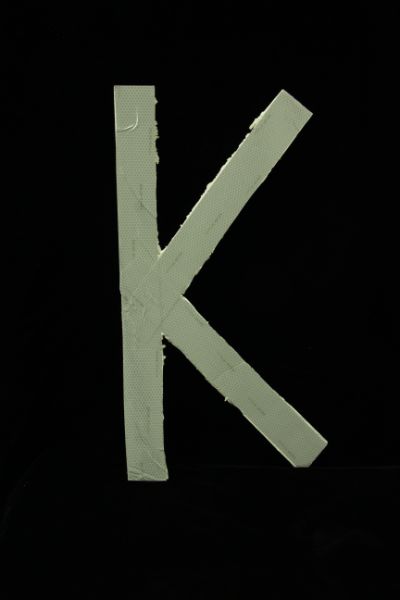}
		\end{overpic}
	\end{subfigure}%
	~ 
	\begin{subfigure}[t]{.105\textwidth}
		\centering
		\begin{overpic}[angle=0,origin=c,width=\textwidth]{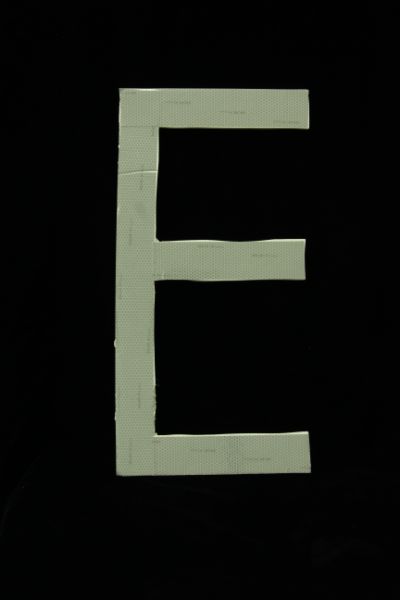}
		\end{overpic}
	\end{subfigure}%
	~ 
	\begin{subfigure}[t]{.105\textwidth}
		\centering
		\begin{overpic}[angle=0,origin=c,width=\textwidth]{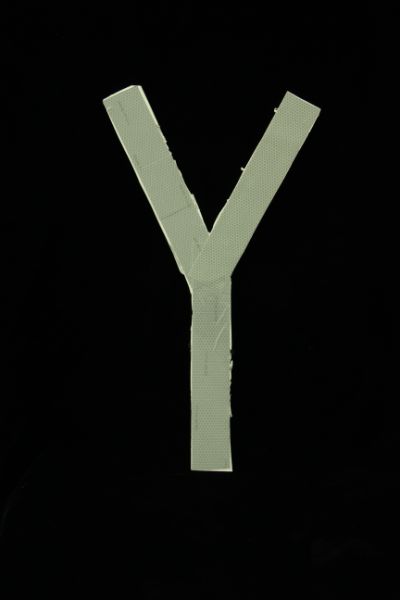}
		\end{overpic}
	\end{subfigure}%
	
	\vspace{1 mm}
	\begin{subfigure}[t]{.105\textwidth}
		\centering
		\begin{overpic}[width=\textwidth]{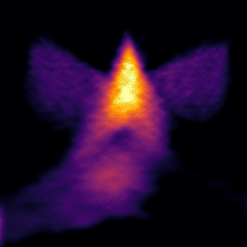}
			\put (3,25) {\textcolor{white}{\scriptsize $\uparrow$}}
			\put (2,12) {\textcolor{white}{\scriptsize $y$}}
			\put (12,2) {\textcolor{white}{\scriptsize $x \rightarrow$}}
		\end{overpic}
	\end{subfigure}%
	~ 
	\begin{subfigure}[t]{.105\textwidth}
		\centering
		\scalebox{-1}[1]{\includegraphics[width=\textwidth]{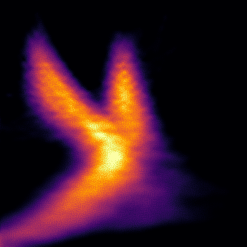}}
	\end{subfigure}%
	~ 
	\begin{subfigure}[t]{.105\textwidth}
		\centering
		\scalebox{-1}[1]{\includegraphics[width=\textwidth]{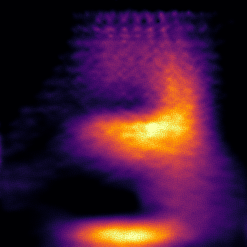}}
	\end{subfigure}%
	~ 
	\begin{subfigure}[t]{.105\textwidth}
		\centering
		\includegraphics[width=\textwidth]{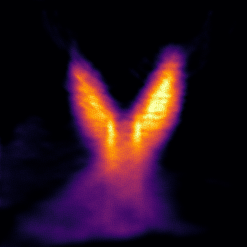}
	\end{subfigure}%
	
	\vspace{1 mm}
	\begin{subfigure}[t]{.105\textwidth}
		\centering
		\begin{overpic}[width=\textwidth]{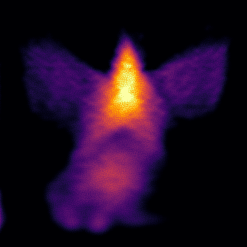}
			\put (3,25) {\textcolor{white}{\scriptsize $\uparrow$}}
			\put (2,12) {\textcolor{white}{\scriptsize $y$}}
			\put (12,2) {\textcolor{white}{\scriptsize $x \rightarrow$}}
		\end{overpic}
		\vspace*{-20pt} 
	\end{subfigure}%
	~ 
	\begin{subfigure}[t]{.105\textwidth}
		\centering
		\scalebox{1}[1]{\includegraphics[width=\textwidth]{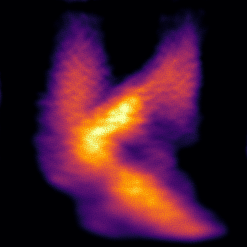}}
	\end{subfigure}%
	~ 
	\begin{subfigure}[t]{.105\textwidth}
		\centering
		\scalebox{-1}[1]{\includegraphics[width=\textwidth]{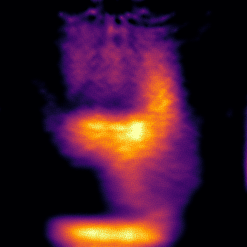}}
	\end{subfigure}%
	~ 
	\begin{subfigure}[t]{.105\textwidth}
		\centering
		\includegraphics[width=\textwidth]{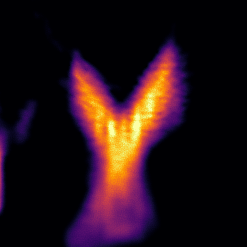}
	\end{subfigure}%
	
	\vspace{1 mm}
	\begin{subfigure}[t]{.105\textwidth}
		\centering
		\begin{overpic}[width=\textwidth]{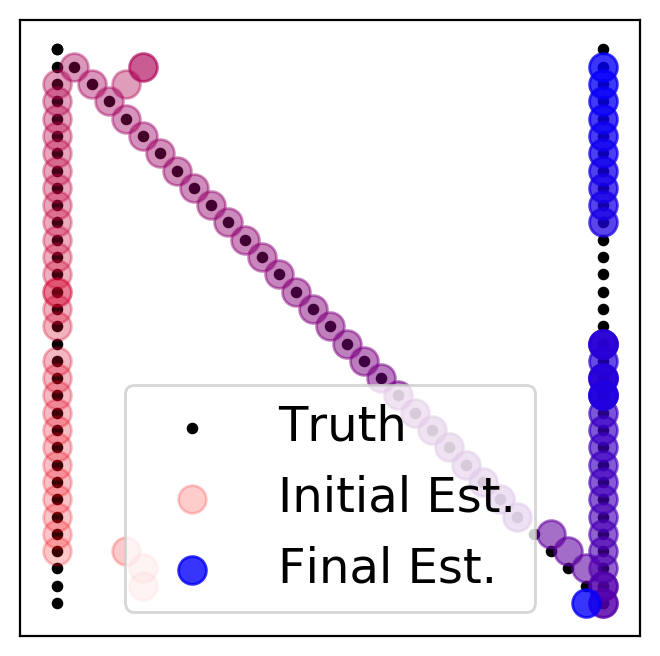}
			\put (-7,25) {{\scriptsize $\uparrow$}}
			\put (-7,12) {{\scriptsize $x$}}				
			\put (12,-5) {{\scriptsize $z \rightarrow$}}
		\end{overpic}	
	\end{subfigure}%
	~ 
	\begin{subfigure}[t]{.105\textwidth}
		\centering
		\scalebox{1}[1]{\includegraphics[width=\textwidth]{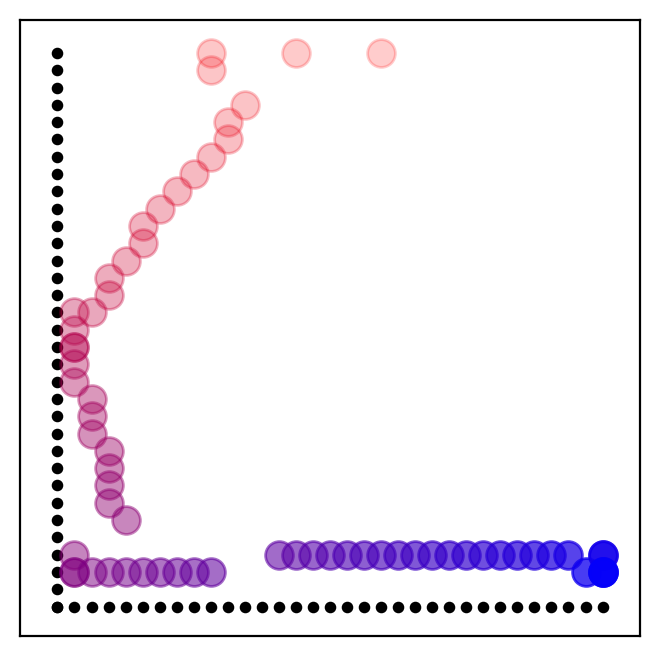}}
	\end{subfigure}%
	~ 
	\begin{subfigure}[t]{.105\textwidth}
		\centering
		\scalebox{1}[1]{\includegraphics[width=\textwidth]{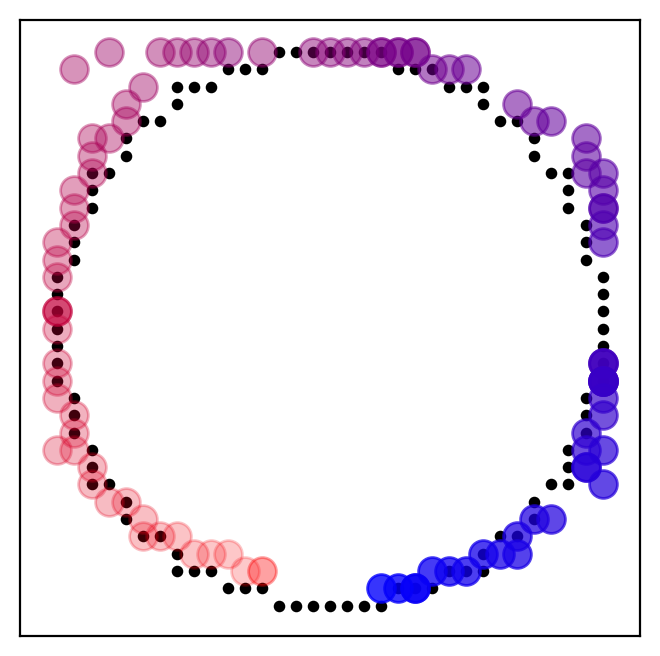}}
	\end{subfigure}%
	~ 
	\begin{subfigure}[t]{.105\textwidth}
		\centering
		\includegraphics[width=\textwidth]{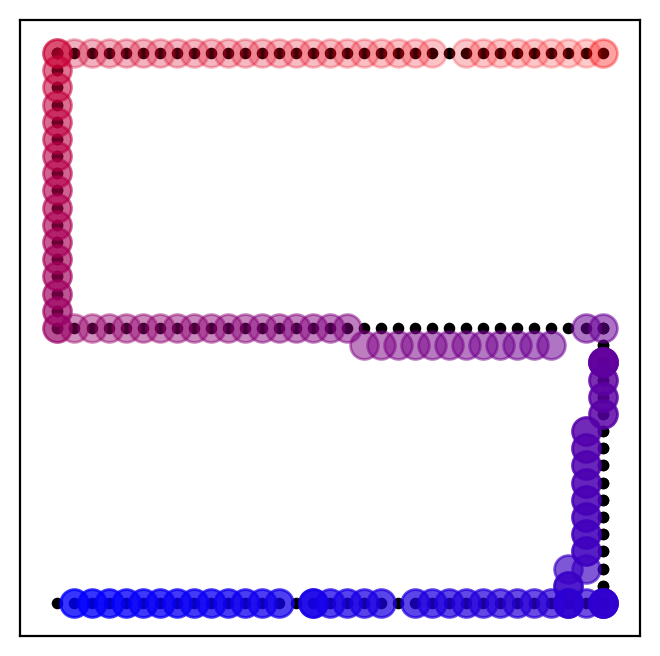}
	\end{subfigure}%
	
	\caption{\textbf{Experimental results.} First row: Images of the hidden objects. Second row: Reconstructions of the hidden objects using GD when their trajectories are known. Third row: EM reconstructions of the hidden objects when their trajectories are unknown. Fourth row: EM estimates of the trajectories of the hidden objects, each of which follows a different trajectory, where the dot color indicates position over time.} 
	\label{fig:ExpResults}
\end{figure}


As in the previous section, in order to form a performance baseline, we first reconstruct the hidden objects using GD assuming their trajectories were known. 
The resulting reconstructions are presented in the second row of Figure \ref{fig:ExpResults}. 
The reconstructions illustrate that ($x$,$z$) translations contain the measurement diversity necessary to perform NLOS reconstructions with real data. The mannequin's pose is recognizable and the letters are all readable. \edit{The mannequin's reconstruction is likely blurrier than the letters' because the mannequin is 3D while the model fit to the data is 2D.}

For the case where object trajectories are not known, we use EM to recover both the shape and trajectory of the hidden object. For each measurement, EM also produces an estimate of the probability distribution of each object's location. We select the highest probability location as the location estimate. {Note that we found that { by the time our algorithm has converged}, the posterior probability estimates of the object locations are essentially one hot. That is, for each instance in time $l$, $w_{\theta_l}$ assigns a probability near one that the object is at a certain position and a probability near 0 that it is at any other position. \editTwo{(During the early iterations of the algorithm, when it does not have an accurate estimate for the shape of the object, the assigned probabilities are nearly uniform.)}

The reconstructed objects and trajectories are presented in the third and fourth rows of Figure \ref{fig:ExpResults}. Again the mannequin's pose is recognizable and the letters are all readable. Moreover, EM's estimated trajectories closely match the ground truth and in all four examples the general path of the object is clearly visible. 

The EM reconstruction of the ``K'' serves to illustrate the translation ambiguity that was discussed in the previous section. Relative to the true trajectory, the reconstructed trajectory is displaced by a positive shift in the $x$ direction. Meanwhile, the object reconstruction is offset by a negative shift in the $x$ direction. Considered jointly, the object's location is recovered accurately. That is, when shifted to the right ($+x$) to undo the error in its trajectory, the EM ``K'' in row three of Figure \ref{fig:ExpResults} overlaps with the GD ``K'' in row two.

The EM reconstructions were computed at a $256\times256$ resolution in five and a half minutes using an Nvidia Titan RTX GPU and a six core Intel CPU. The known-trajectory GD reconstructions took fifteen seconds on the same hardware. 

\editTwo{Code and data to reproduce these results is available at \url{https://github.com/computational-imaging/KeyholeImaging}}.

\Section{Discussion}
\label{sec:discussion}
This work proposes, develops, and experimentally validates keyhole imaging---a new technique to reconstruct the shape and location of a hidden object from NLOS measurement captured along a single optical path. 
Whereas in standard NLOS imaging one scans multiple points across a relay surface to form a synthetic aperture, in keyhole imaging one scans a single point and the trajectory of the object determines the size and shape of the synthetic aperture. This distinction has important implications.

\paragraph{Resolution Limits}
Keyhole imaging performance is upper-bounded by the performance of conventional NLOS system with an equivalent synthetic aperture. 
Our simulation results from the previous section indicate this bound is fairly tight. 
The larger distinction between keyhole and conventional NLOS is that (with planar, horizontal object motion) keyhole's synthetic aperture is a horizontal surface rather than vertical wall. This change results in significantly different resolution limits. 

\begin{figure}[h]
	\centering
	\includegraphics[width=\linewidth]{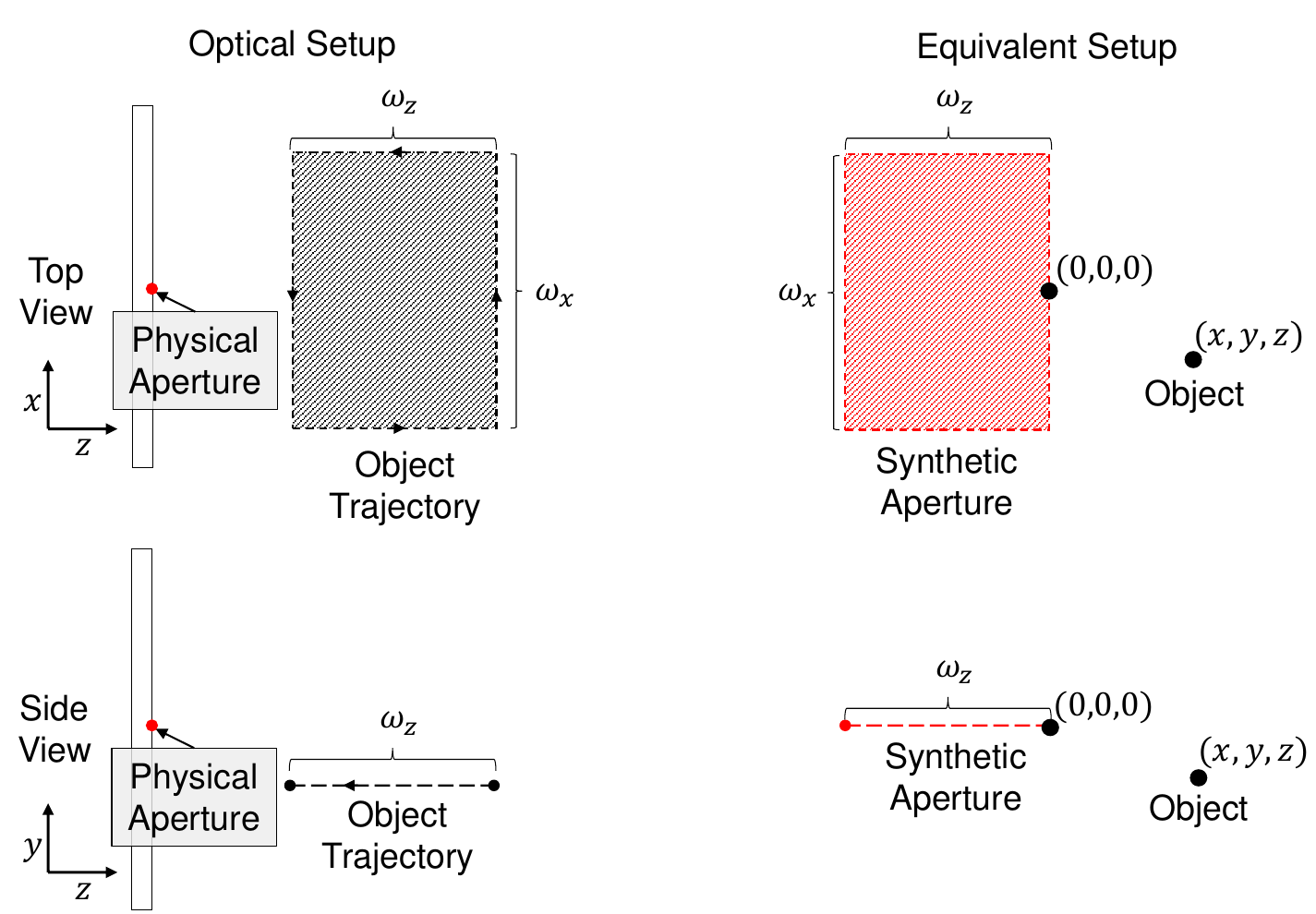}
	\caption{{\bf Synthetic apertures.} The horizontal object motion forms a horizontal synthetic aperture. This aperture has different resolution limits than the vertical synthetic apertures used in standard NLOS imaging.}
	\label{fig:ResLimits1}
\end{figure}

In the appendix, following the analysis conducted in~\cite{OToole:2018}, we derive resolution limits associated with the $\omega_x\times\omega_z$ horizontal synthetic aperture illustrated in Figure~\ref{fig:ResLimits1}. In particular, we assume that the detector has Gaussian jitter with standard deviation $\sigma_{SPAD}$ and derive lower bounds on the resolutions in the $x$, $y$, and $z$ directions, defined simply as the minimum spacing at which two nearby points would produce distinguishable measurements. Our analysis shows that 
\begin{align}
\Delta x &\geq\frac{c\gamma}{2}\sqrt{\frac{y^2+z^2}{(|x|+\omega_x/2)^2}+1},\\
\Delta y 
&\geq\frac{c\gamma}{2}\sqrt{\frac{z^2}{y^2}+1},\label{eqn:deltaY}\\
\Delta z &\geq\frac{c\gamma}{2}\sqrt{\frac{y^2}{(\omega_z+z)^2}+1},
\end{align}
where $c$ is the speed of light and $\gamma=2\sqrt{2\ln 2 }\sigma_{SPAD}$ is the full-width at half maximum of the temporal jitter. The $x$, $y$, and $z$ coordinates are given assuming the origin is at the center of the synthetic aperture with respect to the $x$ direction and on the edge nearest the object with respect to the $z$ direction. 
See Figures~\ref{fig:ResLimits1} and~\ref{fig:SimSetup}(l) for reference.

These results imply that the resolution in any dimension is lower bounded by $\frac{c\gamma}{2}$ and that the resolution in the $x$ and $z$ dimensions improves as the synthetic aperture gets larger with respect to $x$ and $z$ respectively. That is, the  resolution in $x$ and $z$ improves as the object translates more in the $x$ and $z$ directions. The above results also imply that near the height of sampling point on the wall, that is as $y$ approaches $0$, the system has very limited resolution in the vertical ($\Delta y$) direction; the bound described by \eqref{eqn:deltaY} approaches infinity.

\paragraph{Scanning Through a Small Aperture} 
\edit{Imaging through a small aperture/keyhole causes a reduction in light throughput, as compared to standard confocal NLOS imaging. The size of this reduction is determined by the size of the keyhole as well as the distances between the detector, the keyhole, and the relay surface.} In a configuration like ours, the keyhole is significantly larger than both the laser beam width and the SPAD's focus size on the wall. In this context, the keyhole only serves to restrict the maximum usable aperture size of the detector. 
If the detector is $d_1$ away from the keyhole and the keyhole is $d_2$ away from the wall then a $\Gamma$ diameter circular keyhole restricts the usable aperture diameter to $A=\frac{d_1+d_2}{d_2}\Gamma$. With $d_1=50$cm, $d_2=1$m, and $\Gamma=5$mm this works out to $7.5$mm, which corresponds to a minimum effective f-number of $6.6$ for our $50$mm lens.

\paragraph{Laser Power}
In this work we imaged retroreflective objects whose trajectories' total exposure times were one to three minutes. \edit{However, as observed in~\cite{OToole:2018}, at even 1 meter standoff distances diffuse objects return $100\times$ less light than high-quality retroreflective hidden objects and this discrepancy grows quadratically with standoff distance. Large-scale real-time keyhole imaging of non-retroflective objects is possible with more powerful lasers, such as those used in~\cite{Liu:2019,Lindell:2019:Wave}, which are $10\,000\times$ brighter than ours. However, doing so with eye-safe lasers will require operating at lower SNRs than our current results. See the second table in the supplement for predicted photons counts under various operating conditions.   
}

\paragraph{Future Work}
Several avenues exist to improve the proposed EM-based reconstruction scheme. 
First, the present algorithm ignores the fact that motion is generally continuous; that is it treats a motion trajectory that bounces back and forth across the room just a likely as a smooth straight path. \edit{Situations like this can result in artifacts as two mirrored versions of the object are reconstructed on top of one another, as described in Section~\ref{ssec:SimResults}.} To address this problem, motion-continuity and other priors on $\mathbf{\trans}$, which fit naturally into the EM framework, could be incorporated. 
Second, while our current method imposes smoothness and sparsity on the reconstruction, more advanced priors, particularly learned priors, could greatly aid in the reconstruction.  
Third, our present results restrict the trajectories to one of three planes. While our implementation supports three dimensional trajectories \edit{(a very simple 3D trajectory is tested in Appendix~\ref{sec:ExtraSim})}, they also require significantly more computation. Developing more efficient algorithms and implementations. \edit{perhaps by restricting the search space using motion continuity priors or by using neural network based priors~\cite{RealTimeUnknownCOSI},} could make our method more generalizable.  
Finally, extending our work to handle non-rigid body motion and self-occlusion are important open problems.

\Section{Conclusion}
\label{sec:conclusion}
NLOS imaging has emerged as an important research direction in the computational imaging and optics communities and is widely recognized for enabling capabilities that would have been impossible only a few years ago.
While seeing around corners has long required imaging a large visible surface, we demonstrate imaging and tracking using a single visible point by exploiting object motion. Moreover, combining time-of-flight imaging with object motion may be useful for other 3D imaging applications.  
We envision that keyhole imaging could unlock new applications, such as NLOS imaging in constrained and cluttered environments.


%


\appendices
\section{Resolution Analysis}
We now derive the resolution limits associated with a $\omega_x$ by $\omega_z$ horizontal synthetic aperture whose corners are at the points $(-\omega_x/2,0,-\omega_z),(\omega_x/2,0,-\omega_z),(\omega_x/2,0,0),$ and $(-\omega_x/2,0,0)$. 
Following the analysis conducted in~\cite{OToole:2018},
we determine how close to each other two points can get in the $x$, $y$, and $z$ directions before they produce indistinguishable measurements. We say two sets of measurements are indistinguishable if they are separated in time by less that the full width at half maximum, $\gamma=2\sqrt{2\ln 2 }\sigma_{SPAD}$, of the temporal jitter of the system. This occurs if there is no sampling point $p$ within the synthetic aperture for which the distance between $p$ and the two points in the scene, $q_1$ and $q_2$, is greater than or equal to $c\gamma/2$, where $c$ is the speed of light.

\begin{figure*}[h]
	\centering
	\includegraphics[width=.8\linewidth]{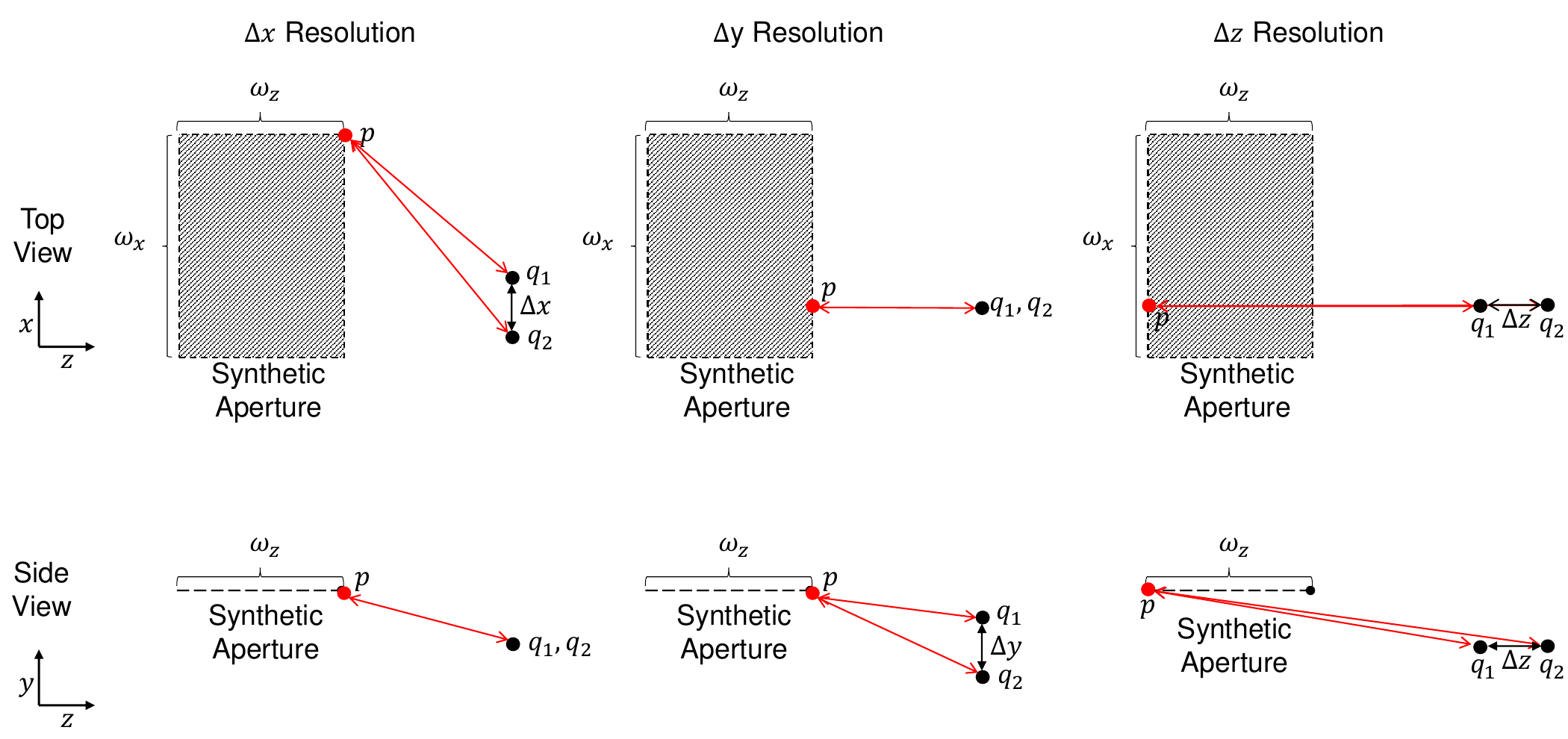}
	\caption{{\bf Resolution Limits.} Two points $q_1$ and $q_2$ are said to be resolvable if, for some point $p$ in the synthetic aperture, the distance between $p$ and $q_1$ is more than $c\gamma/2$ meters longer or shorter than the distance between $p$ and $q_2$.}
	\label{fig:ResLimits2}
\end{figure*}

\subsection{Resolution in x}
We first need to select a point on the synthetic aperture for which two points
$q_1 = (x-\Delta x /2,y,z)$ and $q_2=(x+\Delta x/2, y, z)$ would produce the most easily distinguishable measurements. This is the point $p$ for which the absolute difference between $\|p-q_1\|$ and $\|p-q_2\|$ is maximized. Assuming $\Delta x$ is small, this point can be found by selecting the point $p$ for which the inner product between the vectors $\frac{p-q_1}{\|p-q_1\|}$ and $\frac{q_2-q_1}{\|q_2-q_1\|}$ is maximized. That is, by selecting the point for which $p-q_1$ is best aligned with with $q_1-q_2$. 
Assuming $z\geq0$, this point is $p=(-\text{sign}(x)\omega_x/2,0,0)$. See the left side of Figure~\ref{fig:ResLimits2}.

The resulting distances between $p$ and $q_1$ and $q_2$ are described by
\[
\min(\|q_1-p\|,\|q_2-p\|)=\sqrt{(\omega_x/2+|x|-\Delta x /2)^2+y^2+z^2}
\]
and
\[
\max(\|q_1-p\|,\|q_2-p\|)=\sqrt{(\omega_x/2+|x|+\Delta x /2)^2+y^2+z^2}.
\]

The points $q_1$ and $q_2$ produce distinguishable measurements when ${\big |}\|q_1-p\|-\|q_2-p\|{\big |}\geq c\gamma/2$. This implies that $q_1$ and $q_2$ are resolvable when
\[
\Delta x \geq \frac{c\gamma\sqrt{c^2\gamma^2-4\omega_x^2-16\omega_x |x| -16 |x|^2 -16y^2-16z^2}}{\sqrt{4c^2\gamma^2-16\omega_x^2-64\omega_x |x| -64|x|^2}}.
\]

If we assume $c^2\gamma^2\approx 0$, the above simplifies to 
\begin{align}
	\Delta x \geq\frac{c\gamma}{2}\sqrt{\frac{y^2+z^2}{(|x|+\omega_x/2)^2}+1}.
\end{align}

\subsection{Resolution in y}
As before, we first need to select a point on the synthetic aperture for which two points
$q_1 = (x,y-\Delta y /2,z)$ and $q_2=(x, y+\Delta y/2, z)$ would produce the most easily distinguishable measurements. Assuming $z\geq0$ and $|x|\leq \omega_x/2$, this point is $p=(x,0,0)$. See the middle of Figure~\ref{fig:ResLimits2}.

The resulting distances between $p$ and $q_1$ and $q_2$ are 
\[
\|q_1-p\|=\sqrt{(y-\Delta y/2)^2+z^2}
\]
and
\[
\|q_2-p\|=\sqrt{(y+\Delta y/2)^2+z^2}.
\]

The points $q_1$ and $q_2$ produce distinguishable measurements when ${\big |}\|q_1-p\|-\|q_2-p\|{\big |}\geq c\gamma/2$. This implies that $q_1$ and $q_2$ are resolvable when
\[
\Delta y \geq \frac{c\gamma\sqrt{c^2\gamma^2-16y^2-16z^2}}{\sqrt{4c^2\gamma^2-64y^2}}.
\]

If we assume $c^2\gamma^2\approx 0$, the above simplifies to 
\begin{align}
	\Delta y 
	\geq\frac{c\gamma}{2}\sqrt{\frac{z^2}{y^2}+1}.
\end{align}

\subsection{Resolution in z}
As before, we first need to select a point on the synthetic aperture for which two points
$q_1 = (x,y,z-\Delta z /2)$ and $q_2=(x, y, z+\Delta z /2)$ would produce the most easily distinguishable measurements. Assuming $z\geq0$ and $|x|\leq \omega_x/2$, this point is $p=(x,0,-\omega_z)$. See the right side of Figure~\ref{fig:ResLimits2}.

The resulting distances between $p$ and $q_1$ and $q_2$ are 
\[
\|q_1-p\|=\sqrt{y^2+(\omega_z+z-\Delta z/2)^2}
\]
and
\[
\|q_2-p\|=\sqrt{y^2+(\omega_z+z+\Delta z/2)^2}.
\]

The points $q_1$ and $q_2$ produce distinguishable measurements when ${\big |}\|q_1-p\|-\|q_2-p\|{\big |}\geq c\gamma/2$. This implies that $q_1$ and $q_2$ are resolvable when
\[
\Delta z \geq \frac{c\gamma\sqrt{c^2\gamma^2-16\omega_z^2-32\omega_z z-16y^2-16z^2}}{\sqrt{4c^2\gamma^2-64\omega_z^2-128\omega_z z -64 z^2}}.
\]

If we assume $c^2\gamma^2\approx 0$, the above simplifies to 
\begin{align}
	\Delta z \geq\frac{c\gamma}{2}\sqrt{\frac{y^2}{(\omega_z+z)^2}+1}.
\end{align}

\edit{
\section{Additional Simulation Results}\label{sec:ExtraSim}
}

\newcommand{\mywidthtwo}{.22\textwidth}
\begin{figure}[t]
	\centering
	
	\begin{subfigure}[t]{\mywidthtwo}
		\centering
		\begin{overpic}[width=\textwidth]{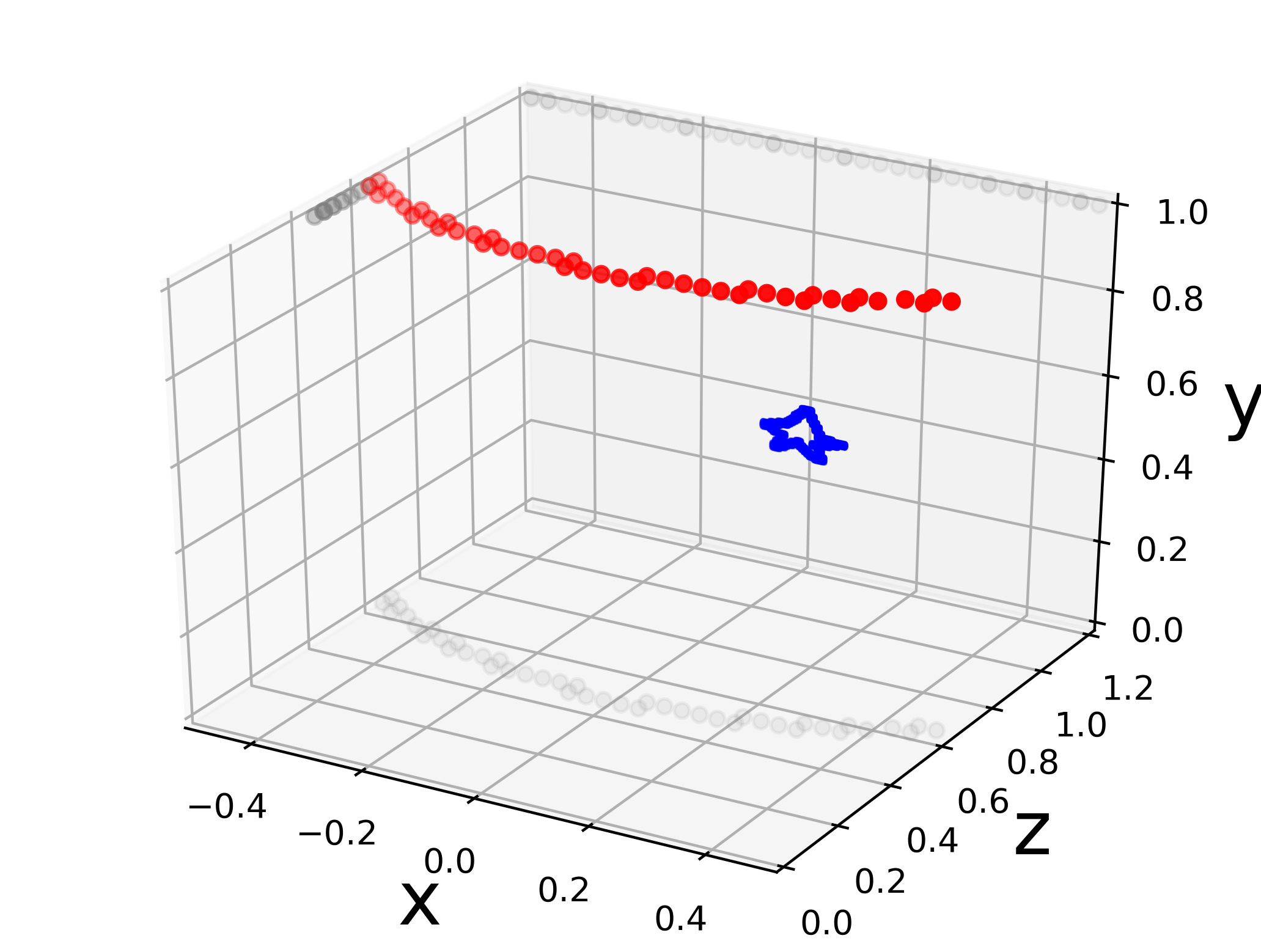}
		\end{overpic}
		\label{fig:ExtraSimSetup_a}
		\caption{}
	\end{subfigure}%
	~
	\begin{subfigure}[t]{\mywidthtwo}
		\centering
		\begin{overpic}[width=\textwidth]{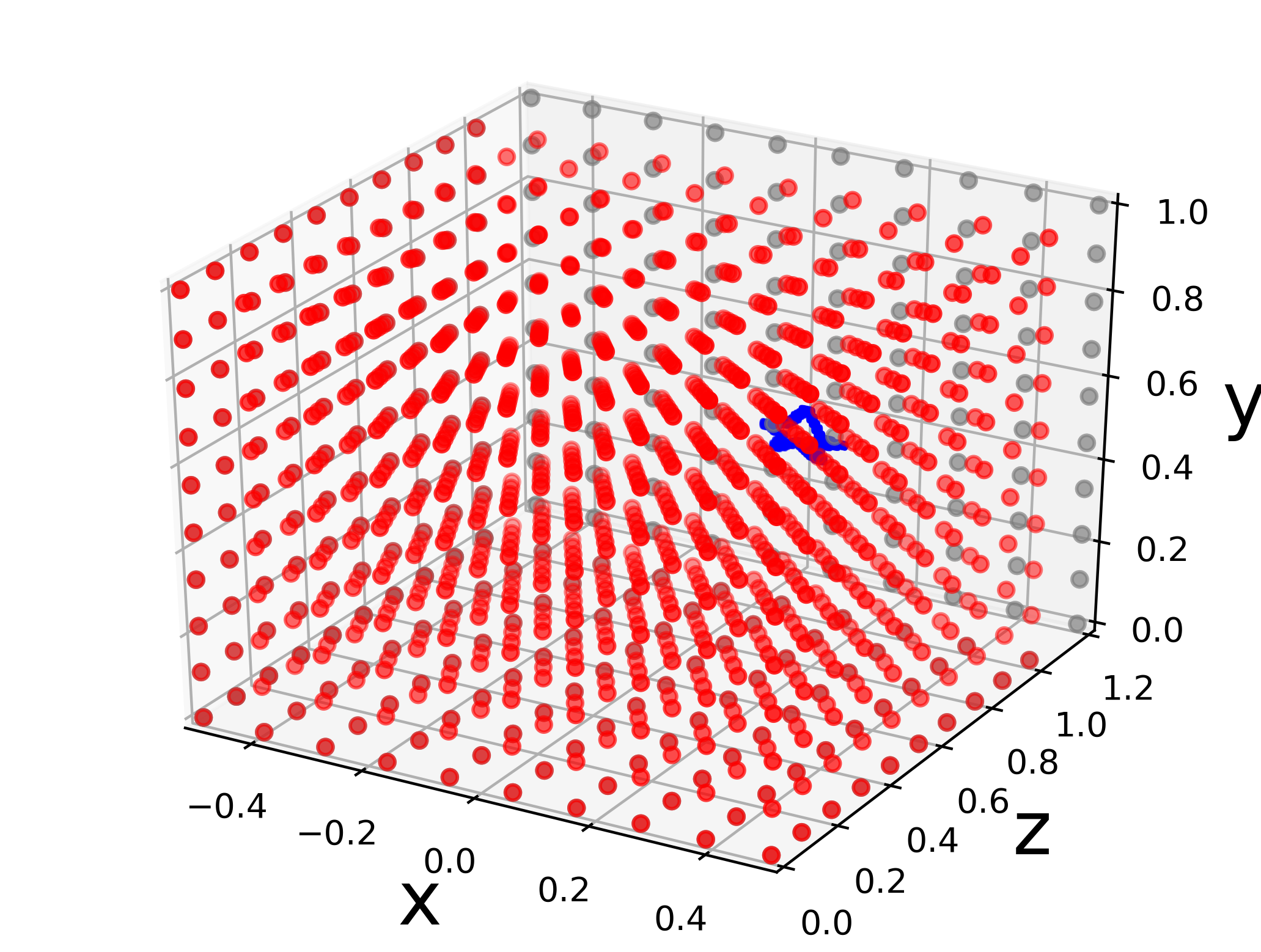}
		\end{overpic}
		\label{fig:ExtraSimSetup_b}
		\caption{}
	\end{subfigure}%
		
	\caption{\edit{\textbf{Rotation-only and 3D trajectory simulation setup.} We simulated reconstructing objects which followed the rotation-only trajectory on the left and the densely sampled 3D trajectory on the right. For the 3D result, the object moved to every point on the grid. }}
	\label{fig:ExtraSimSetup}
\end{figure}

\begin{figure}[t]
	\centering
	
	\hspace*{1.2cm}
	\begin{subfigure}[t]{\mywidth}
	\centering
	\begin{overpic}[width=\textwidth]{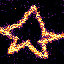}
		\put (-105,45) {{\scriptsize \begin{tabular}{@{}c@{}c@{}} Known\\ Trajectory  \\Reconstructions\end{tabular}}}	
		\put (10,113) {{\scriptsize \begin{tabular}{@{}c@{}c@{}} 50$\times$50 cm\\ Rotation-only \end{tabular}}}	
	\end{overpic}
	\end{subfigure}%
	~
	\begin{subfigure}[t]{\mywidth}
		\centering
		\begin{overpic}[width=\textwidth]{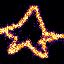}		
		\put (15,113) {{\scriptsize \begin{tabular}{@{}c@{}c@{}} 50$\times$50 cm\\ 3D \end{tabular}}}	
		\end{overpic}		
	\end{subfigure}%
	~
	\begin{subfigure}[t]{\mywidth}
	\centering
	\begin{overpic}[width=\textwidth]{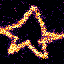}
		\put (10,113) {{\scriptsize \begin{tabular}{@{}c@{}c@{}} 15$\times$15 cm\\ Rotation-only \end{tabular}}}	
	\end{overpic}
	\end{subfigure}%
	~
	\begin{subfigure}[t]{\mywidth}
		\centering
		\begin{overpic}[width=\textwidth]{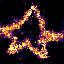}		
			\put (15,113) {{\scriptsize \begin{tabular}{@{}c@{}c@{}} 15$\times$15 cm\\ 3D \end{tabular}}}	
		\end{overpic}		
	\end{subfigure}%

	\vspace{5 pt}
	\hspace*{1.2cm}
	\begin{subfigure}[t]{\mywidth}
	\centering
	\begin{overpic}[width=\textwidth]{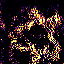}
		\put (-105,45) {{\scriptsize \begin{tabular}{@{}c@{}c@{}} Unknown\\ Trajectory  \\Reconstructions\end{tabular}}}	
	\end{overpic}
	\end{subfigure}%
	~
	\begin{subfigure}[t]{\mywidth}
	\centering
	\begin{overpic}[width=\textwidth]{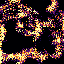}		
	\end{overpic}		
	\end{subfigure}%
	~
	\begin{subfigure}[t]{\mywidth}
	\centering
	\begin{overpic}[width=\textwidth]{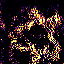}
	\end{overpic}
	\end{subfigure}%
	~
	\begin{subfigure}[t]{\mywidth}
	\centering
	\begin{overpic}[width=\textwidth]{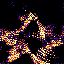}		
	\end{overpic}		
	\end{subfigure}%
	
	\caption{\editTwo{\textbf{Rotation-only and 3D trajectory simulation results.} Reconstructions of the star object with rotation-only and 3D trajectories from measurements with an SNR of $5$. These trajectories are more challenging than the planar, translation-based trajectories from Section~\ref{sec:analysis}. }}
	\label{fig:ExtraSimRecons}
\end{figure}

\edit{
\rowcolors{2}{white}{gray!25}
\begin{table}[t!]%
	\centering
	{\small
		\begin{tabular}{l|cccccc}
			\rowcolor{white}
			\toprule
			\mc{}& \multicolumn{2}{c}{SNR = 5}& \multicolumn{2}{c}{SNR = 15}& \multicolumn{2}{c}{SNR = 50} \\ 
			\cmidrule(lr){2-3}\cmidrule(lr){4-5}\cmidrule(lr){6-7}
			\rowcolor{white}
			\mc{}&  GD & EM & GD&EM & GD&EM \\
			\midrule
			\begin{tabular}{@{}c@{}}Rotation-only Traj.\\ 50 cm $\times$ 50 cm \end{tabular} & \textbf{0.54}&0.43 & \textbf{0.66}&0.46 &\textbf{0.77} &0.46\\
			\begin{tabular}{@{}c@{}}3D Traj.\\ 50 cm $\times$ 50 cm \end{tabular} & \textbf{0.60}&0.43 & \textbf{0.71}&0.44 &\textbf{0.84} &0.44\\
			\begin{tabular}{@{}c@{}}Rotation-only Traj.\\ 15 cm $\times$ 15 cm \end{tabular} & {0.51}&\textbf{0.52} & \textbf{0.60}&0.53 &\textbf{0.68} &0.54\\
			\begin{tabular}{@{}c@{}}3D Traj.\\ 15 cm $\times$ 15 cm \end{tabular} & \textbf{0.60}&0.54 & \textbf{0.68}&0.55 &\textbf{0.75} &0.55\\
			\bottomrule
		\end{tabular}
	}
	\caption{\edit{Comparison of the mean disambiguated SSIM (higher is better) across the 9 tests images with rotation-only and 3D trajectories, various hidden object sizes, and various SNRs. With challenging rotation-only and 3D trajectories, EM works best when restricted to smaller objects, for which the measurements are more easily distinguishable.}}
	\label{tab:ExtraSimResults}
\end{table}%
\rowcolors{2}{}{}
}

In this section we test EM on the more challenging rotation-only and densely-sampled 3D trajectories. \editTwo{Because, under our forward model, the measurements formed by rotating a hidden object are equivalent to the measurements formed by moving the virtual sensor along a circular trajectory around the axis of rotation, we form and represent the rotation-only trajectory using the virtual sensor trajectory illustrated in Figure~\ref{fig:ExtraSimSetup}(a). Our 3D trajectory is illustrated in Figure~\ref{fig:ExtraSimSetup}(b).} The rotation-only trajectory uses the $33\times33$ sampling grid illustrated in Figure~\ref{fig:2l}. In order for EM to fit in the GPU's memory, the 3D sampling grid is restricted to $10\times10\times10$ points; these are the same points that are densely sampled in Figure~\ref{fig:ExtraSimSetup}(b).


We tested these trajectories first under the same conditions as described in Section~\ref{ssec:SimSetup} and then again where we simplified the problem by reducing the size of the simulated images from 50 cm $\times$ 50 cm to 15 cm $\times$ 15 cm. \editTwo{(We kept the resolution at $64\times 64$.)} The smaller objects make it easier for the algorithm to distinguish which measurement belongs to which point on the grid. As Table~\ref{tab:ExtraSimResults} illustrates, EM struggled to accurately reconstruct the hidden objects when they are 50 cm $\times$ 50 cm, but is, on average, roughly as successful as it was with the previous trajectories when the hidden objects are restricted to 15 cm $\times$ 15 cm. 

\edit{
\section*{Acknowledgments}
We would like to thank the reviewers for their feedback and help improving this manuscript. We would also like to thank Prasanna Rangarajan for help with the radiometry analysis.\nocite{PrasannaRevealReport,BelcourBRDFMeasurements}
}

\edit{
C.M.~was supported by an appointment to the Intelligence Community Postdoctoral Research Fellowship Program at Stanford University administered by Oak Ridge Institute for Science and Education (ORISE) through an interagency agreement between the U.S. Department of Energy and the Office of the Director of National Intelligence (ODN). D.L.~was supported by a Stanford Graduate Fellowship. G.W.~was supported by an NSF CAREER Award (IIS 1553333), a Sloan Fellowship, by the KAUST Office of Sponsored Research through the Visual Computing Center CCF grant, and a PECASE by the ARL. 
}

\title{Supplement to ``Keyhole Imaging:\\Non-Line-of-Sight Imaging and Tracking of Moving Objects Along a Single Optical Path''}

\author{Christopher~A.~Metzler,
	David~B.~Lindell,
	and~Gordon~Wetzstein
}

%
%
%


\maketitle

\begin{abstract}
	In this supplement we first derive the maximum powers at which high repetition rate pulsed 670 nm and 1550 nm lasers can operate while remaining eye safe. Then, using these eye safe limits, we predict the number of photons one would record using a meter-scale confocal non-line-of-sight (NLOS) imaging geometry when dealing with Lambertian and retroreflective hidden objects. Next, we show how a quartic falloff model was fit to our experimental data and test the effects using the wrong falloff model have on simulated reconstructions. Finally, we test how sensitive our expectation maximization (EM) based reconstruction algorithm is to its initialization.
\end{abstract}


\IEEEpeerreviewmaketitle

\edit{
	\section{Eye Safe Laser Limits}\label{sec:EyeSafety}
	We follow~\cite{ansi2014american} to work out the maximum powers for which high repetition rate pulsed $670$ nm and $1550$ nm lasers would still be considered eye-safe. According to Table 1 of~\cite{ansi2014american}, for high repetition rate lasers (pulse repetition rate over 100 KHz) the average power limits are more stringent than the per-pulse limits. Table 2 of~\cite{ansi2014american} recommends using a maximum anticipated exposure duration of $.25$ seconds for visible light (due to the blink response) and $10$ seconds for 1550 nm light (which does not induce blinking). Table 5b of \cite{ansi2014american} states that at 670 nm the maximum permissible exposure (MPE) is $1.8t^{0.75}\cdot10^{-3}$ J$\cdot$cm$^{-2}$ and at 1550 nm the MPE is $.1$ W$\cdot$cm$^2$.
	Using Table 9 of \cite{ansi2014american}, we see that for visible wavelengths, the retina has a limiting aperture diameter of $7$ mm. For $1550$ nm, the cornea has a limiting aperture of $1.5t^{0.375}$ mm for exposures between $.3$ and $10$ seconds.}

\edit{
	Thus for 670 nm, the eye-safe limit over a quarter second exposure is  
	\begin{align}
	\frac{1.8(0.25)^{0.75}\cdot10^{-3}\cdot(\pi0.35^2)}{0.25}= 0.98\text{ mW.}
	\end{align}Our 670 nm laser is well below these limits.
}
\\

\edit{
	For 1550 nm, the eye-safe limit is 
	\begin{align}
	.1\cdot \pi(.075(10)^{0.375})^2=9.9\text{ mW.}
	\end{align}
}

\edit{
	\Section{Radiometry}
	\nocite{PrasannaRevealReport}
}

\begin{figure}[h]
	\centering
	\includegraphics[width=.8\linewidth]{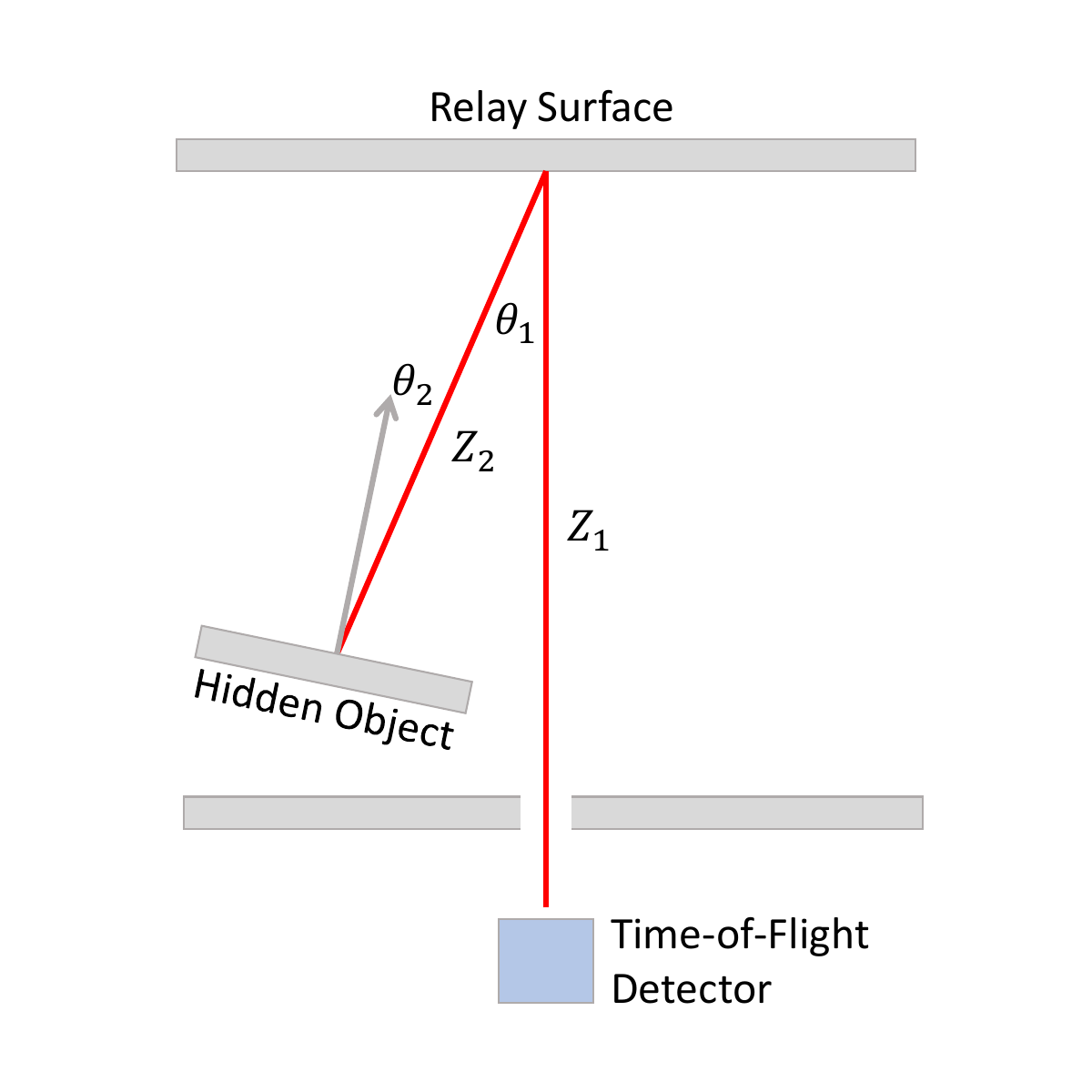}
	\caption{{\bf Radiometry.} A large scale keyhole imaging scene where the time-of-flight detector is orthogonal to the relay wall.}
	\label{fig:Radiometry}
\end{figure}

Keyhole imaging is inherently a confocal imaging geometry. This means it particularly benefits from retroreflective hidden objects. However, many real-world objects are diffuse. Does keyhole imaging (and confocal imaging configurations generally) record enough photons to reconstruct diffuse hidden objects at realistic standoff distances?

While characterizing the exact number of photons required for EM to successfully reconstruct a hidden object is beyond the scope of this work, we can analyze how many photons would be returned in a large-scale imaging geometry with confocal measurements of a diffuse hidden object. Such an analysis tells us if there is any hope of large-scale keyhole imaging with eye-safe lasers -- if on average almost no photons return then it is impossible.
\\

In this section, we analyze the light throughput associated with the confocal NLOS imaging geometry illustrated in Figure~\ref{fig:Radiometry}. We assume the relay wall is Lambertian and make the simplifying assumption that all points on the hidden object are $Z_2$ meters away from the relay wall at an angle of $\theta_1$ with respect to the relay wall's normal vector. We similarly assume that the illuminated point on the relay surface is at an angle of $\theta_2$ with respect to the hidden object's surface normal for all points on the hidden object.

We first analyze the light throughput associated with this system when the hidden object is Lambertian. We then analyze the light throughput when the hidden object is a retroreflector of variable quality, as determined by the size of the lobes over which it reflects light. Our analysis is loosely based off the analysis performed in~\cite{PrasannaRevealReport}. 

\subsection{Lambertian Hidden Object}

\paragraph{From the laser to the relay wall}

An $A_{L}$ sized area of the relay wall is illuminated with a laser with power $P$. The relay wall has an incident irradiance of 
\begin{align}
E_{R1}=\frac{P}{A_{L}}\frac{\text{W}}{\text{m}^{2}}.
\end{align}
\paragraph{From the relay wall to the hidden object}

The Lambertian relay wall, with albedo $\rho_R$, has a radiance of
\begin{align}
L_{R1} = E_{R1} \frac{\rho_R}{\pi}\frac{\text{W}}{\text{m}^{2}\text{sr}}.
\end{align}
That is, if a pupil is viewing the relay wall and the relay wall has a solid angle of $dSA_p$ from the pupil's perspective, then $L_{R1}dSA_p$ Watts of light would make it to each square meter of a surface in the pupil that is orthogonal to the directional vector between the pupil and the relay wall.

If a pupil is $Z_2$ meters from the relay wall at an angle of $\theta_1$ with respect to the relay wall's normal vector, then every unit area of the relay wall spans a solid angle of $\frac{\cos(\theta_1)}{Z_2^2}$ steradians from its perspective. That is, $\frac{dSA_p}{dA_R}=\frac{\cos(\theta_1)}{Z_2^2}$.

Thus the total irradiance on a pupil at this location would be
\begin{align}
E_{p_O} = \int_{A_l} L_{R1} \frac{dSA_p}{dA_R} dA_R = L_{R1} A_{l} \frac{\cos(\theta_1)}{Z_2^2}\frac{\text{W}}{\text{m}^{2}}.
\end{align}

Light that passes through a $dA_p$ area of the pupil (which is orthogonal to the directional vector between the object and the relay wall) is spread out over an area of $\frac{1}{\cos(\theta_2)}$ times larger on the hidden object. Thus $\frac{dA_p}{dA_o}=\cos(\theta_2)$ and
\begin{align}
E_{O} = L_{R1} A_{l} \frac{\cos(\theta_1)\cos(\theta_2)}{Z_2^2}\frac{\text{W}}{\text{m}^{2}}.
\end{align}

\paragraph{From the hidden object back to the relay wall}

The Lambertian hidden object, with albedo $\rho_O$, has a radiance of 
\begin{align}
L_{O} = E_{O} \frac{\rho_O}{\pi}\frac{\text{W}}{\text{m}^{2}\text{sr}}.
\end{align}

From the perspective of a pupil at the relay wall, a unit area of the hidden object spans a solid angle of $\frac{\cos(\theta_2)}{Z_2^2}$ steradians and thus $\frac{dSA_p}{dA_O}=\frac{\cos(\theta_2)}{Z_2^2}$.

Accordingly, the total irradiance on pupil surface orthogonal to the directional vector between the hidden object and the relay wall is 
\begin{align}
E_{p_{R2}} = \int_{A_{O}} L_{O} \frac{dSA}{dA} dA =  L_{O} A_{O} \frac{\cos(\theta_2)}{Z_2^2}\frac{\text{W}}{\text{m}^{2}},
\end{align}
and the irradiance on the relay wall itself (where the light has been spread out over an area $1/\cos(\theta_1)$ times larger than that of the pupil) is
\begin{align}
E_{R2} = \int_{A_{O}} L_{O} \frac{dSA}{dA} dA =  L_{O} A_{O} \frac{\cos(\theta_1)\cos(\theta_2)}{Z_2^2}\frac{\text{W}}{\text{m}^{2}}.
\end{align}

\paragraph{From the relay wall to the detector}

The Lambertian relay wall, with albedo $\rho_R$, has a radiance of 
\begin{align}
L_{R2} = E_{R2} \frac{\rho_R}{\pi}\frac{\text{W}}{\text{m}^{2}\text{sr}}.
\end{align}

From the perspective of the detector's aperture, a unit area of the relay wall spans a solid angle of $\frac{1}{Z_1^2}$ steradians and thus $\frac{dSA}{dA}=\frac{1}{Z_1^2}$.

Thus the detection aperture, with a field of view of $A_{FoV}$ on the wall, experiences an irradiance of 
\begin{align}
E_{Apt} = \int_{A_{FoV}} L_{R2} \frac{dSA}{dA} dA=L_{R2} A_{FoV}\frac{1}{Z_1^2}\frac{\text{W}}{\text{m}^{2}}.
\end{align}

\paragraph{From aperture to photon counts}

By integrating over the camera's entire aperture, we find the total power incident on the detector is 
\begin{align}
P_{detector} = A_{apt}E_{Apt}\text{ W}.
\end{align}

If the detector has a quantum efficienty of $Q$ and the laser has wavelength of $\lambda$ then the detector records
\begin{align}
P_{detector}\frac{Q \lambda}{hc}\text{ photons per second},
\end{align}
where $h$ is Plank's constant and $c$ is the speed of light.

\paragraph{Putting it all together}

We find that the total number of photons per second recorded by the detector is described by
\begin{align}
\frac{Q \lambda}{hc} P {\Big (}\frac{\rho_r^2\rho_O}{\pi^3}{\Big )} A_{apt} A_{FoV} A_O \frac{\cos^2(\theta_1)\cos^2(\theta_2)}{Z_2^4Z_1^2}.
\end{align}

\subsection{Retroreflective Hidden Object}

\edit{
	Retroreflective objects reflect the majority of their incident light over a narrow lobe centered at the angle of the incident illumination; $\theta_2$ in Figure~\ref{fig:Radiometry}. The quality of the retro-reflector determines the width of this lobe; a higher quality retroreflector directs the light over a smaller range of angles. 
}

\edit{
	Consider a simplified retroreflective model where the retroreflector reflects all the light coming from direction $\theta_2$ uniformly over all angles within $\eta/2$ of $\theta_2$, where the lobe width $\eta$ is determined by the quality of the retroreflector. That is, the retroreflector sends all the incident light to a region spanning $\pi\frac{\eta^2}{4}$ steradians from its perspective.
}

\edit{
	Denote the area on the relay wall that is imaged by the detector the ``virtual detector''. It has an area $A_{FoV}$. Under the aforementioned model, when the virtual detector subtends a solid angle larger than $\pi\frac{\eta^2}{4}$ then the light it gathers is proportional to all the light that reaches the retroreflector. 
}

\edit{
	However, the size of $\eta$ depends on the quality of the retroreflector; retroreflective tape has a lobe width of $\eta\approx5\text{--}10$ degrees~\cite{BelcourBRDFMeasurements}. When $\eta$ is large, the retroreflector may direct the light over a solid angle larger than that subtended by the virtual detector. In this case, the fraction of light that reaches the virtual detector from the retroreflector is proportional to the solid angle subtended by the virtual detector over the solid angle subtended by the lobe.
}

We now analyze the light throughput associated with each scenario.

%
\subsubsection{Small lobes, $SA_{FoV}>SA_{lobe}$}

When the solid angle subtended by the lobe, $SA_{lobe}$ is smaller than the solid angle subtended by the relay wall $SA_{FoV}$, the total power of the light that reaches the relay wall from the object is that same the power of the light that reaches the hidden object. That is, the power of the light on the relay wall from the hidden object is 
\begin{align}
\int_{A_O} E_{O} dA = A_O E_{O} \text{ W.}
\end{align}

This light is directed exclusively an area $A_{lobe}$ on the relay wall and thus the irradiance on the relay wall from the hidden object is 
\begin{align}\label{eqn:retro_ER2}
E_{R2} =  \frac{ A_O E_{O}}{A_{lobe}} \frac{\text{W}}{\text{m}^{2}},
\end{align}
over the area $A_{lobe}$ and $0$ elsewhere.

Likewise, 
\begin{align}\label{eqn:retro_LR2}
L_{R2} = E_{R2} \frac{\rho_R}{\pi}\frac{\text{W}}{\text{m}^{2}\text{sr}},
\end{align}
over the area $A_{lobe}$ and $0$ elsewhere.

Because $A_{lobe}<A_{FoV}$ and the radiance outside the lobe is $0$, when we integrate over the virtual detector to determine the irradiance on the camera's aperture we find
\begin{align}\label{eqn:retro_Eapt}
E_{Apt} 
&=\int_{A_{FoV}} L_{R2} \frac{dSA}{dA} dA,\nonumber\\
&=\int_{A_{lobe}} L_{R2} \frac{dSA}{dA} dA,\nonumber\\
&=L_{R2} A_{lobe}\frac{1}{Z_1^2}\frac{\text{W}}{\text{m}^{2}}.
\end{align}

The $A_{lobe}$ term in \eqref{eqn:retro_Eapt} cancels out the $A_{lobe}$ term in \eqref{eqn:retro_ER2} and we eventually find that the detector records
\begin{align}
\frac{Q \lambda}{hc} P {\Big (}\frac{\rho_r^2}{\pi^2}{\Big )} A_{apt} A_O \frac{\cos(\theta_1)\cos(\theta_2)}{Z_1^2Z_2^2},
\end{align}
photons per second.

\subsubsection{Large lobes, $SA_{FoV}<SA_{lobe}$}

Assume, as before, that the solid angle subtended by the lobe, $SA_{lobe}$ is smaller than the solid angle subtended by the entire relay wall $SA_{FoV}$. Then, as before, all the light that reaches the retroreflector reaches the relay wall and is spread out over an area $A_{lobe}$. Thus, the radiance from the relay wall is again
\begin{align}
L_{R2} = E_{R2} \frac{\rho_R}{\pi}\frac{\text{W}}{\text{m}^{2}\text{sr}},
\end{align}
over the area $A_{lobe}$ and $0$ elsewhere, where $E_{R2}$ is defined as in \eqref{eqn:retro_ER2}.

In calculating the irradiance on the camera's aperture, we arrive at
\begin{align}
E_{Apt} 
&=\int_{A_{FoV}} L_{R2} \frac{dSA}{dA} dA,\nonumber\\
&=L_{R2} A_{FoV}\frac{1}{Z_1^2}\frac{\text{W}}{\text{m}^{2}}.
\end{align}

Without an $A_{lobe}$ term to cancel in the expression for $E_{Apt}$, if we substitute in 
\begin{align}
&A_{lobe} = \nonumber\\
&\frac{1}{2}\pi Z_2^2 \sin(\eta/2)\tan(\eta/2) {\Big [} \frac{1}{\cos(\eta/2-\theta_1)}+\frac{1}{\cos(\eta/2+\theta_1)} {\Big ]}\label{eqn:ellipsearea}\\
&\approx \frac{\pi Z_2^2\tan^2(\eta/2)}{\cos(\theta_1)},\label{eqn:circlearea}
\end{align}
(\eqref{eqn:ellipsearea} is determined by the area of the ellipse formed by intersecting the cone of light from the retroreflector with the planar relay wall $Z_2$ meters away, \eqref{eqn:circlearea} is the circular approximation of this ellipse)
we find that the detector now records 
\begin{align}
\frac{Q \lambda}{hc} P {\Big (}\frac{\rho_r^2}{\pi^3}{\Big )} A_{apt} A_{FoV} A_O \frac{\cos^2(\theta_1)\cos(\theta_2)}{\tan^2(\eta/2)Z_1^2Z_2^4},
\end{align}
photons per second.

\subsection{Quantitative Example}
\begin{table}[t!]%
	\centering
	{\small
		\begin{tabular}{l|cc}
			\rowcolor{white}
			\toprule
			$Z_1$ & 2.5 m& \\
			$Z_2$ & 2 m &\\
			$\theta_1$ & 10 degrees& \\
			$\theta_2$ & 10 degrees& \\
			$\rho_O$ & 1&\\
			$\rho_R$ & 1 &\\
			$A_O$ & 1 m$^2$ &\\
			$A_L$ & $\pi (.5)^2$ mm$^2$& 1mm beam diameter\\
			$A_{FoV}$ & 4 cm$^2$ &\\
			$A_{apt}$ & $\pi (\frac{1}{2}\frac{50}{3})^2 $ mm$^2$& f-number of 3 on 50mm lens\\
			$Q$ & 30\% &\\
			\bottomrule
		\end{tabular}
	}
	\caption{\textbf{Parameters} These numbers are used for the light throughput estimates.}
	\label{tab:Parameters}
\end{table}%

\begin{table}[t!]%
	\centering
	{\small
		\begin{tabular}{l|ccc}
			\rowcolor{white}
			\toprule
			&\begin{tabular}{@{}c@{}}670~nm\\ 0.1~mW  \end{tabular}  &\begin{tabular}{@{}c@{}}1550~nm\\ 9.9~mW  \end{tabular}  & \begin{tabular}{@{}c@{}}532~nm\\ 1~W  \end{tabular}\\
			\midrule
			Lambertian & 2.7K & 613K &21.3M \\
			Perfect  Retroreflector   &86.7M &19.9B &688B \\
			\begin{tabular}{@{}c@{}}Retroreflective Tape\\ (5 degree lobe) \end{tabular} & 1.4M & 326M& 11.3B\\
			\bottomrule
		\end{tabular}
	}
	\caption{\textbf{Third-bounce Photons per Second} Predicted third-bounce photons per second detected with different hidden objects and lasers (K~=~thousand, M~=~million, B~=~billion).}
	\label{tab:PhotonCounts}
\end{table}%

Consider the setup described by Table~\ref{tab:Parameters}, where a one square meter hidden object is two meters from a relay wall. We compare the number of third-bounce photons are analysis predicts would be recorded with a 670~nm 0.1~mW laser like ours, a barely eye-safe 9.9~mW 1550~nm laser, and a 1~W 532~nm laser, as used in~\cite{Liu:2019,Lindell:2019:Wave}. We perform this analysis for a Lambertian hidden object, a perfectly retro-reflective hidden object, and a hidden object covered in real-world retroreflective tape with a lobe size, $\eta$, of 5 degrees. Our results are presented in Table~\ref{tab:PhotonCounts}.

With a Lambertian hidden object 2 meters from the relay wall a barely eye-safe 9.9~mW 1550~nm laser would record 613 thousand third-bounce photons per second while a 1~W 532~nm laser would record 21 million third-bounce photons per second. For reference, our experiments used hidden objects covered in retroreflective tape at a shorter standoff and recorded 7--15 million photons per reconstruction. These results suggest meter-scale real-time keyhole imaging of diffuse hidden objects is possible, but that doing so in real-time with eye-safe lasers will require operating at lower SNRs than our present results.

\section{Falloff Model}\label{sec:DropoffModel}
The bidirectional reflectance distribution function (BRDF) of the hidden object determines the falloff model, $g_{\textrm{exp}}$, associated with our measurements. Consider a confocal NLOS imaging geometry with a hidden planar object parallel to the relay surface (Figure~\ref{fig:Radiometry} with $\theta_1=\theta_2=\theta$). A Lambertian surface should have falloff proportional to $Z_2^4 \cos^{-4}(\theta)$ while a perfectly retroreflective surface should have falloff proportional to $ Z_2^2 \cos^{-2}(\theta)$.


We fit a falloff model to our real-world experimental data as follows. We place a $31\times31$ cm square covered in our retroreflective tape parallel to the wall such that it is $19.5$ cm below where the laser hit the relay wall. It is then moved to a series of known locations, as illustrated by the red virtual sensor locations in Figure~\ref{fig:Calibration_A}. The resulting real-world measurements are presented in Figure~\ref{fig:Calibration_B}. Using the object's known shape and trajectory, we predict what the observations would look like under an idea Lambertian model, Figure~\ref{fig:Calibration_C}, and an ideal retroreflective model, Figure~\ref{fig:Calibration_D}. Both models' predictions are scaled so as to the have the same maximum as the real-world measurements. While neither is perfect, the Lambertian model is a better fit to the real-world data.

Recall retroreflective tape has a lobe width $\eta\approx5\text{--}10$ degrees~\cite{BelcourBRDFMeasurements}. At a $0.62$ m standoff perpendicular to the relay wall a $5$ degree beam width corresponds to a 23 cm$^2$ area on the relay wall. Our detector images a smaller area and should therefore (assuming the radiance is uniform over the entire lobe) be subject to $Z_2^4 \cos^{-3}(\theta)$ falloff, which is very similar to the Lambertian model we fit.

\begin{figure*}[t]
	\centering
	
	\begin{subfigure}[t]{.45\textwidth}
		\centering
		\begin{overpic}[width=\textwidth]{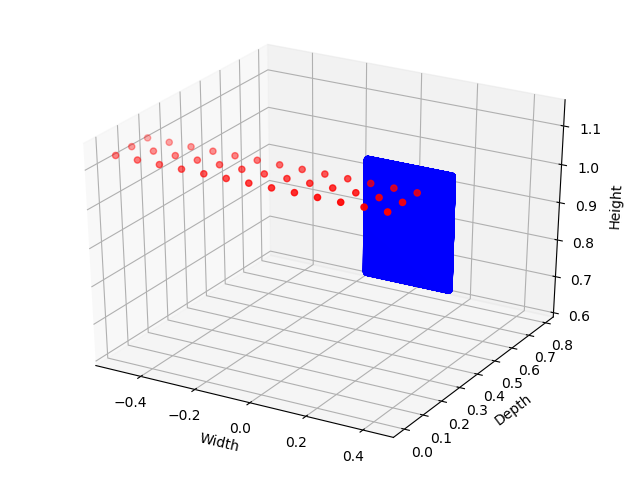}
		\end{overpic}
		\caption{Experimental Setup}
		\label{fig:Calibration_A}
	\end{subfigure}%
	~
	\begin{subfigure}[t]{.45\textwidth}
		\centering
		\begin{overpic}[width=\textwidth]{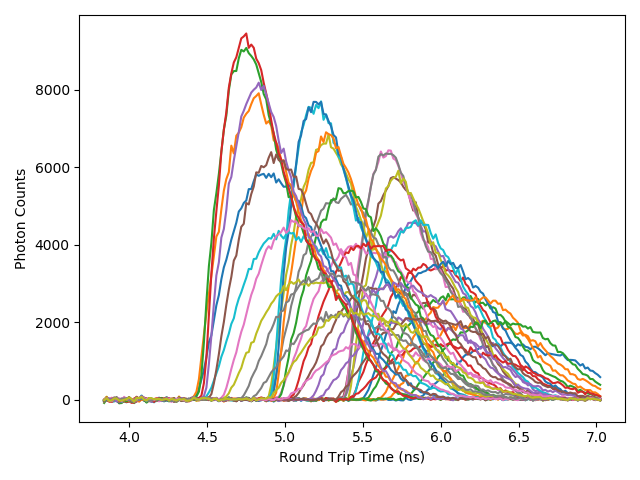}
		\end{overpic}
		\caption{Experimental Measurements}
		\label{fig:Calibration_B}
	\end{subfigure}%
	\vspace{1mm}
	
	\begin{subfigure}[t]{.45\textwidth}
		\centering
		\begin{overpic}[width=\textwidth]{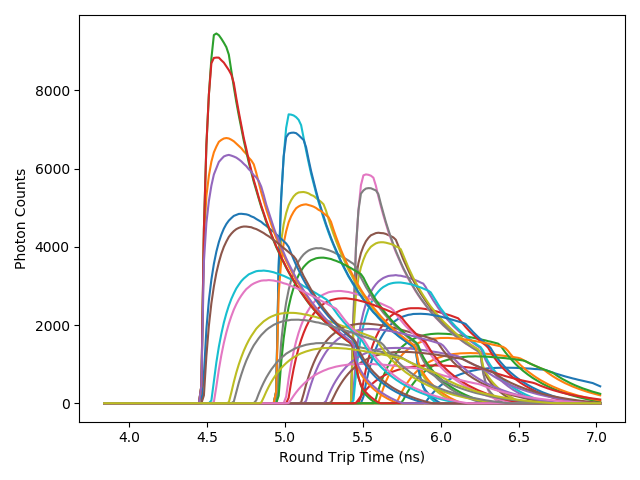}
		\end{overpic}
		\caption{Lambertian Predictions}
		\label{fig:Calibration_C}
	\end{subfigure}%
	~
	\begin{subfigure}[t]{.45\textwidth}
		\centering
		\begin{overpic}[width=\textwidth]{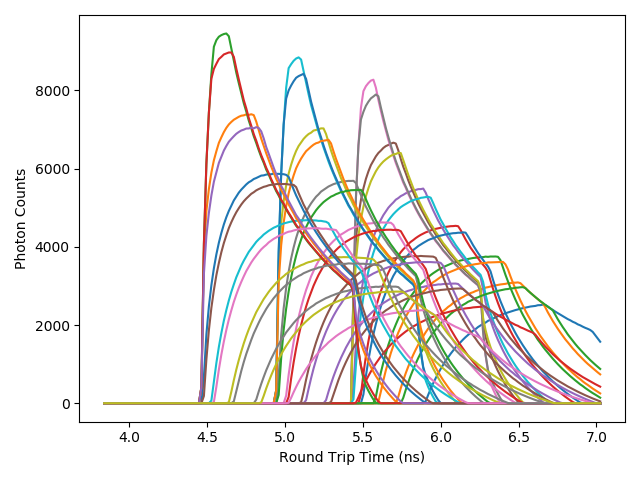}
		\end{overpic}
		\caption{Retroreflective Predictions}
		\label{fig:Calibration_D}
	\end{subfigure}%
	\caption{\edit{\textbf{Determining the falloff model.} A representation of our experimental setup is illustrated in (a), where the object is illustrated in blue and the virtual sensor locations are illustrated in red. This setup gave rise to the measurements (b). The measurements are more similar to those predicted by a Lambertian falloff model (c) than a retroreflective falloff model (d): The real data drops off much more quickly than the retroreflective predictions.}}
\end{figure*}

\edit{
	\section{Sensitivity to Model Mismatch}
	Our method assumes that the falloff model is known in advance, but in the real-world this is rarely the case. In this section we simulate measuring a diffuse hidden object, $g(x)=\left\| \xglob \right\|_2^4\cos^{-4}(\phi)$, and reconstructing it with a retroreflective model, $g(x)=\left\| \xglob \right\|_2^2\cos^{-2}(\phi)$, as well as measuring a retroreflective object and reconstructing it using a diffuse model. We use the same dataset as in Section~IV of the main text,
	operate at an SNR of 15, and use trajectory (i) from Figure~2 of the main text.
}

\edit{
	The top row of Figure~\ref{fig:ModelMismatchSimResults} demonstrates the results of reconstructing a retroreflective hidden object using a diffuse model: The regions of the reconstructions close to the sensor (top) are too dim and the regions far form the sensor (bottom) are too bright, but the overall structure is largely correct.
	The bottom row of Figure~\ref{fig:ModelMismatchSimResults} demonstrates the results of reconstructing a diffuse hidden object using a retroreflective model: The reconstructions were more prone to failure and the regions of the reconstructions close to the sensor (top) are too bright and the regions far form the sensor (bottom) are too dim.
}

\begin{figure*}
	\centering
	\hspace*{1.2cm}
	\begin{subfigure}[t]{\mywidth}
		\centering
		\begin{overpic}[width=\textwidth]{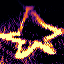}
			\put (-120,45) {{\scriptsize \begin{tabular}{@{}c@{}} Retroflective Data\\ Lambertian Model \end{tabular}}}	
		\end{overpic}
	\end{subfigure}%
	~
	\begin{subfigure}[t]{\mywidth}
		\centering
		\begin{overpic}[width=\textwidth]{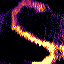}
		\end{overpic}
	\end{subfigure}%
	~
	\begin{subfigure}[t]{\mywidth}
		\centering
		\begin{overpic}[width=\textwidth]{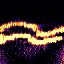}
		\end{overpic}
	\end{subfigure}%
	~
	\begin{subfigure}[t]{\mywidth}
		\centering
		\begin{overpic}[width=\textwidth]{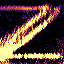}
		\end{overpic}
	\end{subfigure}%
	~
	\begin{subfigure}[t]{\mywidth}
		\centering
		\begin{overpic}[width=\textwidth]{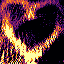}
		\end{overpic}
	\end{subfigure}%
	~
	\begin{subfigure}[t]{\mywidth}
		\centering
		\begin{overpic}[width=\textwidth]{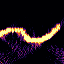}
		\end{overpic}
	\end{subfigure}%
	~
	\begin{subfigure}[t]{\mywidth}
		\centering
		\begin{overpic}[width=\textwidth]{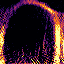}
		\end{overpic}
	\end{subfigure}%
	~
	\begin{subfigure}[t]{\mywidth}
		\centering
		\begin{overpic}[width=\textwidth]{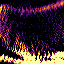}
		\end{overpic}
	\end{subfigure}%
	~
	\begin{subfigure}[t]{\mywidth}
		\centering
		\begin{overpic}[width=\textwidth]{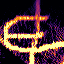}
		\end{overpic}
	\end{subfigure}%
	
	\vspace{17 pt}
	\hspace*{1.2cm}
	\begin{subfigure}[t]{\mywidth}
		\centering
		\begin{overpic}[width=\textwidth]{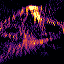}
			\put (-124,45) {{\scriptsize \begin{tabular}{@{}c@{}} Lambertian Data \\ Retroflective Model  \end{tabular}}}	
		\end{overpic}
	\end{subfigure}%
	~
	\begin{subfigure}[t]{\mywidth}
		\centering
		\begin{overpic}[width=\textwidth]{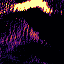}
		\end{overpic}
	\end{subfigure}%
	~
	\begin{subfigure}[t]{\mywidth}
		\centering
		\begin{overpic}[width=\textwidth]{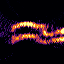}
		\end{overpic}
	\end{subfigure}%
	~
	\begin{subfigure}[t]{\mywidth}
		\centering
		\begin{overpic}[angle=0,width=\textwidth]{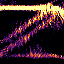}
		\end{overpic}
	\end{subfigure}%
	~
	\begin{subfigure}[t]{\mywidth}
		\centering
		\begin{overpic}[width=\textwidth]{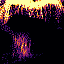}
		\end{overpic}
	\end{subfigure}%
	~
	\begin{subfigure}[t]{\mywidth}
		\centering
		\begin{overpic}[width=\textwidth]{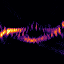}
		\end{overpic}
	\end{subfigure}%
	~
	\begin{subfigure}[t]{\mywidth}
		\centering
		\begin{overpic}[width=\textwidth]{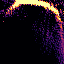}
		\end{overpic}
	\end{subfigure}%
	~
	\begin{subfigure}[t]{\mywidth}
		\centering
		\begin{overpic}[width=\textwidth]{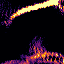}
		\end{overpic}
	\end{subfigure}%
	~
	\begin{subfigure}[t]{\mywidth}
		\centering
		\begin{overpic}[width=\textwidth]{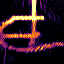}
		\end{overpic}
	\end{subfigure}%
	\vspace{12 pt}
	
	\caption{\edit{\textbf{Simulations with model mismatch.} Top: Reconstructions of objects sampled, along trajectory (i), using a retroreflective model and reconstructed using a Lambertian model. Bottom: Reconstructions of objects sampled using a Lambertian model and reconstructed using a retroreflective model.}}
	\label{fig:ModelMismatchSimResults}
\end{figure*}

\section{Sensitivity to Initialization}
\edit{
	Our implementation of EM relies upon annealing~\cite{ueda1998deterministic}. With annealing, EM starts with an extremely high temperature parameter, which causes the algorithm to assume it has an unreliable estimate of the object's trajectory. Over time, this temperature parameter is reduced and the algorithm trusts the trajectory estimate more and more.
}

\edit{
	During the first few iterations of annealed EM, while the algorithm does not trust its estimates of the object's trajectory, it tries to find a single albedo estimate that explains all the observations regardless of the object's trajectory. This behavior results in the algorithm first finding a blob-like structure that loosely explains all the observations, regardless of how it was initialized.} Figure \ref{fig:MultiRecons} shows multiple reconstructions of the dataset from Section~IV of the main text 
using different random initializations. Thanks to annealing, they are nearly indistinguishable.

\begin{figure*}[t]
	\centering
	\hspace*{1.2cm}
	\begin{subfigure}[t]{.105\textwidth}
		\centering
		\begin{overpic}[width=\textwidth]{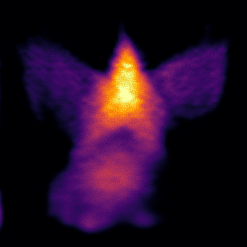}
			\put (3,25) {\textcolor{white}{\scriptsize $\uparrow$}}
			\put (2,12) {\textcolor{white}{\scriptsize $y$}}
			\put (12,2) {\textcolor{white}{\scriptsize $x \rightarrow$}}
			\put (-70,45) {{\scriptsize \begin{tabular}{@{}c@{}} Initialization \\ 1 \end{tabular}}}
		\end{overpic}
		\vspace*{-20pt} 
	\end{subfigure}%
	~ 
	\begin{subfigure}[t]{.105\textwidth}
		\centering
		\scalebox{1}[1]{\includegraphics[width=\textwidth]{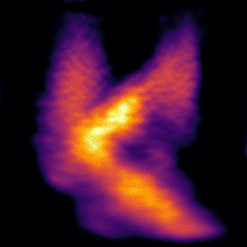}}
	\end{subfigure}%
	~ 
	\begin{subfigure}[t]{.105\textwidth}
		\centering
		\scalebox{1}[1]{\includegraphics[width=\textwidth]{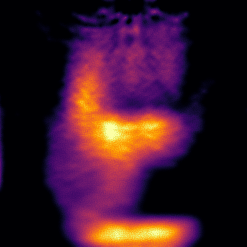}}
	\end{subfigure}%
	~ 
	\begin{subfigure}[t]{.105\textwidth}
		\centering
		\includegraphics[width=\textwidth]{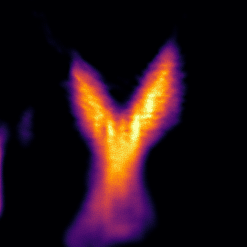}
	\end{subfigure}%
	~
	\begin{subfigure}[t]{.105\textwidth}
		\centering
		\begin{overpic}[width=\textwidth]{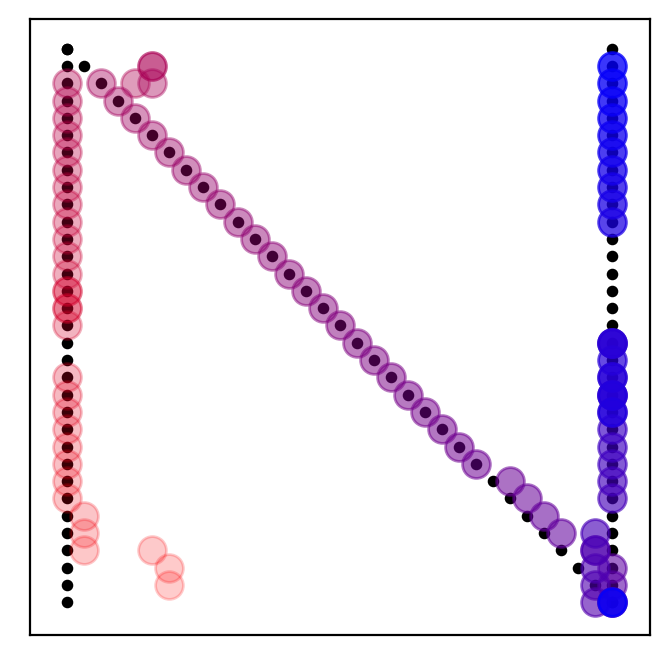}
			\put (-7,25) {{\scriptsize $\uparrow$}}
			\put (-7,12) {{\scriptsize $x$}}				
			\put (12,-5) {{\scriptsize $z \rightarrow$}}
		\end{overpic}	
	\end{subfigure}%
	~ 
	\begin{subfigure}[t]{.105\textwidth}
		\centering
		\scalebox{1}[1]{\includegraphics[width=\textwidth]{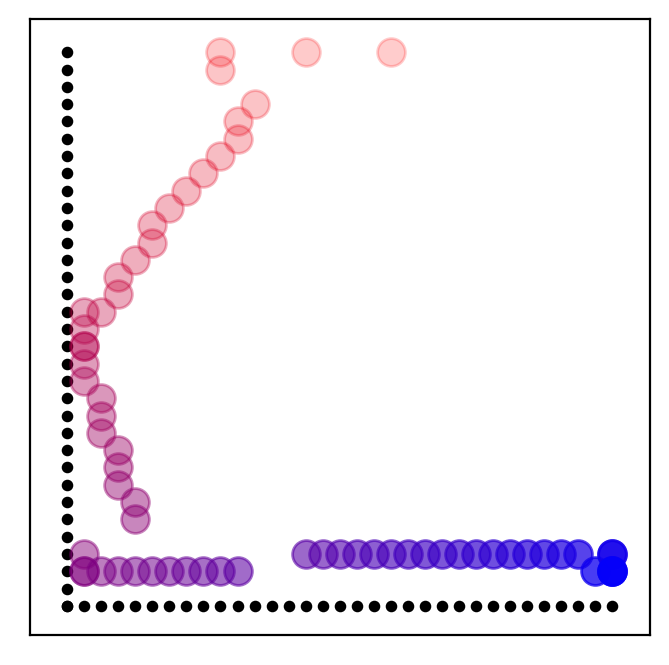}}
	\end{subfigure}%
	~ 
	\begin{subfigure}[t]{.105\textwidth}
		\centering
		\scalebox{1}[1]{\includegraphics[width=\textwidth]{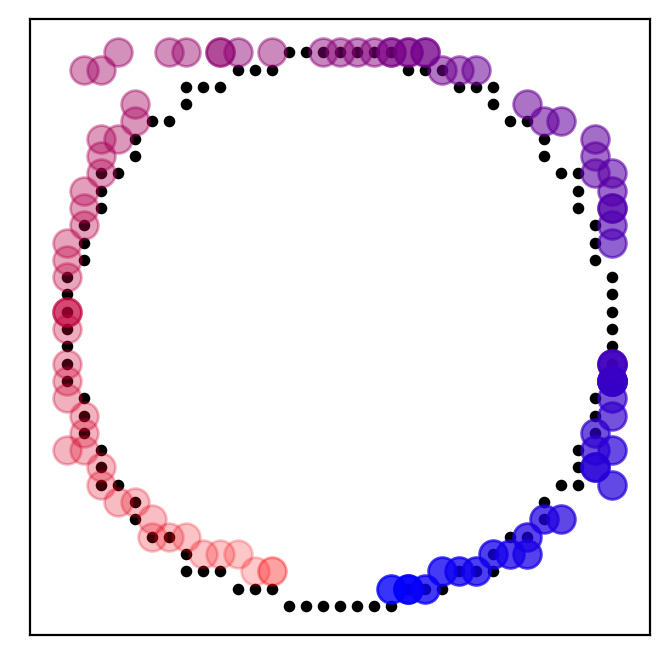}}
	\end{subfigure}%
	~ 
	\begin{subfigure}[t]{.105\textwidth}
		\centering
		\includegraphics[width=\textwidth]{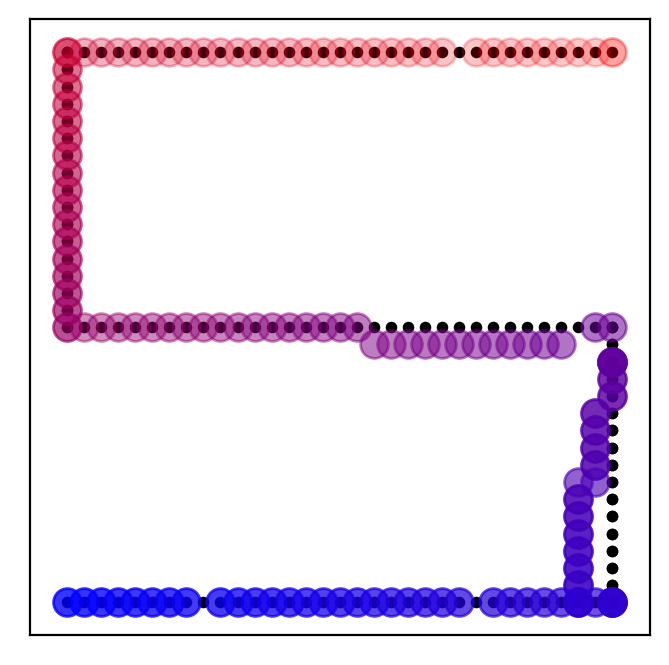}
	\end{subfigure}%
	\vspace{1 mm}
	\hspace*{1.2cm}
	\begin{subfigure}[t]{.105\textwidth}
		\centering
		\begin{overpic}[width=\textwidth]{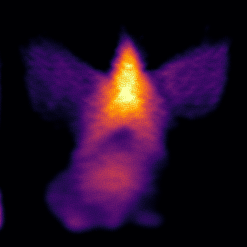}
			\put (3,25) {\textcolor{white}{\scriptsize $\uparrow$}}
			\put (2,12) {\textcolor{white}{\scriptsize $y$}}
			\put (12,2) {\textcolor{white}{\scriptsize $x \rightarrow$}}
			\put (-70,45) {{\scriptsize \begin{tabular}{@{}c@{}} Initialization \\ 2 \end{tabular}}}
		\end{overpic}
		\vspace*{-20pt} 
	\end{subfigure}%
	~ 
	\begin{subfigure}[t]{.105\textwidth}
		\centering
		\scalebox{1}[1]{\includegraphics[width=\textwidth]{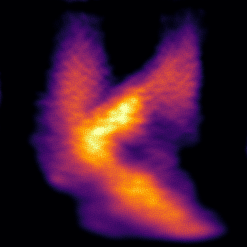}}
	\end{subfigure}%
	~ 
	\begin{subfigure}[t]{.105\textwidth}
		\centering
		\scalebox{1}[1]{\includegraphics[width=\textwidth]{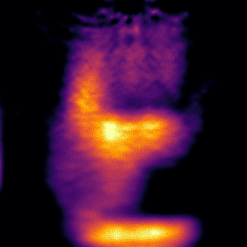}}
	\end{subfigure}%
	~ 
	\begin{subfigure}[t]{.105\textwidth}
		\centering
		\includegraphics[width=\textwidth]{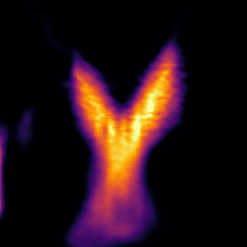}
	\end{subfigure}%
	~
	\begin{subfigure}[t]{.105\textwidth}
		\centering
		\begin{overpic}[width=\textwidth]{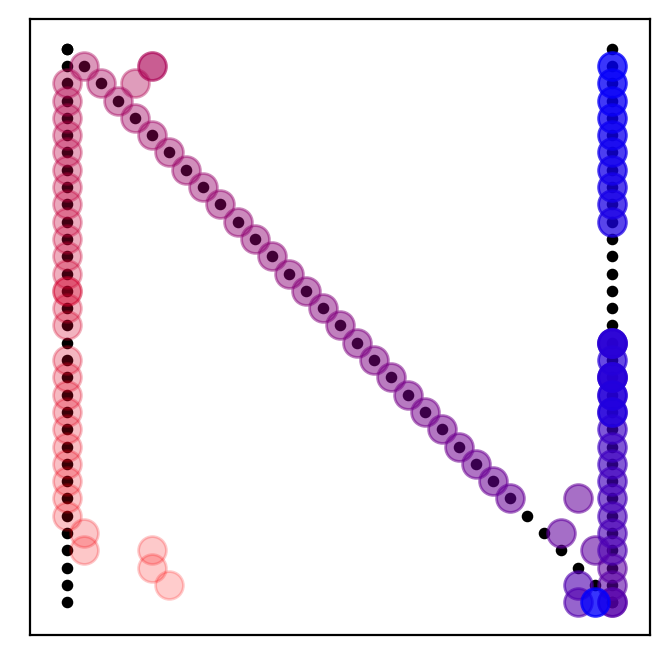}
			\put (-7,25) {{\scriptsize $\uparrow$}}
			\put (-7,12) {{\scriptsize $x$}}				
			\put (12,-5) {{\scriptsize $z \rightarrow$}}
		\end{overpic}	
	\end{subfigure}%
	~ 
	\begin{subfigure}[t]{.105\textwidth}
		\centering
		\scalebox{1}[1]{\includegraphics[width=\textwidth]{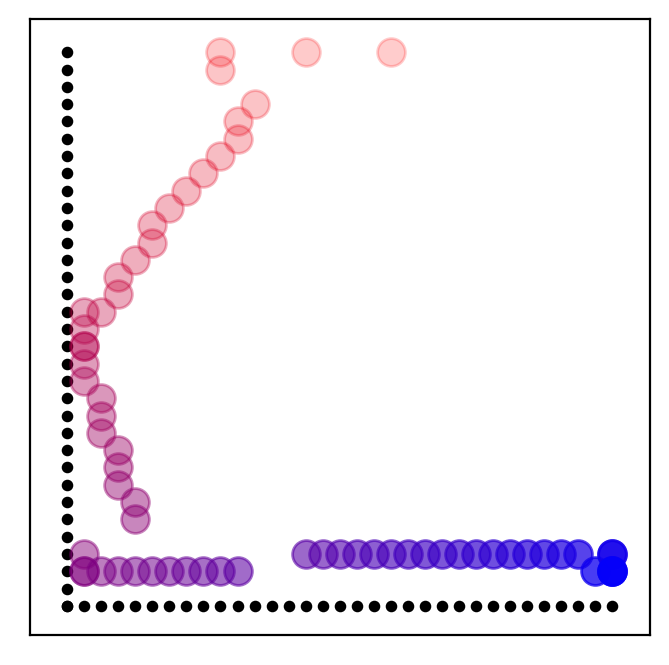}}
	\end{subfigure}%
	~ 
	\begin{subfigure}[t]{.105\textwidth}
		\centering
		\scalebox{1}[1]{\includegraphics[width=\textwidth]{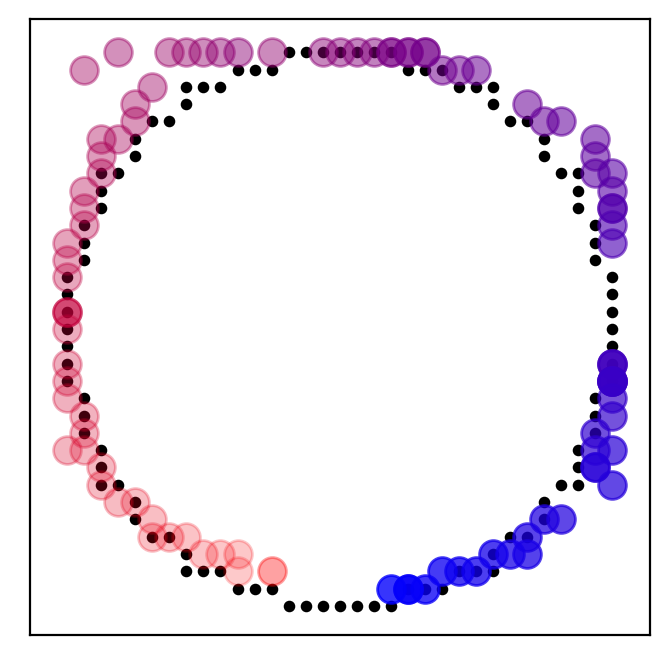}}
	\end{subfigure}%
	~ 
	\begin{subfigure}[t]{.105\textwidth}
		\centering
		\includegraphics[width=\textwidth]{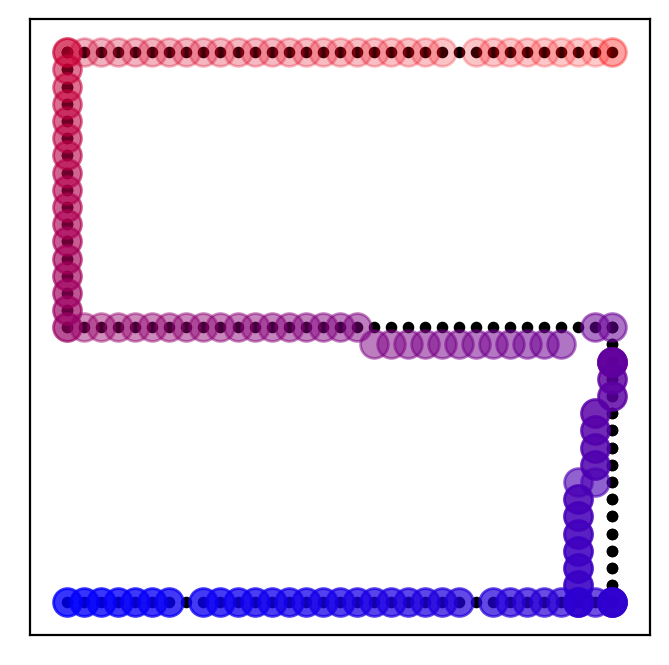}
	\end{subfigure}%
	\vspace{1 mm}	
	\hspace*{1.2cm}
	\begin{subfigure}[t]{.105\textwidth}
		\centering
		\begin{overpic}[width=\textwidth]{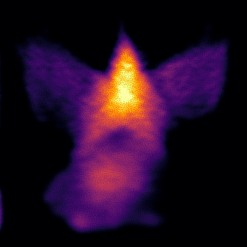}
			\put (3,25) {\textcolor{white}{\scriptsize $\uparrow$}}
			\put (2,12) {\textcolor{white}{\scriptsize $y$}}
			\put (12,2) {\textcolor{white}{\scriptsize $x \rightarrow$}}
			\put (-70,45) {{\scriptsize \begin{tabular}{@{}c@{}} Initialization \\ 3 \end{tabular}}}
		\end{overpic}
		\vspace*{-20pt} 
	\end{subfigure}%
	~ 
	\begin{subfigure}[t]{.105\textwidth}
		\centering
		\scalebox{1}[1]{\includegraphics[width=\textwidth]{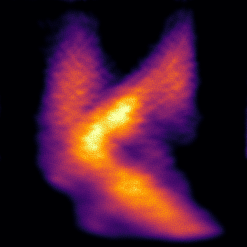}}
	\end{subfigure}%
	~ 
	\begin{subfigure}[t]{.105\textwidth}
		\centering
		\scalebox{1}[1]{\includegraphics[width=\textwidth]{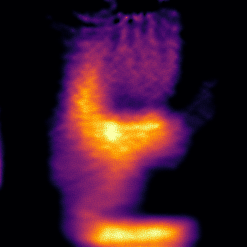}}
	\end{subfigure}%
	~ 
	\begin{subfigure}[t]{.105\textwidth}
		\centering
		\includegraphics[width=\textwidth]{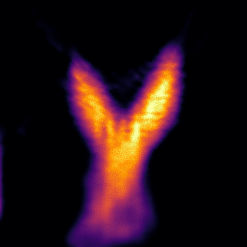}
	\end{subfigure}%
	~
	\begin{subfigure}[t]{.105\textwidth}
		\centering
		\begin{overpic}[width=\textwidth]{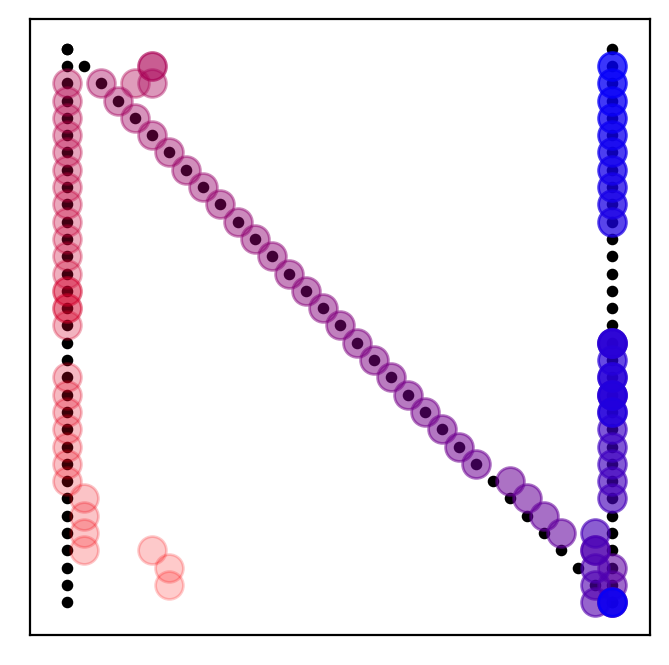}
			\put (-7,25) {{\scriptsize $\uparrow$}}
			\put (-7,12) {{\scriptsize $x$}}				
			\put (12,-5) {{\scriptsize $z \rightarrow$}}
		\end{overpic}	
	\end{subfigure}%
	~ 
	\begin{subfigure}[t]{.105\textwidth}
		\centering
		\scalebox{1}[1]{\includegraphics[width=\textwidth]{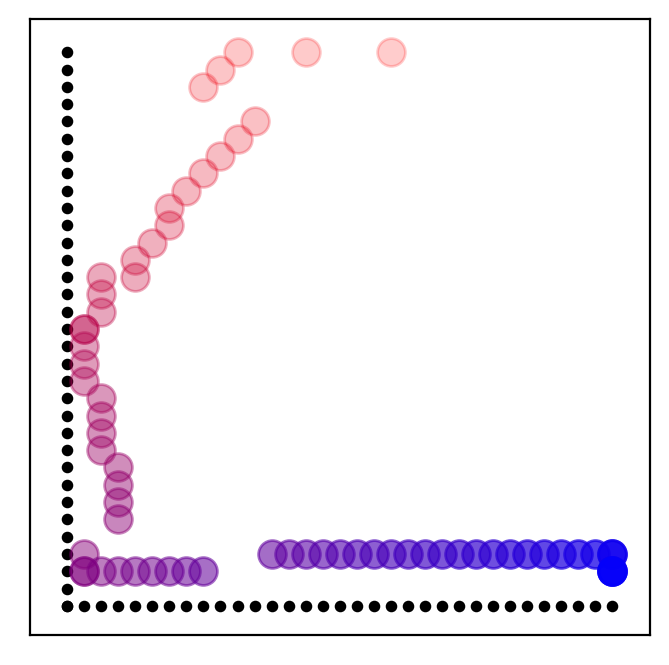}}
	\end{subfigure}%
	~ 
	\begin{subfigure}[t]{.105\textwidth}
		\centering
		\scalebox{1}[1]{\includegraphics[width=\textwidth]{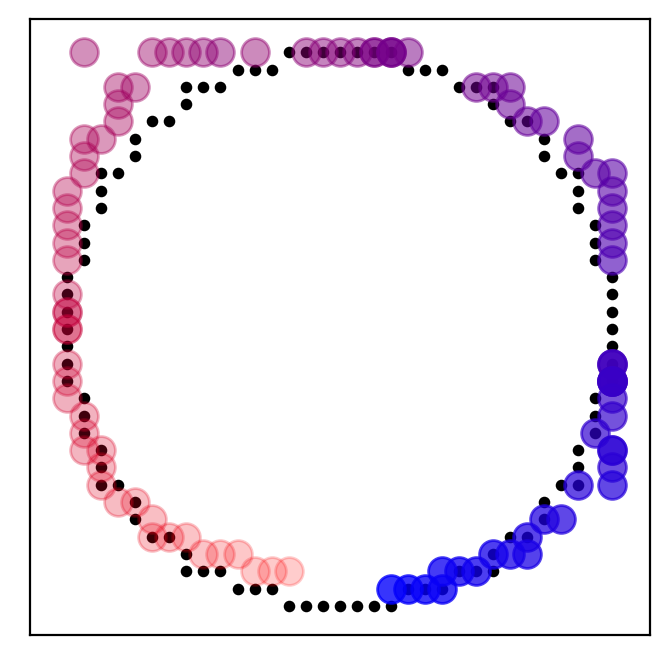}}
	\end{subfigure}%
	~ 
	\begin{subfigure}[t]{.105\textwidth}
		\centering
		\includegraphics[width=\textwidth]{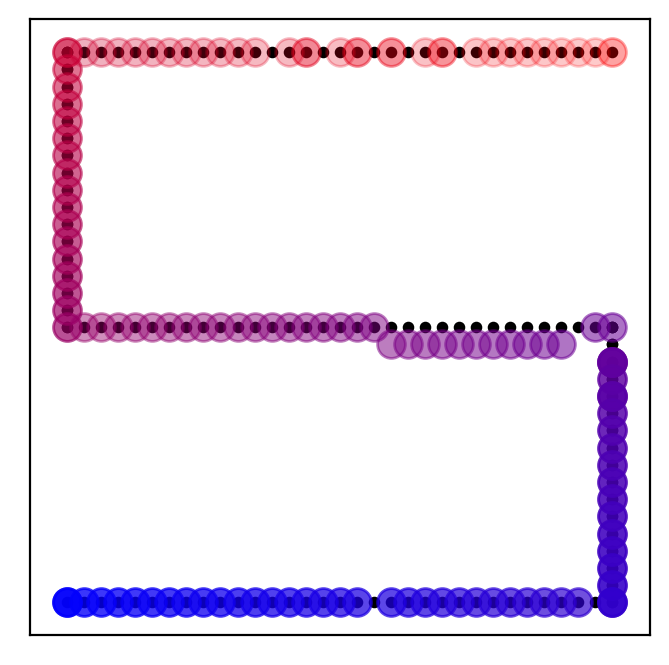}
	\end{subfigure}%
	\vspace{1 mm}
	\hspace*{1.2cm}	
	\begin{subfigure}[t]{.105\textwidth}
		\centering
		\begin{overpic}[width=\textwidth]{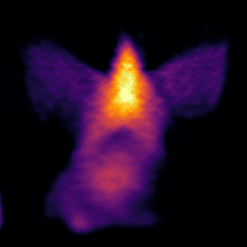}
			\put (3,25) {\textcolor{white}{\scriptsize $\uparrow$}}
			\put (2,12) {\textcolor{white}{\scriptsize $y$}}
			\put (12,2) {\textcolor{white}{\scriptsize $x \rightarrow$}}
			\put (-70,45) {{\scriptsize \begin{tabular}{@{}c@{}} Initialization \\ 4 \end{tabular}}}
		\end{overpic}
		\vspace*{-20pt} 
	\end{subfigure}%
	~ 
	\begin{subfigure}[t]{.105\textwidth}
		\centering
		\scalebox{1}[1]{\includegraphics[width=\textwidth]{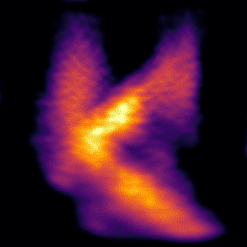}}
	\end{subfigure}%
	~ 
	\begin{subfigure}[t]{.105\textwidth}
		\centering
		\scalebox{1}[1]{\includegraphics[width=\textwidth]{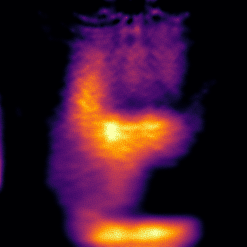}}
	\end{subfigure}%
	~ 
	\begin{subfigure}[t]{.105\textwidth}
		\centering
		\includegraphics[width=\textwidth]{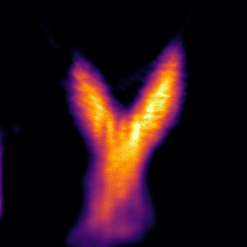}
	\end{subfigure}%
	~
	\begin{subfigure}[t]{.105\textwidth}
		\centering
		\begin{overpic}[width=\textwidth]{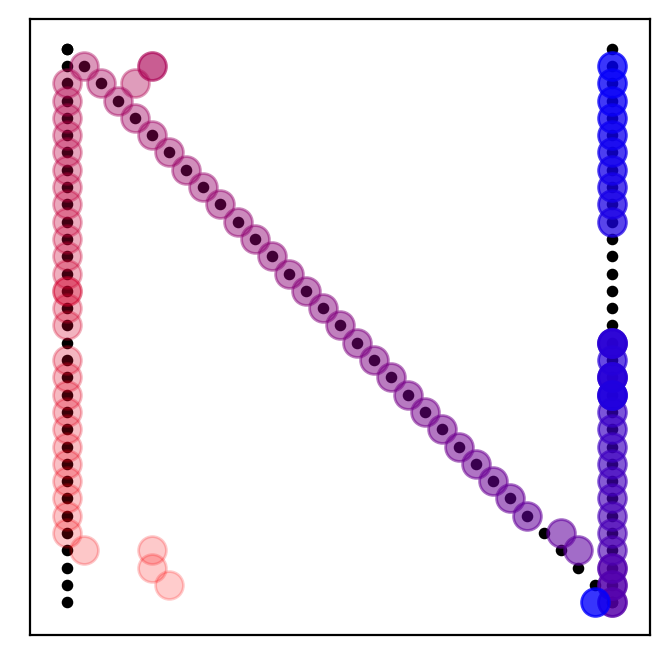}
			\put (-7,25) {{\scriptsize $\uparrow$}}
			\put (-7,12) {{\scriptsize $x$}}				
			\put (12,-5) {{\scriptsize $z \rightarrow$}}
		\end{overpic}	
	\end{subfigure}%
	~ 
	\begin{subfigure}[t]{.105\textwidth}
		\centering
		\scalebox{1}[1]{\includegraphics[width=\textwidth]{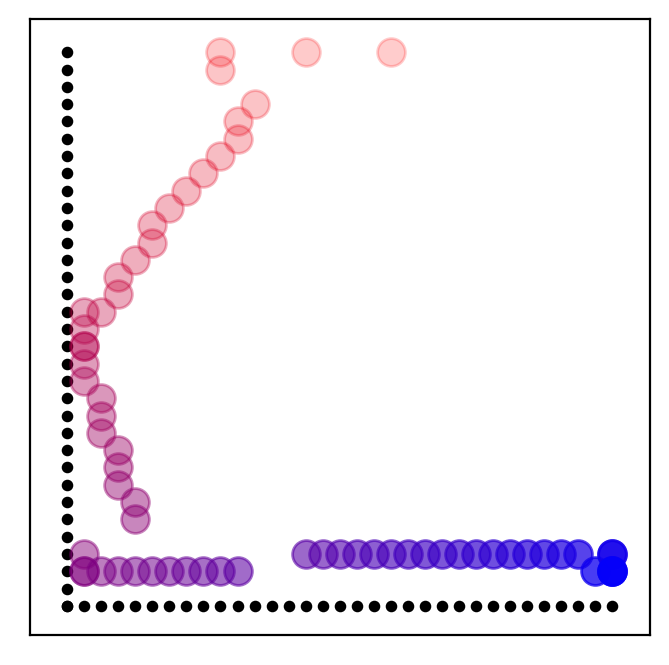}}
	\end{subfigure}%
	~ 
	\begin{subfigure}[t]{.105\textwidth}
		\centering
		\scalebox{1}[1]{\includegraphics[width=\textwidth]{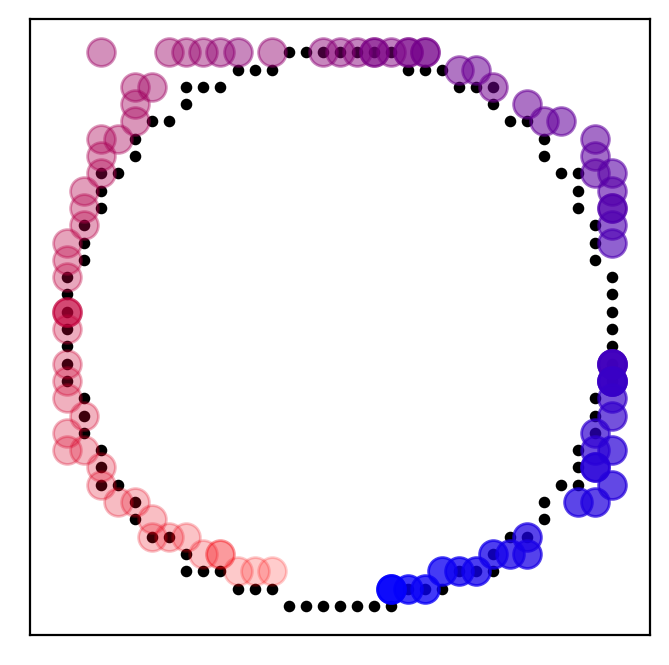}}
	\end{subfigure}%
	~ 
	\begin{subfigure}[t]{.105\textwidth}
		\centering
		\includegraphics[width=\textwidth]{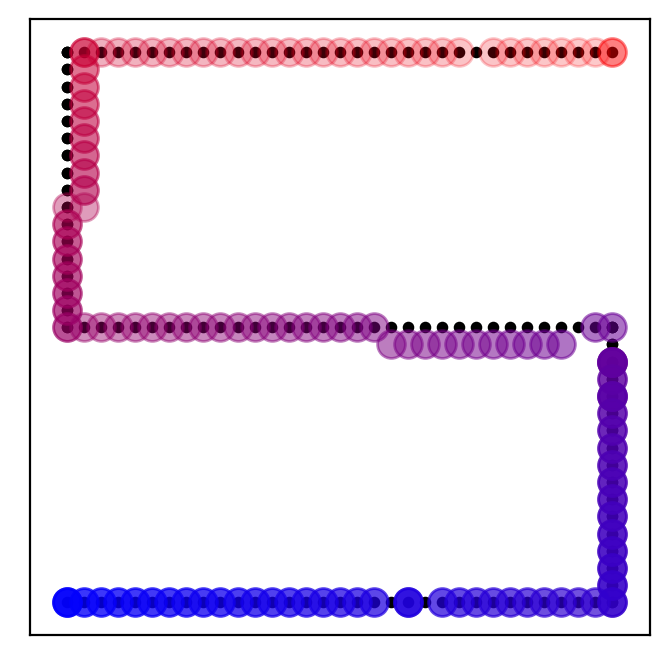}
	\end{subfigure}%
	\caption{\textbf{Reconstructions with different initializations.} Each row displays reconstructions of the same dataset using different random initializations of the albedos. \edit{(The mannequin goes with the ``N'', the ``K'' goes with the ``L'', the ``E'' goes with the ``O'', and the ``Y'' goes with the ``S''.)} Thanks to annealing, our EM algorithm is largely insensitive to how it is initialized.} 
	\label{fig:MultiRecons}
\end{figure*}

\bibliographystyle{IEEEtran}
\bibliography{IEEEabrv,egbib}

\begin{thebibliography}{10}
\providecommand{\url}[1]{#1}
\csname url@samestyle\endcsname
\providecommand{\newblock}{\relax}
\providecommand{\bibinfo}[2]{#2}
\providecommand{\BIBentrySTDinterwordspacing}{\spaceskip=0pt\relax}
\providecommand{\BIBentryALTinterwordstretchfactor}{4}
\providecommand{\BIBentryALTinterwordspacing}{\spaceskip=\fontdimen2\font plus
\BIBentryALTinterwordstretchfactor\fontdimen3\font minus
  \fontdimen4\font\relax}
\providecommand{\BIBforeignlanguage}[2]{{%
\expandafter\ifx\csname l@#1\endcsname\relax
\typeout{** WARNING: IEEEtran.bst: No hyphenation pattern has been}%
\typeout{** loaded for the language `#1'. Using the pattern for}%
\typeout{** the default language instead.}%
\else
\language=\csname l@#1\endcsname
\fi
#2}}
\providecommand{\BIBdecl}{\relax}
\BIBdecl

\bibitem{Velten:2012recovering}
A.~Velten, T.~Willwacher, O.~Gupta, A.~Veeraraghavan, M.~Bawendi, and
  R.~Raskar, ``Recovering three-dimensional shape around a corner using
  ultrafast time-of-flight imaging,'' \emph{Nature Communications}, vol.~3, p.
  745, 2012.

\bibitem{buttafava2015non}
M.~Buttafava, J.~Zeman, A.~Tosi, K.~Eliceiri, and A.~Velten,
  ``Non-line-of-sight imaging using a time-gated single photon avalanche
  diode,'' \emph{Opt. Express}, vol.~23, no.~16, pp. 20\,997--21\,011, 2015.

\bibitem{Gariepy:2016}
G.~Gariepy, F.~Tonolini, R.~Henderson, J.~Leach, and D.~Faccio, ``Detection and
  tracking of moving objects hidden from view,'' \emph{Nature Photonics},
  vol.~10, pp. 23--26, 2016.

\bibitem{OToole:2018}
M.~O'Toole, D.~B. Lindell, and G.~Wetzstein, ``Confocal non-line-of-sight
  imaging based on the light cone transform,'' \emph{Nature}, pp. 338--341,
  2018.

\bibitem{xin2019theory}
S.~Xin, S.~Nousias, K.~N. Kutulakos, A.~C. Sankaranarayanan, S.~G. Narasimhan,
  and I.~Gkioulekas, ``A theory of fermat paths for non-line-of-sight shape
  reconstruction,'' in \emph{Proceedings of the IEEE Conference on Computer
  Vision and Pattern Recognition}, 2019, pp. 6800--6809.

\bibitem{Liu:2019}
X.~Liu, I.~Guillen, M.~L. Manna, J.~H. Nam, S.~A. Reza, T.~H. Le, A.~Jarabo,
  D.~Gutierrez, and A.~Velten, ``Non-line-of-sight imaging using phasor-field
  virtual wave optics,'' \emph{Nature}, vol. 572, pp. 620--623, 2019.

\bibitem{Lindell:2019:Wave}
D.~B. Lindell, G.~Wetzstein, and M.~O’Toole, ``Wave-based non-line-of-sight
  imaging using fast f--k migration,'' \emph{ACM Trans. Graph. (SIGGRAPH)},
  vol.~38, no.~4, p. 116, 2019.

\bibitem{faccio2020non}
D.~Faccio, A.~Velten, and G.~Wetzstein, ``Non-line-of-sight imaging,''
  \emph{Nature Reviews Physics}, pp. 1--10, 2020.

\bibitem{young2020non}
S.~I. Young, D.~B. Lindell, B.~Girod, D.~Taubman, and G.~Wetzstein,
  ``Non-line-of-sight surface reconstruction using the directional light-cone
  transform,'' in \emph{Proceedings of the IEEE/CVF Conference on Computer
  Vision and Pattern Recognition}, 2020, pp. 1407--1416.

\bibitem{bouman2017turning}
K.~L. Bouman, V.~Ye, A.~B. Yedidia, F.~Durand, G.~W. Wornell, A.~Torralba, and
  W.~T. Freeman, ``Turning corners into cameras: Principles and methods,'' in
  \emph{Proceedings of the IEEE International Conference on Computer Vision},
  2017, pp. 2270--2278.

\bibitem{prickett1980principles}
M.~J. {Prickett} and C.~C. {Chen}, ``{Principles of inverse synthetic aperture
  radar/ISAR/imaging},'' in \emph{EASCON '80; Electronics and Aerospace Systems
  Conference}, Jan. 1980, pp. 340--345.

\bibitem{liu2019analysis}
X.~Liu, S.~Bauer, and A.~Velten, ``Analysis of feature visibility in
  non-line-of-sight measurements,'' in \emph{Proceedings of the IEEE Conference
  on Computer Vision and Pattern Recognition}, 2019, pp. 10\,140--10\,148.

\bibitem{freund1990looking}
I.~Freund, ``Looking through walls and around corners,'' \emph{Physica A:
  Statistical Mechanics and its Applications}, vol. 168, no.~1, pp. 49--65,
  1990.

\bibitem{Kirmani:2009}
A.~Kirmani, T.~Hutchison, J.~Davis, and R.~Raskar, ``Looking around the corner
  using transient imaging,'' in \emph{2009 IEEE 12th International Conference
  on Computer Vision}.\hskip 1em plus 0.5em minus 0.4em\relax IEEE, 2009, pp.
  159--166.

\bibitem{Velten:2012:Visualizing}
A.~Velten, D.~Wu, A.~Jarabo, B.~Masia, C.~Barsi, C.~Joshi, E.~Lawson,
  M.~Bawendi, D.~Gutierrez, and R.~Raskar, ``Femto-photography: Capturing and
  visualizing the propagation of light,'' \emph{ACM Trans. Graph.}, vol.~32,
  2013.

\bibitem{Pandharkar:2011}
R.~Pandharkar, A.~Velten, A.~Bardagjy, E.~Lawson, M.~Bawendi, and R.~Raskar,
  ``Estimating motion and size of moving non-line-of-sight objects in cluttered
  environments,'' in \emph{Proceedings of the IEEE International Conference on
  Computer Vision}.\hskip 1em plus 0.5em minus 0.4em\relax IEEE, 2011, pp.
  265--272.

\bibitem{Wu:2014}
D.~Wu, G.~Wetzstein, C.~Barsi, T.~Willwacher, Q.~Dai, and R.~Raskar,
  ``Ultra-fast lensless computational imaging through 5d frequency analysis of
  time-resolved light transport,'' \emph{Int. J. Comput. Vision}, vol. 110,
  no.~2, pp. 128--140, 2014.

\bibitem{Gupta:12}
O.~Gupta, T.~Willwacher, A.~Velten, A.~Veeraraghavan, and R.~Raskar,
  ``Reconstruction of hidden 3d shapes using diffuse reflections,'' \emph{Opt.
  Express}, vol.~20, no.~17, pp. 19\,096--19\,108, Aug 2012.

\bibitem{laurenzis2015multiple}
M.~Laurenzis, J.~Klein, E.~Bacher, and N.~Metzger, ``Multiple-return
  single-photon counting of light in flight and sensing of non-line-of-sight
  objects at shortwave infrared wavelengths,'' \emph{Optics letters}, vol.~40,
  no.~20, pp. 4815--4818, 2015.

\bibitem{tsai2017geometry}
C.-Y. Tsai, K.~N. Kutulakos, S.~G. Narasimhan, and A.~C. Sankaranarayanan,
  ``The geometry of first-returning photons for non-line-of-sight imaging,'' in
  \emph{Proceedings of the IEEE Conference on Computer Vision and Pattern
  Recognition}, 2017, pp. 7216--7224.

\bibitem{arellano2017fast}
V.~Arellano, D.~Gutierrez, and A.~Jarabo, ``Fast back-projection for non-line
  of sight reconstruction,'' \emph{Opt. Express}, vol.~25, no.~10, pp.
  11\,574--11\,583, 2017.

\bibitem{pediredla2017reconstructing}
A.~K. Pediredla, M.~Buttafava, A.~Tosi, O.~Cossairt, and A.~Veeraraghavan,
  ``Reconstructing rooms using photon echoes: A plane based model and
  reconstruction algorithm for looking around the corner,'' in \emph{2017 IEEE
  International Conference on Computational Photography (ICCP)}.\hskip 1em plus
  0.5em minus 0.4em\relax IEEE, 2017, pp. 1--12.

\bibitem{Xu:18}
F.~Xu, G.~Shulkind, C.~Thrampoulidis, J.~H. Shapiro, A.~Torralba, F.~N.~C.
  Wong, and G.~W. Wornell, ``Revealing hidden scenes by photon-efficient
  occlusion-based opportunistic active imaging,'' \emph{Opt. Express}, vol.~26,
  no.~8, pp. 9945--9962, 2018.

\bibitem{jin2015recovering}
C.~Jin, Z.~Song, S.~Zhang, J.~Zhai, and Y.~Zhao, ``Recovering three-dimensional
  shape through a small hole using three laser scatterings,'' \emph{Optics
  letters}, vol.~40, no.~1, pp. 52--55, 2015.

\bibitem{heide2019non}
F.~Heide, M.~O'Toole, K.~Zang, D.~B. Lindell, S.~Diamond, and G.~Wetzstein,
  ``Non-line-of-sight imaging with partial occluders and surface normals,''
  \emph{ACM Trans. Graph.}, vol.~38, no.~3, p.~22, 2019.

\bibitem{heide2014diffuse}
F.~Heide, L.~Xiao, W.~Heidrich, and M.~B. Hullin, ``Diffuse mirrors: 3d
  reconstruction from diffuse indirect illumination using inexpensive
  time-of-flight sensors,'' in \emph{Proceedings of the IEEE Conference on
  Computer Vision and Pattern Recognition}, 2014, pp. 3222--3229.

\bibitem{kadambi2016occluded}
A.~Kadambi, H.~Zhao, B.~Shi, and R.~Raskar, ``Occluded imaging with
  time-of-flight sensors,'' \emph{ACM Trans. Graph.}, vol.~35, no.~2, p.~15,
  2016.

\bibitem{bertolotti2012non}
J.~Bertolotti, E.~G. van Putten, C.~Blum, A.~Lagendijk, W.~L. Vos, and A.~P.
  Mosk, ``Non-invasive imaging through opaque scattering layers,''
  \emph{Nature}, vol. 491, no. 7423, p. 232, 2012.

\bibitem{katz2012looking}
O.~Katz, E.~Small, and Y.~Silberberg, ``Looking around corners and through thin
  turbid layers in real time with scattered incoherent light,'' \emph{Nature
  Photonics}, vol.~6, no.~8, pp. 549--553, 2012.

\bibitem{katz2014non}
O.~Katz, P.~Heidmann, M.~Fink, and S.~Gigan, ``Non-invasive single-shot imaging
  through scattering layers and around corners via speckle correlations,''
  \emph{Nature Photonics}, vol.~8, no.~10, p. 784, 2014.

\bibitem{PrasannaCosi}
A.~Viswanath, P.~Rangarajan, D.~MacFarlane, and M.~P. Christensen, ``Indirect
  imaging using correlography,'' in \emph{Proc. COSI}, 2018.

\bibitem{metzler2020deep}
C.~A. Metzler, F.~Heide, P.~Rangarajan, M.~M. Balaji, A.~Viswanath,
  A.~Veeraraghavan, and R.~G. Baraniuk, ``Deep-inverse correlography: towards
  real-time high-resolution non-line-of-sight imaging,'' \emph{Optica}, vol.~7,
  no.~1, pp. 63--71, 2020.

\bibitem{lindell2019acoustic}
D.~B. Lindell, G.~Wetzstein, and V.~Koltun, ``Acoustic non-line-of-sight
  imaging,'' in \emph{Proceedings of the IEEE Conference on Computer Vision and
  Pattern Recognition}, 2019, pp. 6780--6789.

\bibitem{an2019diffraction}
I.~An, D.~Lee, J.-w. Choi, D.~Manocha, and S.-e. Yoon, ``Diffraction-aware
  sound localization for a non-line-of-sight source,'' in \emph{2019
  International Conference on Robotics and Automation (ICRA)}.\hskip 1em plus
  0.5em minus 0.4em\relax IEEE, 2019, pp. 4061--4067.

\bibitem{torralba2012accidental}
A.~Torralba and W.~T. Freeman, ``Accidental pinhole and pinspeck cameras:
  Revealing the scene outside the picture,'' in \emph{2012 IEEE Conference on
  Computer Vision and Pattern Recognition}.\hskip 1em plus 0.5em minus
  0.4em\relax IEEE, 2012, pp. 374--381.

\bibitem{batarseh2018passive}
M.~Batarseh, S.~Sukhov, Z.~Shen, H.~Gemar, R.~Rezvani, and A.~Dogariu,
  ``Passive sensing around the corner using spatial coherence,'' \emph{Nature
  Communications}, vol.~9, no.~1, p. 3629, 2018.

\bibitem{Thrampoulidis:2018}
C.~Thrampoulidis, G.~Shulkind, F.~Xu, W.~T. Freeman, J.~H. Shapiro,
  A.~Torralba, F.~N.~C. Wong, and G.~W. Wornell, ``Exploiting occlusion in
  non-line-of-sight active imaging,'' \emph{IEEE Trans. Comput. Imaging},
  vol.~4, no.~3, 2018.

\bibitem{saunders2019computational}
C.~Saunders, J.~Murray-Bruce, and V.~K. Goyal, ``Computational periscopy with
  an ordinary digital camera,'' \emph{Nature}, vol. 565, no. 7740, p. 472,
  2019.

\bibitem{brooks2019single}
J.~Brooks and D.~Faccio, ``A single-shot non-line-of-sight range-finder,''
  \emph{Sensors}, vol.~19, no.~21, p. 4820, 2019.

\bibitem{o2018real}
M.~O'Toole, D.~B. Lindell, and G.~Wetzstein, ``Real-time non-line-of-sight
  imaging,'' in \emph{ACM SIGGRAPH 2018 Emerging Technologies}, 2018, pp. 1--2.

\bibitem{klein2016tracking}
J.~Klein, C.~Peters, J.~Mart{\'\i}n, M.~Laurenzis, and M.~B. Hullin, ``Tracking
  objects outside the line of sight using 2d intensity images,''
  \emph{Scientific Reports}, vol.~6, p. 32491, 2016.

\bibitem{chan2017non}
S.~Chan, R.~E. Warburton, G.~Gariepy, J.~Leach, and D.~Faccio,
  ``Non-line-of-sight tracking of people at long range,'' \emph{Opt. Express},
  vol.~25, no.~9, pp. 10\,109--10\,117, 2017.

\bibitem{smith2018tracking}
B.~M. Smith, M.~O'Toole, and M.~Gupta, ``Tracking multiple objects outside the
  line of sight using speckle imaging,'' in \emph{Proceedings of the IEEE
  Conference on Computer Vision and Pattern Recognition}, 2018, pp. 6258--6266.

\bibitem{Boger:2019}
J.~Boger-Lombard and O.~Katz, ``Passive optical time-of-flight for non
  line-of-sight localization,'' \emph{Nature Communications}, vol.~10, 2019.

\bibitem{basu2000uniqueness}
S.~Basu and Y.~Bresler, ``Uniqueness of tomography with unknown view angles,''
  \emph{IEEE Trans. Image Process.}, vol.~9, no.~6, pp. 1094--1106, June 2000.

\bibitem{cheng2015primer}
Y.~Cheng, N.~Grigorieff, P.~A. Penczek, and T.~Walz, ``A primer to
  single-particle cryo-electron microscopy,'' \emph{Cell}, vol. 161, no.~3, pp.
  438--449, 2015.

\bibitem{singer2018mathematics}
A.~Singer, ``Mathematics for cryo-electron microscopy,'' \emph{arXiv preprint
  arXiv:1803.06714}, 2018.

\bibitem{wang2019two}
L.~Wang and Z.~Zhao, ``Two-dimensional tomography from noisy projection tilt
  series taken at unknown view angles with non-uniform distribution,'' in
  \emph{2019 IEEE International Conference on Image Processing (ICIP)}.\hskip
  1em plus 0.5em minus 0.4em\relax IEEE, 2019, pp. 1242--1246.

\bibitem{bandeira2017optimal}
A.~S. Bandeira, J.~Niles-Weed, and P.~Rigollet, ``Optimal rates of estimation
  for multi-reference alignment,'' \emph{Mathematical Statistics and Learning},
  vol.~2, no.~1, pp. 25--75, 2020.

\bibitem{zehni2019geometric}
M.~Zehni, S.~Huang, I.~Dokmani{\'c}, and Z.~Zhao, ``Geometric invariants for
  sparse unknown view tomography,'' in \emph{ICASSP 2019-2019 IEEE
  International Conference on Acoustics, Speech and Signal Processing
  (ICASSP)}.\hskip 1em plus 0.5em minus 0.4em\relax IEEE, 2019, pp. 5027--5031.

\bibitem{coifman2008graph}
R.~R. Coifman, Y.~Shkolnisky, F.~J. Sigworth, and A.~Singer, ``Graph laplacian
  tomography from unknown random projections,'' \emph{IEEE Trans. Image
  Process.}, vol.~17, no.~10, pp. 1891--1899, 2008.

\bibitem{singer2013two}
A.~Singer and H.-T. Wu, ``Two-dimensional tomography from noisy projections
  taken at unknown random directions,'' \emph{SIAM J. Imaging Sci.}, vol.~6,
  no.~1, pp. 136--175, 2013.

\bibitem{basu2000feasibility}
S.~Basu and Y.~Bresler, ``Feasibility of tomography with unknown view angles,''
  \emph{IEEE Trans. Image Process.}, vol.~9, no.~6, pp. 1107--1122, 2000.

\bibitem{scheres2012relion}
S.~H. Scheres, ``Relion: implementation of a {Bayesian} approach to cryo-{EM}
  structure determination,'' \emph{Journal of Structural Biology}, vol. 180,
  no.~3, pp. 519--530, 2012.

\bibitem{dellaert2002expectation}
F.~Dellaert, ``The expectation maximization algorithm,'' Georgia Institute of
  Technology, Tech. Rep., 2002.

\bibitem{NEURIPS2019_9015}
A.~Paszke, S.~Gross, F.~Massa, A.~Lerer, J.~Bradbury, G.~Chanan, T.~Killeen,
  Z.~Lin, N.~Gimelshein, L.~Antiga, A.~Desmaison, A.~Kopf, E.~Yang, Z.~DeVito,
  M.~Raison, A.~Tejani, S.~Chilamkurthy, B.~Steiner, L.~Fang, J.~Bai, and
  S.~Chintala, ``Pytorch: An imperative style, high-performance deep learning
  library,'' in \emph{Advances in Neural Information Processing Systems
  32}.\hskip 1em plus 0.5em minus 0.4em\relax Curran Associates, Inc., 2019,
  pp. 8024--8035.

\bibitem{kingma2014adam}
D.~P. Kingma and J.~Ba, ``Adam: A method for stochastic optimization,''
  \emph{arXiv preprint arXiv:1412.6980}, 2014.

\bibitem{ueda1998deterministic}
N.~Ueda and R.~Nakano, ``Deterministic annealing em algorithm,'' \emph{Neural
  Networks}, vol.~11, no.~2, pp. 271--282, 1998.

\bibitem{thoma2017hasyv2}
M.~Thoma, ``The hasyv2 dataset,'' \emph{arXiv preprint arXiv:1701.08380}, 2017.

\bibitem{ssim}
Z.~Wang, A.~C. Bovik, H.~R. Sheikh, and E.~P. Simoncelli, ``Image quality
  assessment: from error visibility to structural similarity,'' \emph{IEEE
  Trans. Image Process.}, vol.~13, no.~4, pp. 600--612, 2004.

\bibitem{RealTimeUnknownCOSI}
C.~A. Metzler and G.~Wetzstein, ``Real-time unknown-view tomography using
  recurrent neural networks with applications to keyhole imaging,'' in
  \emph{Computational Optical Sensing and Imaging}.\hskip 1em plus 0.5em minus
  0.4em\relax Optical Society of America, 2020, pp. CTh5C--1.

\bibitem{PrasannaRevealReport}
P.~Rangarajan and M.~P. Christensen, ``Obtaining multipath \& non-line-of-sight
  information by sensing coherence \& intensity with emerging novel
  techniques,'' Southern Methodist University, Tech. Rep., 2020.

\bibitem{BelcourBRDFMeasurements}
L.~Belcour, R.~Pacanowski, M.~Delahaie, A.~Laville-Geay, and L.~Eupherte,
  ``Brdf measurements and analysis of retroreflective materials,''
  \emph{Journal of the Optical Society of America}, vol.~31, pp. 2561--2572,
  2014.

\bibitem{ansi2014american}
``{American National Standard for Safe Use of Lasers ANSI Z136.}'' American
  National Standards Institute, Washington, DC, Standard, 2014.

\end{thebibliography}

\end{document}